\documentclass{article}

\PassOptionsToPackage{numbers, compress}{natbib}
\usepackage{arxiv}
\usepackage{natbib}

\usepackage{algorithm}
\usepackage{algpseudocode}
\usepackage[utf8]{inputenc} 
\usepackage[T1]{fontenc}    
\usepackage{hyperref}       
\usepackage{url}            
\usepackage[normalem]{ulem}
\usepackage{booktabs}       
\usepackage{amsfonts}       
\usepackage{microtype}      
\usepackage{enumitem}
\usepackage[table,xcdraw]{xcolor} 
\usepackage[T1]{fontenc}
\usepackage{graphicx}
\usepackage{multirow}
\usepackage{subcaption}

\usepackage{amsmath,amssymb,amsthm,mathtools}
\newtheorem{theorem}{Theorem}
\newtheorem{proposition}{Proposition}

\theoremstyle{definition}
\newtheorem{definition}{Definition}
\newtheorem{remark}{Remark}

\newcommand{\Tdist}{\mathcal{T}}
\newcommand{\Rdist}{\mathcal{R}}
\newcommand{\E}{\mathbb{E}}
\newcommand{\R}{\mathbb{R}}
\newcommand{\KL}{D_{\mathrm{KL}}}
\newcommand{\Dtrain}{\mathcal{D}_{\mathrm{train}}}

\newcommand{\F}{\mathbf{F}}
\newcommand{\U}{\mathbf{U}}
\newcommand{\V}{\mathbf{V}}

\newcommand{\I}{\mathbf{I}}
\newcommand{\norm}[1]{\left\lVert #1 \right\rVert}

\newcommand{\argmax}{\mathop{\mathrm{arg\,max}}}
\newcommand{\argmin}{\mathop{\mathrm{arg\,min}}}

\renewcommand{\t}{\theta}

\newcommand{\Dfish}{\ensuremath{D_{\mathrm{F}}}}
\newcommand{\Dkl}{\ensuremath{D_{\mathrm{KL}}}}
\newcommand{\Dtask}{\ensuremath{D_{\mathrm{task}}}}

\newcommand{\tr}{\mathrm{tr}}
\newcommand{\diag}{\mathrm{diag}}
\newcommand{\safe}{\textsc{Safe-Drift}}

\title{\safe: Data Selection for Supervised Fine-tuning with Controllable Off-Target Drifts}

\author{%
  Yeo Jin Jung$^*$ \\
  Department of Statistics \\
  University of Chicago \\
  \texttt{yeojinjung@uchicago.edu} \\
  \And
  Yating Liu$^*$ \\
  Department of Statistics \\
  University of Chicago \\
  \texttt{yatingliu@uchicago.edu} \\
  \AND
  Lalchand Pandia \\
  Department of Computer Science \\
  University of Chicago \\
  \texttt{lcpandia@uchicago.edu} \\
  \And
  Claire Donnat \\
  Department of Statistics \\
  University of Chicago \\
  \texttt{cdonnat@uchicago.edu} \\
}

\begin{document}

\maketitle

\begin{abstract}
 
We propose \safe, a data-selection framework for supervised fine-tuning that improves a target capability while controlling unwanted drift in the model's other behaviors. We formalize this as a constrained optimization problem that maximizes alignment with the target gradient, subject to a budget on off-target drift which we quantify using the Fisher information of a reference distribution. Under this objective, the optimal, closed-form update is a damped natural-gradient step that downweights reference-sensitive directions. To estimate these high-dimensional gradients and work within the right geometry,  we further derive a scalable algorithm that approximates the Fisher information and the target gradient in a shared low-dimensional subspace and analyze how entanglement of the target and reference task can affect the update. We evaluate \textsc{Safe-Drift} against state-of-the-art data selection methods across diverse domains including coding and medical question answering, yielding competitive target performance while reducing degradation on specified off-target behaviors.
\end{abstract}

\section{Introduction}\label{sec:intro}
Supervised fine-tuning (SFT) is an effective technique to adapt large language models (LLMs) to specialized tasks. In many of such tasks, a critical bottleneck is the scarcity of high-quality examples that precisely define the desired behavior. For instance, in medicine, clinician-validated cases may serve as gold-standard target examples; larger biomedical corpora may provide a much wider but noisier training pool~\citep{zhang2023alpacare, singhal2023large}. Similarly, in law, a small number of expert-annotated examples may define the target task, while larger collections of legal documents may offer a broader but less aligned training pool  \cite{colombo2024saullm,niklaus2024flawn,zheng2021does}. In all of these examples,  estimates of the target distribution are high-value and expensive to label, whereas the available training pool is broader, cheaper, and only partially aligned with the task of interest. This pool of candidate examples, which we refer to as the candidate pool, can provide a useful training signal, but training on all of it is not necessarily safe or effective. Carefully selected subsets have indeed been shown to outperform training on the full candidate pool in various settings \citep{xia2024less,biderman2024loralearnsforgets,zhou2023lima, ye2025limoreasoning} (see Table~\ref{tab:main_results} for additional examples). 

At the same time, fine-tuning does not solely affect the capability we intend to improve \citep{zhang2026instruction}. Growing evidence shows that seemingly benign finetuning can weaken safety alignment and degrade behavior outside the target task \citep{qi2023fine,betley2026training}. Beyond safety, fine-tuning can also lead to the forgetting of more general capabilities \citep{luo2025empirical}. We refer to these unintended changes as \emph{off-target drifts}.

These observations suggest that fine-tuning is not just a matter of collecting more data, but selecting the right ones: given a large and noisy candidate pool, we must choose examples that improve the target task without inducing unnecessary off-target drift. The key question is therefore: \textbf{\emph{which examples from the candidate pool should we train on in order to improve the target task while safeguarding other model behaviors?}}

Existing data selection methods broadly fall into two (and often mutually exclusive) families. The first selects candidate examples that appear most valuable for the target task, either through explicit quality proxies \citep{cao2023instruction,wenzek2020ccnet,wettig2024qurating,xie2023dataselectionlanguagemodels} or through direct influence estimation \citep{xia2024less,ilyas2022datamodels}. The second prioritizes diversity or distributional coverage of selected candidates to improve generalization \citep{jung2025prismatic,yang2025data}. We refer the reader to \cite{albalak2024survey} for a comprehensive review. 
None of these methods explicitly constrain the off-target drift induced by the selected subset, nor do they account for the possible entanglement between target-improving and reference-sensitive parameter directions \citep{park2026dualsteering,siu2025steeringsafety,wu2026knowing,kozlowski2025semantic}.

\begin{figure}
    \centering
\includegraphics[width=0.9\linewidth]{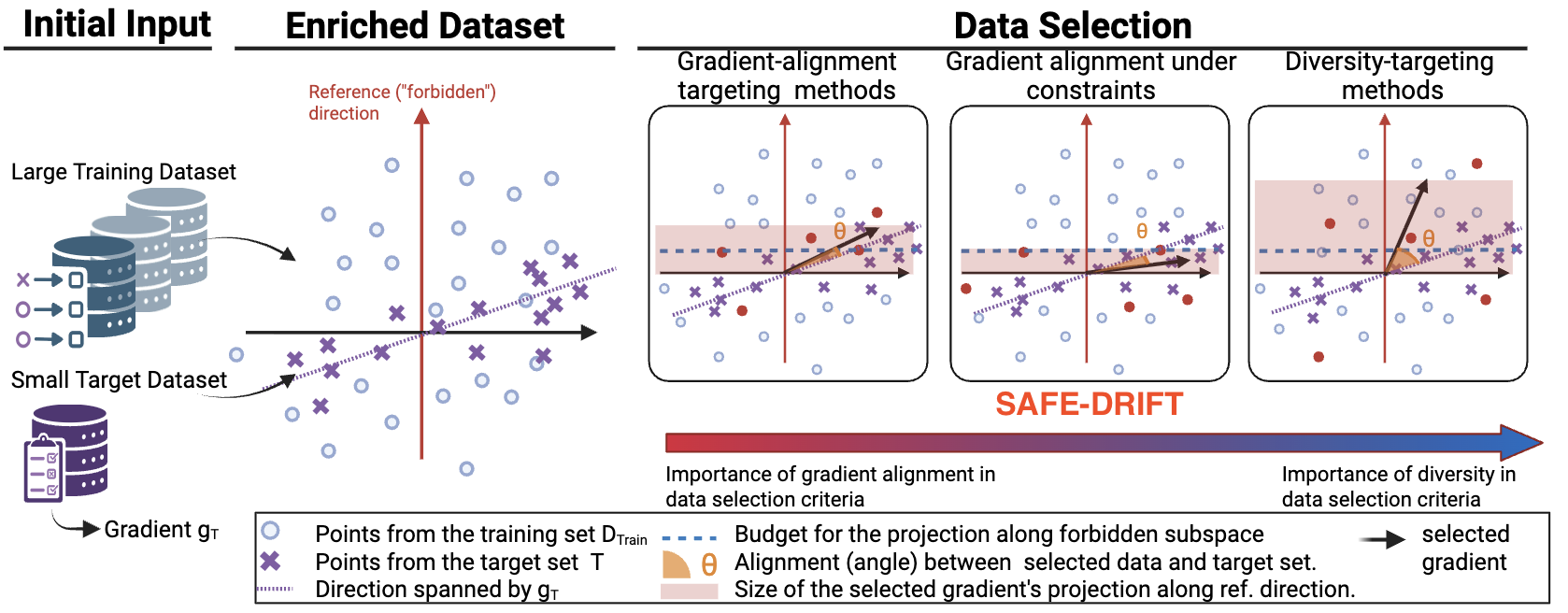}
    \vspace{-2mm}
    \caption{\textit{Trade-offs in data selection: gradient alignment selects candidate gradients aligned with the task but can cause substantial off-target drift (e.g., movement along a "forbidden dimension", in red). Diversity-based selection can mitigate this but does not reliably prevent off-target drift. \safe~combines both approaches by constraining updates along the forbidden dimension. } }
    \vspace{-2mm} 
    \label{fig:pipeline}
\end{figure}

If the gradients that yield the largest improvement in the target task overlap with directions associated with worse performance on a reference task, then aggressively selecting them can induce substantial off-target drift. Diversity-based selection, on the other hand, spreads the update across many gradient directions, but may not deliver the best target performance.
Optimal data selection must therefore navigate potential tensions between target improvement and reference degradation.
Here, we propose an algorithm that explicitly models data selection as a constrained optimization problem to limit unwanted movement along reference-sensitive directions, as illustrated in Figure~\ref{fig:pipeline}.

{\bf Contributions.}  We summarize our contributions as follows:
\begin{itemize}[noitemsep, topsep=0em, leftmargin=1em]
    \item \textbf{A drift-constrained formulation of data selection.} 
    We introduce \safe, which selects data that maximizes target gradient alignment subject to an explicit budget on off-target drift, quantified by the Fisher information of a user-specified reference distribution. (Section~\ref{sec:methods}).
    \item \textbf{A connection between target-aligned and diversity-based selection.}
    We show that the optimal update under a reference-drift budget is a damped natural-gradient direction. This clarifies when target-aligned selection cannot safeguard against off-target drift, when diversity can reduce the drift, and how both arise as limiting or special cases of a single constrained update problem (Section~\ref{subsec:oracle}).
    \item \textbf{A scalable low-rank implementation.} We build a shared low-rank basis that merges leading reference-sensitive directions with residual task-relevant directions (Section~\ref{sec:real}) and work with gradients projected onto the derived low-rank basis, making the selection algorithm practical. We also propose diagnostics of target--reference entanglement using the low-rank projections and analyze how it affects the off-target drift control.
    \item \textbf{Experiments demonstrating controllable drift.}
    Across four settings---code generation, medical question answering, emotional support dialogues, and instruction-following experiments---\safe{} achieves competitive target gains while reducing off-target drift on the specified reference behaviors. We provide further robustness and ablation results on the off-target drift budget and different design choices  (Section~\ref{sec:experiments}).
\end{itemize}

\section{Method}\label{sec:methods}
Consider a base language model with parameters $\theta_0 \in \mathbb{R}^d$ that defines a conditional distribution over output tokens $y \in \mathcal{Y}$ given input tokens $x \in \mathcal{X}$. Let $z=(x,y)$ denote an input--output pair, and let $\ell(z;\theta)$ denote the training loss (typically the negative log-likelihood of the output sequence, or of a single answer token in a multiple-choice setting). We assume that we are given a large candidate training pool \(\mathcal{D}_{\mathrm{train}} = \{z_i\}_{i=1}^{N_C}\) and a small target set \(\mathcal{D}_{\mathrm{tar}} = \{z_j\}_{j=1}^{N_T}\) with \(N_T \ll N_C\) that encodes the capability we wish to improve. We are also given a reference set \(\mathcal{D}_{\mathrm{ref}} = \{x_i\}_{i=1}^{N_R}\) of prompts from a reference distribution $\Rdist$ over inputs, on which the model's output distribution should be preserved. Let the target risk at parameter $\theta \in \mathbb{R}^d$ be defined as:
\begin{equation}
\label{eq:target-risk-methods}
L_{\Tdist}(\theta)
:=
\E_{z\sim \Tdist}\bigl[\ell(z;\theta)\bigr]
=
\E_{x\sim \Tdist_X}\E_{y\sim q_{\Tdist}(\cdot\mid x)}\bigl[\ell((x,y);\theta)\bigr],
\end{equation}
where $\Tdist_X$ is the marginal of the target distribution $\Tdist$ on inputs and $q_{\Tdist}(\cdot\mid x)$ denotes the gold-standard conditional distribution. Our goal is to select a subset \(\mathcal S \subset \mathcal{D}_{\mathrm{train}}\) of size $S$ such that fine-tuning on \(\mathcal S\) minimizes the target risk $L_{\Tdist}(\theta)$ while preserving the model's output distribution on $\mathcal{R}$.

We proceed in two stages. We first characterize the optimal local update, which minimizes $L_{\Tdist}(\theta)$ while preserving behavior under $\Rdist$. We then identify a subset $\mathcal{S} \subset \mathcal{D}_{\mathrm{train}}$ of size $S$ whose aggregate effect most closely matches the optimal update. The model is fine-tuned on $\mathcal{S}$ in one pass, without any online reweighting or iterative refinement. The first subsection characterizes the oracle update, assuming access to the full target and reference populations. We then derive the subset-selection algorithm and its low-rank implementation in the noisy empirical setting using the finite datasets $\mathcal{D}_{\mathrm{train}}$, $\mathcal{D}_{\mathrm{tar}}$, and $\mathcal{D}_{\mathrm{ref}}$.

\subsection{Optimal parameter update under a drift budget}
\label{subsec:oracle}

We begin with a population-level oracle problem in which both the population gradient of the target loss $g_{\Tdist} = \E[\nabla_{\theta} L_{\Tdist}(\theta_0)]$ and the reference distribution $\Rdist$ are known exactly. Here $\theta_0 \in \mathbb{R}^d$ denotes the base model parameters, and the fine-tuned model is parametrized by $\theta = \theta_0 + \Delta\theta$ for some update $\Delta\theta \in
\mathbb{R}^d$. The oracle question we ask is: \textit{what is the optimal parameter update $\Delta\theta$ that maximizes the target performance while safeguarding against off-target drift?}
\vspace{-0.3cm}

\paragraph{Off-target drift on reference distribution.}
For the fine-tuned model $\theta_0 + \Delta\theta$, define the expected drift on the reference distribution as the Kullback-Leibler divergence with the base model,
\begin{equation}
\label{eq:off-target-drift}
D_{\Rdist}(\Delta\theta)
:=
\E_{x\sim \Rdist}
\left[
\KL\!\bigl(
 p_{\theta_0}(\cdot\mid x)
 \,\|\,
 p_{\theta_0+\Delta\theta}(\cdot\mid x)
\bigr)
\right].
\end{equation}
The quantity $D_{\Rdist}(\Delta\theta)$ captures the off-target effects of fine-tuning: how much the parameter update $\Delta\theta$ alters model behavior on $\mathcal{R}$, outside the target task.
We note here that $D_{\Rdist}(\Delta\theta)$ is defined globally for any $\Delta \theta$. However, because our results consider small fine-tuning updates, it is natural to study the local behavior of $D_{\Rdist}(\Delta\theta)$ around $\Delta\theta = 0$. We next show that, to second order, off-target drift is a quadratic form induced by the Fisher information of the model on the reference distribution.

\begin{definition}[Reference Fisher]
\label{def:reference-fisher}
Let $s^R_{\theta_0}(x,y) := \nabla_\theta \log p_{\theta_0}(y\mid x).$
The \emph{reference Fisher information} at $\theta_0$ is: 
\begin{equation}
\label{eq:reference-fisher}
\F_{\Rdist}
:=
\E_{x\sim \Rdist}
\E_{y\sim p_{\theta_0}(\cdot\mid x)}
\bigl[s^R_{\theta_0}(x,y)s^R_{\theta_0}(x,y)^\top\bigr].
\end{equation}
\end{definition}

\begin{proposition}[Local reference behavior]
\label{prop:local-kl}
Assume a common parameter-independent support and standard smoothness and integrability conditions so that differentiation may be exchanged with expectation and with the normalization integral near $\theta_0$. Then, as $\Delta\theta\to 0$,
\begin{equation}
\label{eq:local-kl-expansion}
D_{\Rdist}(\Delta\theta)
=
\frac{1}{2}\Delta\theta^\top \F_{\Rdist}\Delta\theta
+ o\bigl(\norm{\Delta\theta}_2^2\bigr).
\end{equation}
\end{proposition}
The smoothness conditions and proof are provided in
Appendix~\ref{app:prop_1}. Proposition~\ref{prop:local-kl} shows that, for small fine-tuning updates, the cost $D_{\Rdist}(\Delta \theta)$ is estimated as a quadratic form with the curvature given by $\F_\Rdist$, rather than the classic Euclidean norm $\|\Delta\theta\|_2^2$:
directions with large curvature under $\F_\Rdist$ induce larger second-order changes in the model's output distribution on $x\sim \Rdist$, whereas directions lying near the null space of $\F_\Rdist$ have small quadratic off-target drift.

\paragraph{Target gain as gradient alignment.}
\label{subsec:target_gain}

Recall the population gradient loss of the target at the base model $\theta_0$:
$g_{\Tdist} := \nabla_\theta L_{\Tdist}(\theta_0).$
For a small update $\Delta \theta$, a first order Taylor approximation at $\theta_0$ gives,
\begin{equation}\label{eq:loss_taylor}
    L_{\Tdist}(\theta_0+\Delta\theta)-L_{\Tdist}(\theta_0)=
g_{\Tdist}^\top \Delta\theta
+o(\|\Delta\theta\|_2).
\end{equation}

Under a Euclidean norm bound and $g_{\Tdist} \neq 0$, the update providing the steepest decrease in target loss is proportional to $g_{\Tdist}$. However, this strategy might induce updates in the space spanned by $\mathbf{F}_{\Rdist}$, creating a potential trade-off between maximizing target accuracy, and minimizing off-target drift.

\paragraph{The oracle update problem.} We construct a bridge of data selection by formulating it as a constrained optimization problem of $\Delta\theta$: we maximize the target-side gain $-g_{\Tdist}^\top \Delta\theta$, while allowing only a small budget of quadratic off-target drift $\frac{1}{2}\Delta\theta^\top \F_{\Rdist}\Delta\theta$ on the reference distribution. This leads to the following oracle problem: {\it among all small parameter updates, choose the one that maximizes target gain while spending at most $\rho$, a fixed drift budget:}
\begin{equation}
\label{eq:oracle-primal}
\Delta\theta^\star
\in
\argmax_{\Delta\theta\in\R^d}
\Bigl\{-g_{\Tdist}^\top \Delta\theta\Bigr\}
\quad\text{subject to}\quad
\frac12\Delta\theta^\top \F_{\Rdist}\Delta\theta \le \rho,
\qquad
\norm{\Delta\theta}_2 \le  \epsilon.
\end{equation}

There are two central parameters: $\rho>0$, which sets the budget on quadratic off-target drift, and $\epsilon>0$ which constrains the magnitude of the update. 

The constraint $\epsilon$ serves two purposes: it ensures that $\Delta \theta$ is small enough for the local approximations in \ref{eq:local-kl-expansion} and \ref{eq:loss_taylor} to hold, and it regularizes the update in directions where $g_{\Tdist}$ lies in the null space of $\mathbf{F}_\mathcal{R}$. In practice, a fixed optimizer and its learning rate $\eta$ can determine $\epsilon$. For stochastic gradient descent, for instance, we have $\|\Delta \theta\|_2 \leq \eta /\sqrt{S} \| \mathbf{G}\|_{\text{op}}$ where $S$ is the selection budget and $\|\mathbf{G}\|_{\text{op}}$ is the spectral norm of the candidate gradient matrix, both quantities that are fixed characteristics of the problem.

With $\epsilon$ fixed, the sole quantity in \ref{eq:oracle-primal} that must be specified is $\rho$, the maximum off-target drift we are willing to tolerate.
To make this quantity interpretable, let $B_x  \subseteq \mathcal{Y}$ be a set of undesirable outputs for prompt $x$, for instance, a set of toxic or stereotyped responses. By Pinsker's and Jensen's inequalities,
\begin{equation*}
    \E_{x\sim \Rdist}[(p_{\t}(B_x\mid x) -p_{\t_0}(B_x\mid x))_+] \leq \sqrt{D_{\Rdist}(\Delta\theta)/2}.
\end{equation*}
For a local oracle update, Appendix~\ref{app:prop_1} gives $D_{\Rdist}(\Delta\theta) \leq \rho+M\epsilon^3/6$, where $M$ is the expected third-derivative envelope and $\epsilon$ lies
within the stated neighborhood.
Therefore, for any undesirable reference output $B_x \subseteq \mathcal{Y}$, the expected increase in the probability of $B_x$ is bounded by $\sqrt{(\rho+M\epsilon^3/6)/2}$. In practice, we treat $\rho$ as a hyperparameter that requires tuning to efficiently safeguard against deleterious off-target drifts; we ablate over a range of $\rho$'s in Section~\ref{sec:experiments}. We now provide the optimal oracle solution of \ref{eq:oracle-primal}.

\begin{proposition}[Oracle Update $\Delta \theta^*$]
\label{prop:oracle-safe-direction} 
Assume the target gradient vector satisfies  $g_{\Tdist} \neq 0$. When both the drift and the norm constraint multipliers are strictly positive, the solution to \eqref{eq:oracle-primal}
is,
\begin{align*}   
\Delta\theta^\star
=
-\epsilon\,
\frac{(\F_\Rdist+\alpha \mathbf{I})^{-1}g_{\Tdist}}
{\|(\F_\Rdist+\alpha \mathbf{I})^{-1}g_{\Tdist}\|_2},
\end{align*}
where $\alpha>0$ is chosen to satisfy $    \frac{1}{2}(\Delta\theta^\star)^\top 
   \F_\Rdist \Delta\theta^\star = \rho.$

The remaining boundary and singular Fisher cases are characterized in Appendix~\ref{app:safe_drift_opt}; a tight constraint can have zero multiplier and drift-only solution may include null space components. In the strictly positive-multiplier regime, where both constraints are active, the maximum-gain direction is proportional to the damped natural gradient of the target objective under the reference distribution. 
\end{proposition}

A complete characterization of all cases, along with proofs and an explicit formula for $\alpha$ as a function of $(\epsilon, \rho)$, is
provided in Appendices~\ref{app:safe_drift_opt} and~\ref{app:alpha_choice}.

\begin{remark}[Connection to existing data selection methods]
Proposition~\ref{prop:oracle-safe-direction} places existing data selection methods within a common framework and helps explain the conditions under which each paradigm is preferred. When $\rho \to \infty$, the drift constraint is never active and the optimal
update reduces to $\Delta\theta^\star \propto -g_\Tdist$, recovering pure gradient alignment with the target---the principle underlying LESS~\citep{xia2024less}. This connection also reveals when target-aligned selection can be insufficient:
if $g_T$ has large components along the leading eigenvectors of $\mathbf{F}_\Rdist$, which we define as \textit{entanglement}, then selecting based on gradient alignment with $g_\Tdist$ necessarily induces large off-target drift. In this regime, diversity-based selection can be preferable as the parameter update is not necessarily concentrated along $g_T$ and may avoid reference-sensitive directions. 
\end{remark}

\subsection{Algorithm: Safe-Drift Data Selection}
Let $w \in \{0,1\}^{N_C}$ denote the selection vector,
where $w_i = 1$ indicates that candidate example $i$ is selected for fine-tuning. Let $\mathbf{G}\in \mathbb{R}^{d \times N_C}$ be the matrix of candidate loss gradients at $\theta_0$. We approximate the subset-induced parameter update by one
step with fixed gradients and preconditioner:
$
\Delta\theta(w) = -\eta \mathbf{P} \mathbf{G}w/|\sum_{i}w_i|,
$
where $\mathbf{P} \in \mathbb{R}^{d \times d}$ denotes the preconditioner induced by the
optimizer and $\eta>0$ is the learning rate. 
For example, $\mathbf{P}=\I_d$ recovers stochastic gradient descent, while adaptive methods such as Adam~\citep{kingma2014adam} use a diagonal second-order approximation for local curvature. We defer a further discussion of the choice of $\mathbf{P}$ to
Appendix~\ref{app:preconditioner}.  In our setting, the goal is to approximate the optimal safe update $\Delta\theta^\star$ in Proposition~\ref{prop:oracle-safe-direction} that maximizes the target gain while satisfying the constraint on off-target drift. Accordingly, we aim to construct a subset of exactly $S$ examples by,
\begin{align}
\label{eq:selector-objective}
w^*= \arg\min_{w\in\{0,1\}^{N_C
},\;\mathbf 1^\top w = S}\|\Delta\theta(w)-\Delta\theta^\star\|_{\F_{\Rdist}+\alpha \I}^2,
\end{align}
where $
\|x\|_{\F_{\Rdist}+\alpha \I}^2 := x^\top (\F_{\Rdist}+\alpha \I)x$ for a vector $x \in \mathbb{R}^d$ with $\alpha>0$. 

To approximately solve \eqref{eq:selector-objective} in practice, we use a greedy forward selector, following the common use of greedy matching procedures in sparse approximation \citep{mallat1993matching} and gradient-based subset selection \citep{killamsetty2021grad}. We detail the full pipeline in Algorithm~\ref{alg:reduced_safe_selector} below and Algorithm \ref{alg:greedy_safe_selector} in Appendix~\ref{app:selection}.
\begin{remark}
We carry out the reconstruction within the geometry defined by the reference set, characterized as the norm $\|\cdot \|_{\F_{\Rdist}+\alpha \I}^2$ in \eqref{eq:selector-objective}, which enables us to assign higher penalties to reconstructions that have larger components along costly off-target directions. In Appendix~\ref{app:ablation:norm}, we show that this approach yields stronger safeguarding mechanisms than reconstructions based on the standard Euclidean norm.
\end{remark}

\subsection{Low-rank estimation of Fisher information and gradients}
\label{sec:real}

We turn to the problem of estimating the gradients of the candidates and target, as well as the reference Fisher information required to solve problems~\eqref{eq:oracle-primal} and \eqref{eq:selector-objective}. In a very high-dimensional parameter space, plug-in estimation in the full parameter space is computationally unstable, memory-intensive and noisy. In particular, the empirical reference Fisher is rank-deficient whenever the number of reference samples is smaller than the number of trainable parameters.
Luckily, prior works have shown that language-model updates often lie in a low-dimensional subspace or admit a low-rank update structure \citep{aghajanyan2021intrinsic,hu2022lora,zhao2024galore}. Our goal is therefore to construct a low-rank projection,
$
\widehat \U_K \in \mathbb{R}^{d\times K}$ with $K \ll d,
$
that captures the reference-sensitive directions while also retaining the dominant
task-relevant gradient structure. 

\paragraph{Raw estimation.}
Define the estimated gradient by
$\widehat g^{\mathrm{raw}}_i:=\nabla_\theta \ell(z_i;\theta_0)
$.
 The target gradient is then estimated
by the sample mean $\widehat g_{\mathrm{tar}}^{\mathrm{raw}}:=\frac{1}{N_T}\sum_{z_i\in \mathcal D_{\mathrm{tar}}}\nabla_\theta \ell(z_i;\theta_0)$.
Let \(\mathcal D_{\mathrm{ref}}=\{x_j\}_{j=1}^{N_R}\) be the held-out set of reference inputs, and let  the score vector for the reference data be denoted:
\begin{align*}
s^R_j
:=
\nabla_\theta \log p_{\theta_0}(y_j\mid x_j)\in\mathbb{R}^d,
\qquad
y_j \sim p_{\theta_0}(\cdot\mid x_j).
\end{align*}
  Let \(\mathbf{S}_R\in\mathbb{R}^{N_R\times d}\) denote the matrix of stacked score vectors. 

\paragraph{Estimating the low-rank projection $\widehat {\U}_K$.} We construct $\widehat {\U}_K$ in two stages.
\begin{description}[noitemsep, leftmargin=2em, topsep=0em]
    \item[Step 1: low-rank approximation of ${\F}_\Rdist$.] We estimate the leading
eigenspace of $\F_\Rdist$, and retain the top $K_R$ strictly positive eigenpairs, $(\widehat{\U}_{K_R}, \widehat{\mathbf{\Lambda}}_{K_R})$, from the raw empirical reference Fisher
$
\widehat{\F}_{R}^{\mathrm{raw}}
=
\frac{1}{N_R}\mathbf{S}_R^\top \mathbf{S}_R
$\footnote{In implementation, we never form the ambient \(d\times d\) matrix explicitly. Instead, we work with the smaller \(N_R\times N_R\) Gram matrix whose nonzero eigenvalues coincide with those of $\widehat{\F}_R^{\mathrm{raw}}$ (see Appendix~\ref{app:fisher_shrinkage}.)}.
\item[Step 2: low-rank approximation of target and candidate gradients on the residual subspace.] Since the target and candidate gradients may
contain informative structure outside this reference eigenspace, we project them onto its orthogonal complement and extract a residual
task basis from the projected gradients.  Let $\widetilde g_i$ denote the residualized gradients:
\begin{align}\label{eq:orthogonal_target}
\widetilde g_i
:=
(\I_{d}-\widehat \U_{K_R}\widehat \U_{K_R}^\top)\widehat g_{i}^{\mathrm{raw}}, \qquad \forall i\in[N_C+N_T].
\end{align}
Stacking the residualized gradients gives
$
\widetilde{\mathbf{G}}_{TC}
\in \mathbb{R}^{d\times (N_T+N_C)}
$.
We then estimate an additional residual basis by retaining the top
$K_{T}$ left singular vectors corresponding to top positive singular values of $\widetilde{\mathbf{G}}_{TC}$ , yielding,
$
\widehat \U_{K_T} \in \mathbb{R}^{d\times K_{T}}.
$

\end{description}
Stacking the two bases yields the final working basis $\widehat {\U}_K$, which by construction spans both
the reference-sensitive and residual task-relevant directions:
\begin{equation}
    \label{eq:common_basis}
    \widehat {\U}_K
    := \begin{bmatrix} \widehat{\U}_{K_R} & \widehat{\U}_{K_T} \end{bmatrix}
    \in \mathbb{R}^{d \times K},
    \qquad K := K_R + K_{T},
    \qquad \widehat {\U}_K^\top \widehat {\U}_K = \I_K.
\end{equation}

We evaluate the effect of the approximation of $\mathbf{F}_{\Rdist}$ and the gradients under this low rank projection in Appendices~\ref{app:exp_linear} and ~\ref{app:spectral}. As expected, the suitability of the approximation depends on the spectrum of the eigenvalues of $\F_{\Rdist}$ and the residual target gradient, and the projection error of $g_{\Tdist}$. The selection of $K$ is discussed in Appendix \ref{app:working_basis} and ~\ref{app:K_ablation}. 

\begin{figure}[ht]
    \centering
    \begin{subfigure}[t]{0.49\textwidth}
        \centering
    \includegraphics[width=0.85\linewidth]{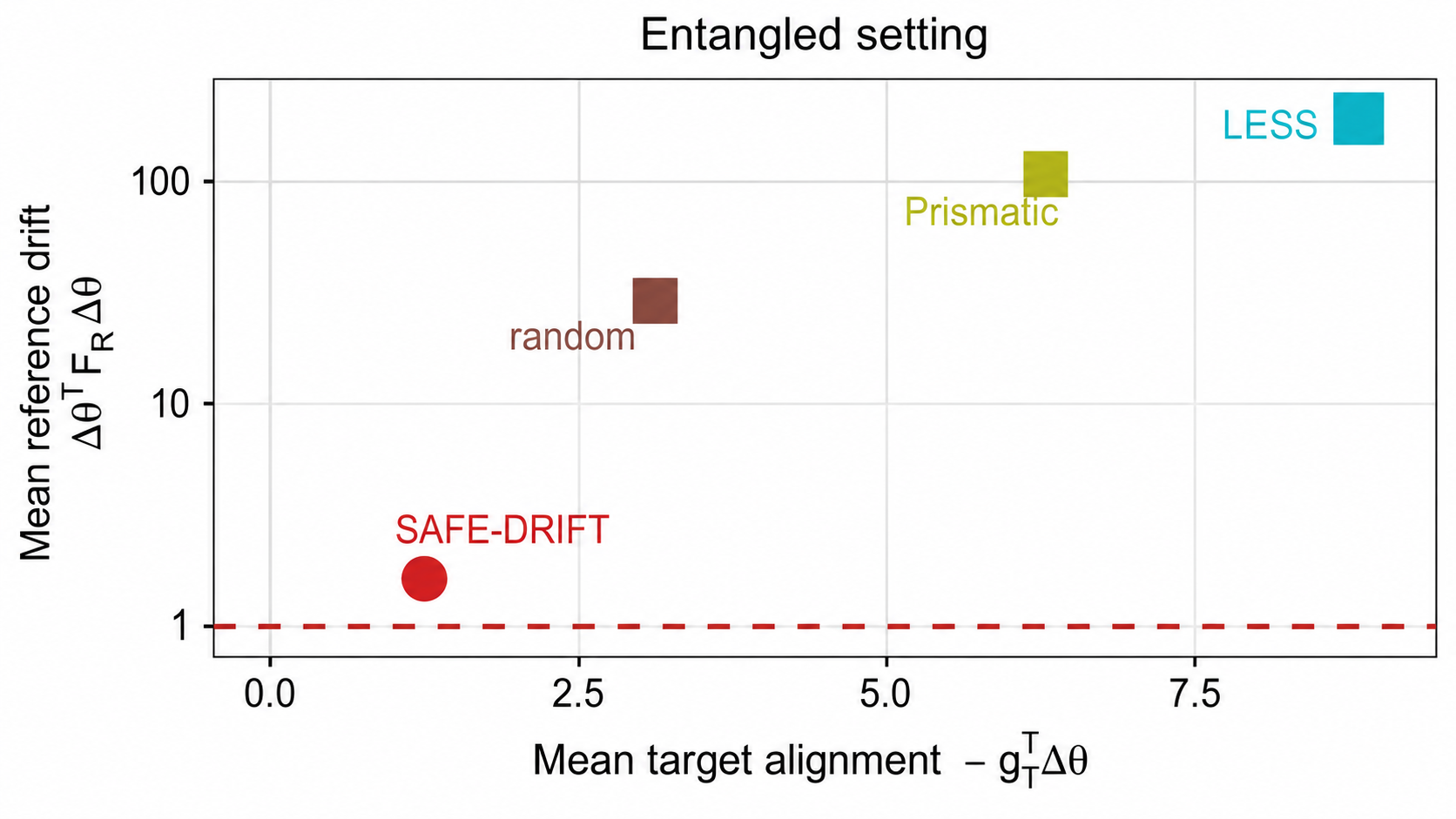}
    \end{subfigure}
        \centering
        \begin{subfigure}[t]{0.49\textwidth}
    \includegraphics[width=0.85\linewidth]{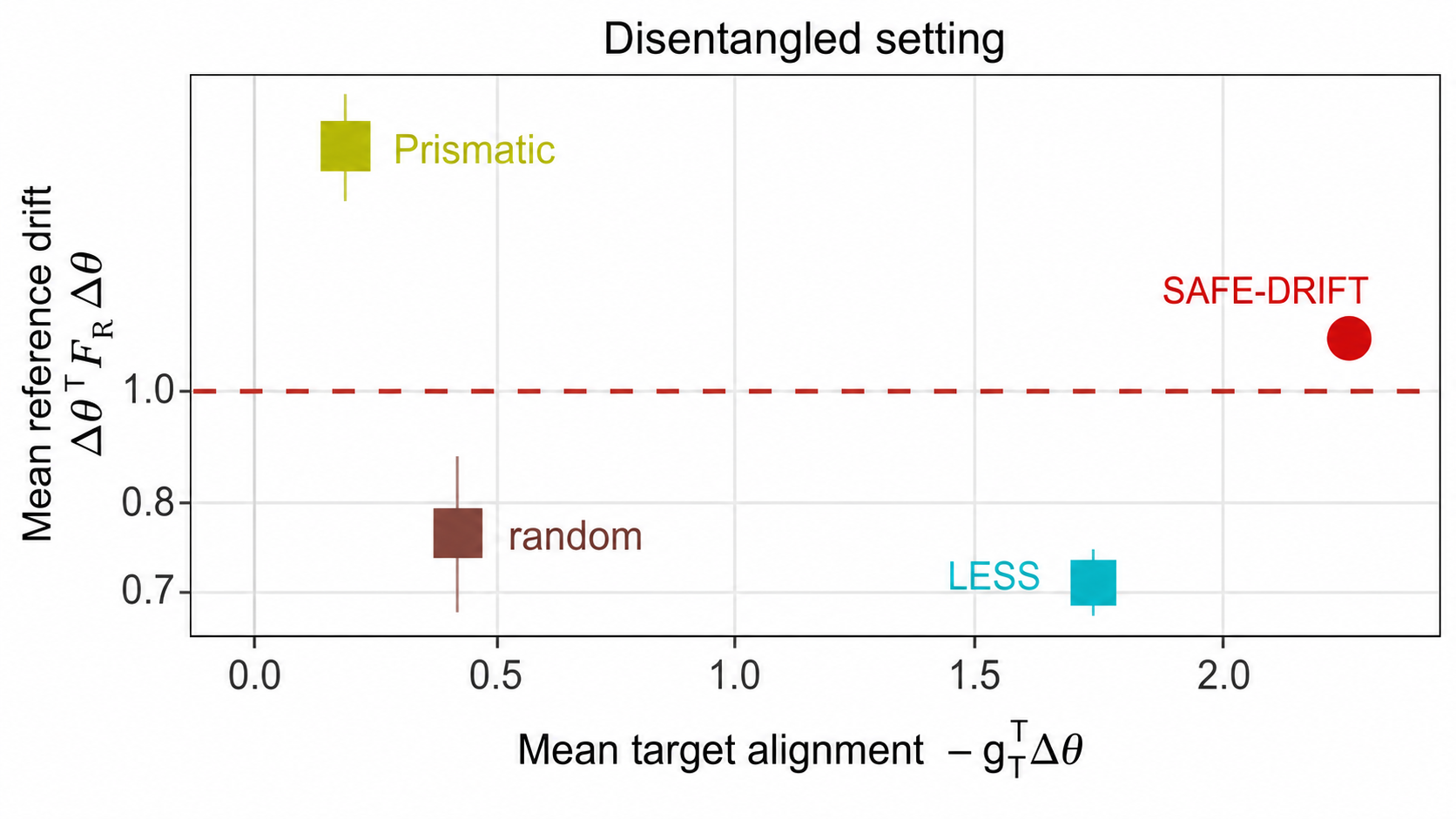}
    \end{subfigure}
\caption{\it{Trade-off between target gradient alignment and off-target drift for an ``entangled regime" and a ``disentangled regime" in the synthetic linear experiment (Appendix~\ref{app:exp_linear}). The results are averaged over 100 runs. The red line denotes the budget $\rho$. \safe{} is the only method able to control the drift.}}
\label{fig:lin-exp-data-selection-tradeoffs}
\end{figure}

\paragraph{Entanglement diagnostics.} 
Proposition~\ref{prop:oracle-safe-direction} shows that how strongly the drift budget $\rho$ affects the update depends on the target--reference \emph{entanglement}. With the low-rank estimation described above, entanglement can be characterized as the overlap between the target-relevant eigenspace and the reference-sensitive eigenspace. We first illustrate the effect of entanglement in a simple synthetic experiment where we use a linear-Gaussian model with known exact per-example gradients and exact $\mathbf{F}_{\Rdist}=\sigma^{-2}\mathbf{\Sigma}_{\Rdist}$ (Appendix~\ref{app:exp_linear}), where $\mathbf{\Sigma}_{\Rdist}$ has one expensive direction $u_1$ that costs $25\times$ more drift than $u_2$. The two regimes differ only in the target: in the \emph{entangled} regime $-g_{\Tdist}$ lies mostly along $u_1$, and in the \emph{disentangled} regime $-g_{\Tdist}=u_2$. In the entangled regime (Figure~\ref{fig:lin-exp-data-selection-tradeoffs}a), other baseline selectors incur drift cost, whereas \safe{} achieves the lowest mean plotted drift at the cost of some target alignment. In the disentangled regime (Figure~\ref{fig:lin-exp-data-selection-tradeoffs}b) the budget does not bind and \safe{} matches the target-aligned baseline LESS~\citep{xia2024less}.

On real data, we measure entanglement using the working basis $\widehat \U_K$: we report the proportion of candidates with positive damped target alignment in both reference and residual blocks, denoted as $\pi_{Q_{\mathrm{ent}}}$, while we use $\pi_{Q_{\mathrm{clean}}}$ to indicate the proportion of candidates with positive residual alignment and non-positive reference-block alignment. We provide their definitions and discuss their interpretation in Appendix~\ref{appsec:entanglement_analysis}.

\subsection[Safe update and selection in practice]{Safe update $\widehat \Delta\theta_K^\star$ and selection $\hat w$ in practice.}
The target and candidate coordinates in the common basis can be calculated by,
\begin{align}\label{eq:tar_cand_in_common_basis}
 \widehat b_{i}^{(K)}:= \widehat\U_K^\top \widehat g^{\mathrm{raw}}_i  
     \text{ for all }z_i\in \mathcal D_{\mathrm{train}}\bigcup \mathcal D_{\mathrm{tar}}, \qquad \widehat b_{\mathrm{tar}}^{(K)} := \frac{1}{N_T}\sum_{i:z_i\in\mathcal{D}_{\mathrm{tar}}} \widehat b_i^{(K)} .
\end{align}
Given $\alpha>0$ and the scaling factor $\widehat \mu_K\geq 1$ that satisfy the reference and norm budgets in \eqref{eq:oracle-primal}, which we detail in Appendix~\ref{app:working_basis}, the practical safe update $\Delta\theta^*$ using the low rank basis is, 
\begin{align}\label{eq:real_updated_direction_tar}
\widehat{\Delta}\theta^{\star}_K(\alpha)
=
-\frac{1}{\widehat \mu_K}\, \left(\diag(\widehat{\mathbf{\Lambda}}_{K_R}, \mathbf{0}_{K_T})+\alpha \I_K \right)^{-1} \widehat\U_K^\top \widehat g_{\mathrm{tar}}^{\mathrm{raw}}\in\mathbb{R}^K.
\end{align}
Define the reduced preconditioner
$
\mathbf{P}_K:=  \widehat\U_K^\top \mathbf{P} \widehat\U_K \in\mathbb{R}^{K\times K}
$, where $\mathbf{P}$ is the fixed preconditioner. Similarly, for the selection vector $w\in\{0,1\}^{N_C}$ with $\mathbf 1^\top w = S$, the projected-gradient surrogate update is,
\begin{align}
\label{eq:adam_updated_direction_cand} 
\widehat \Delta\theta_K(w)
=
-\frac{\eta}{S} \mathbf{P}_K\mathbf{B}_K w\in\mathbb{R}^{K},
\qquad
\mathbf{B}_K :=
\begin{bmatrix}
\widehat b^{(K)}_1 & \cdots & \widehat b^{(K)}_{N_C}
\end{bmatrix}.
\end{align}
We can now plug in $\widehat{\Delta}\theta^{\star}_K(\alpha)$ and $\widehat \Delta\theta_K(w)$ in \eqref{eq:selector-objective} to solve for the final selection $\hat w$. In practice, we turn the norm objective to a score for each candidate and select the top $S$ points using the forward-greedy scores as detailed in Algorithm~\ref{alg:reduced_safe_selector} and Appendix~\ref{app:working_basis}.

\begin{algorithm}[H]
\caption{\safe{} data selection in the low-rank space}
\label{alg:reduced_safe_selector}
\begin{algorithmic}[1]
\Statex \textbf{Input:}
Damping parameter $\alpha>0$, target update \(\widehat{\Delta}\theta_K^\star(\alpha)\),
preconditioner \(\mathbf P_K\),
candidate gradients $\mathbf{B}_K$,
learning rate $\eta$,
subset budget \(1\leq S\leq N_C\), shortlist size \(S\leq m \leq N_C\).

\State Form the reduced reference Fisher $
\mathbf{M}_{\alpha,K}=\diag(\widehat{\mathbf{\Lambda}}_{K_R},\mathbf 0_{K_T})+\alpha \I_K$.

\State Whiten the reduced objective via \(\mathbf{M}_{\alpha,K}^{1/2}\), and set $$
\widetilde{\mathbf{\Delta}}_K \leftarrow -\frac{\eta}{S}\mathbf{M}_{\alpha,K}^{1/2} \mathbf{P}_K\mathbf{B}_K,\qquad \widetilde{\Delta}\theta_K^\star \leftarrow\mathbf{M}_{\alpha,K}^{1/2} \widehat{\Delta}\theta_K^\star(\alpha) .
$$

\State Compute singleton scores for each sample in $\mathcal{D}_{train}$
\[
s_i
=
\langle  \widetilde{\Delta}_K(i) ,\widetilde{\Delta}\theta_K^\star\rangle
-\frac12\|\widetilde{\Delta}_K(i)\|_2^2,
\qquad i=1,\dots,N_C.
\]
\State Choose the top $S$ candidates using the forward greedy selector (Algorithm~\ref{alg:greedy_safe_selector} in Appendix~\ref{app:selection}).

\Statex \textbf{Output:} Selection $\widehat w$, selected subset
$
\widehat{\mathcal S}:=\{i\in[N_C]:\widehat w_i=1\},
$
and the update $\widehat{\Delta}\theta_K(\widehat w)$.
\end{algorithmic}
\end{algorithm}

\section{Experiments}
\label{sec:experiments}
We compare \safe{} against four established data selection baselines. These include (1) random sampling; (2) LESS~\citep{xia2024less}, a target-alignment method which leverages LoRA-based low-dimensional gradient similarity to select data; (3) DSIR~\citep{xie2023dataselectionlanguagemodels}, a target-alignment method based on hashed $n$-gram features;  (4) Prismatic Synthesis~\citep{jung2025prismatic}, a diversity-based method that clusters candidate gradients and samples from underrepresented clusters.\footnote{Although introduced for synthetic data generation, we adapt it for data selection, motivated by its use of gradient diversity to capture meaningful variation across data instances.} None of these explicitly controls off-target drift. We additionally compare against two methods that preserve reference behavior: (a) replay, the standard method for catastrophic forgetting \cite{luo2025empiricalstudycatastrophicforgetting}, and KL-regularized fine-tuning, which penalizes divergence from the base model on reference data.

All experiments use open-weight OLMo-2-7B models (OLMo-2-1124-7B or OLMo-2-1124-7B-Instruct) \citep{olmo20252olmo2furious} as the base model with parameter ${\theta_0}$; the robustness study of Section~\ref{sec:main} additionally evaluates two model families across five scales. The full experimental protocol is summarized in Table~\ref{tab:datasets}. We have additional results about off-target drifts across the settings in Table \ref{tab:rho_correlations}.

\begin{table}[t]
\centering
\resizebox{\textwidth}{!}{%
\begin{tabular}{cccccccccc}
\toprule
\textbf{Setting} &
  \textbf{\begin{tabular}[c]{@{}c@{}}Candidate\\ ($N_C$)\end{tabular}} &
  \textbf{\begin{tabular}[c]{@{}c@{}}Target\\ ($N_T$)\end{tabular}} &
  \textbf{\begin{tabular}[c]{@{}c@{}}Reference\\ ($N_R$)\end{tabular}} &
  \textbf{\begin{tabular}[c]{@{}c@{}}Base\\ Model\end{tabular}} &
  \textbf{\begin{tabular}[c]{@{}c@{}}LoRA\\ rank\end{tabular}} &
  \textbf{\begin{tabular}[c]{@{}c@{}}LoRA\\ alpha\end{tabular}} &
  \textbf{\begin{tabular}[c]{@{}c@{}}Rank\\ $K_R+K_T$\end{tabular}} &
  $\boldsymbol{(\rho,\, \epsilon)}$ &
  \textbf{\begin{tabular}[c]{@{}c@{}}Entanglement\\ $(\pi_{Q_{\mathrm{clean}}},\, \pi_{Q_{\mathrm{ent}}})$\end{tabular}} \\
\midrule
\textbf{1} &
  \begin{tabular}[c]{@{}c@{}}CodeAlpaca\\ (20K) \citep{codealpaca}\end{tabular} &
  \begin{tabular}[c]{@{}c@{}}HumanEval\\ (164) \citep{chen2021codex}\end{tabular} &
  \begin{tabular}[c]{@{}c@{}}StereoSet\\ (4.2K) \citep{nadeem2020stereoset}\end{tabular} &
  Base & 16 & 16 & $512 + 128$ & $(1\times10^{-5},\, 1)$ & $(0.16,\, 0.31)$ \\
\midrule
\textbf{2} &
  \begin{tabular}[c]{@{}c@{}}MedMCQA\\ (2K) \citep{pmlr-v174-pal22a}\end{tabular} &
  \begin{tabular}[c]{@{}c@{}}MedQA\\ (256) \citep{jin2021disease}\end{tabular} &
  \begin{tabular}[c]{@{}c@{}}BBQ (253) \citep{parrish-etal-2022-bbq}\\ TruthfulQA (256) \citep{lin2022truthfulqameasuringmodelsmimic}\end{tabular} &
  \begin{tabular}[c]{@{}c@{}}Base-\\ Instruct\end{tabular} & 16 & 32 & $64 + 16$ & $(1,\, 1)$ & $(0.22,\, 0.77)$ \\
\midrule
\textbf{3} &
  \begin{tabular}[c]{@{}c@{}}Empathetic Dialogues\\ (5K) \citep{Rashkin2018TowardsEO}\end{tabular} &
  \begin{tabular}[c]{@{}c@{}}ESConv\\ (100) \citep{liu-etal-2021-towards}\end{tabular} &
  \begin{tabular}[c]{@{}c@{}}GSM8K\\ (256) \citep{cobbe2021gsm8k}\end{tabular} &
  \begin{tabular}[c]{@{}c@{}}Base-\\ Instruct\end{tabular} & 8 & 16 & $64 + 16$ & $(0.01,\, 1)$ & $(0.36,\, 0.08)$ \\
\midrule
\textbf{4} &
  \begin{tabular}[c]{@{}c@{}}FLAN\\ (100K) \citep{wang2023far}\end{tabular} &
  \begin{tabular}[c]{@{}c@{}}IFEval\\ (42) \citep{zhou2023instructionfollowingevaluationlargelanguage}\end{tabular} &
  \begin{tabular}[c]{@{}c@{}}MMLU\\ (285) \citep{hendryckstest2021}\end{tabular} &
  Base & 32 & 128 & $128 + 32$ & $(2\times10^{-5},\, 2)$ & $(0.22,\, 0.26)$ \\
\bottomrule
\end{tabular}}
\caption{\textit{Summary of experimental settings. $N_C$: candidate pool size. $N_T$: target set size (used for $\widehat{g}_{\mathrm{tar}}$, validation, and test). $N_R$: reference set size (used for Fisher estimation and off-target drift evaluation). Sizes here describe the source pools; estimator denominators in Section~\ref{sec:real} use the corresponding estimation splits. Setting-specific evaluation sets are detailed in Appendix~\ref{appsec:real_exp_app}. The definitions of $\pi_{Q_\mathrm{clean}}$ and  $\pi_{Q_\mathrm{ent}}$ are defined and illustrated in Appendix \ref{appsec:entanglement_analysis}. All runs use learning rate $\eta = 2\times10^{-4}$.}}
\label{tab:datasets}
\end{table}

\paragraph{Measuring off-target drift.} There are three ways we measure off-target drifts. We write $\Dfish = \Delta\theta^\top \hat {\mathbf{F}}_{R}^{(K_R)} \Delta\theta$ for the \emph{predicted} drift, the reference Fisher cost the selection procedure optimizes; $\Dkl$ for the \emph{realized} drift, the token-weighted KL divergence between the base and fine-tuned model on held-out reference data; and $\Dtask$ for the \emph{behavioral} drift, the change in accuracy on the reference benchmarks. \safe{} controls $\Dfish$ by construction, and as the drift budget changes, $\Dfish$ tracks the realized drift $\Dkl$ closely (Spearman correlation $0.967$ and $0.670$ in Settings 2 and 3; Table~\ref{tab:rho_correlations}). Unless stated otherwise, results report $\Dtask$ and $\Dkl$; we invoke $\Dfish$ only where the claim concerns the mechanism.

\subsection{Main Results}\label{sec:main}
The settings in Table~\ref{tab:datasets} fall into two regimes, depending on whether fine-tuning on the candidate pool degrades or improves model behavior on the reference task. Settings 1, 2, and 4 fall in the first, where target gains come at the cost of off-target drift. Setting 3 falls in the second: its entanglement score $\pi_{Q_{\mathrm{ent}}} = 0.08$ (Table~\ref{tab:datasets}) is far below that of the other settings, so target-improving candidates lie largely outside the reference-sensitive directions and the target can be improved at zero off-target cost. Table~\ref{tab:main_results} reports the target performance and the off-target behavioral drift $\Dtask$ across all settings.
\begin{table}[t]
\centering
\resizebox{\textwidth}{!}{%
\begin{tabular}{llccccc}
\toprule
Setting & Metric & Random & DSIR & LESS & Prismatic & \textsc{Safe-Drift} \\
\midrule
 & \cellcolor[HTML]{EFEFEF}$\Delta$ HumanEval pass@1
 & \cellcolor[HTML]{EFEFEF}$\mathbf{0.136 \pm 0.005}$
 & \cellcolor[HTML]{EFEFEF}$0.122 \pm 0.015$
 & \cellcolor[HTML]{EFEFEF}$0.116 \pm 0.003$
 & \cellcolor[HTML]{EFEFEF}$0.118 \pm 0.016$
 & \cellcolor[HTML]{EFEFEF}$\underline{0.132 \pm 0.004}$ \\
 & \cellcolor[HTML]{EFEFEF}$\Delta$ HumanEval pass@10
 & \cellcolor[HTML]{EFEFEF}$\mathbf{0.207 \pm 0.023}$
 & \cellcolor[HTML]{EFEFEF}$0.150 \pm 0.010$
 & \cellcolor[HTML]{EFEFEF}$0.162 \pm 0.017$
 & \cellcolor[HTML]{EFEFEF}$0.175 \pm 0.026$
 & \cellcolor[HTML]{EFEFEF}$\underline{0.183 \pm 0.023}$ \\
\cmidrule{2-7}
\multirow{-3}{*}{1 \ (15\%)} & $\Delta$ StereoSet score
 & $\underline{0.537 \pm 0.519}$ & $1.162 \pm 0.455$ & $0.999 \pm 0.706$
 & $1.012 \pm 0.745$ & $\mathbf{0.148 \pm 0.662}$ \\
\midrule
 & \cellcolor[HTML]{EFEFEF}$\Delta$ MedQA acc.
 & \cellcolor[HTML]{EFEFEF}$\underline{0.033 \pm 0.004}$
 & \cellcolor[HTML]{EFEFEF}$\mathbf{0.039 \pm 0.016}$
 & \cellcolor[HTML]{EFEFEF}$0.021 \pm 0.003$
 & \cellcolor[HTML]{EFEFEF}$0.027 \pm 0.013$
 & \cellcolor[HTML]{EFEFEF}$0.015 \pm 0.006$ \\
\cmidrule{2-7}
 & $\Delta$ TruthfulQA score
 & $-0.069 \pm 0.023$ & $-0.083 \pm 0.019$ & $\underline{-0.062 \pm 0.005}$
 & $-0.080 \pm 0.022$ & $\mathbf{-0.038 \pm 0.010}$ \\
\multirow{-3}{*}{2 \ (25\%)} & $\Delta$ BBQ acc.
 & $-0.026 \pm 0.036$ & $-0.004 \pm 0.011$ & $\mathbf{0.014 \pm 0.042}$
 & $-0.071 \pm 0.000$ & $\underline{0.007 \pm 0.033}$ \\
\midrule
 & \cellcolor[HTML]{EFEFEF}$\Delta$ ESConv
 & \cellcolor[HTML]{EFEFEF}$\underline{0.047 \pm 0.080}$
 & \cellcolor[HTML]{EFEFEF}$0.039 \pm 0.007$
 & \cellcolor[HTML]{EFEFEF}$\mathbf{0.053 \pm 0.007}$
 & \cellcolor[HTML]{EFEFEF}$0.037 \pm 0.008$
 & \cellcolor[HTML]{EFEFEF}$0.039 \pm 0.006$ \\
\cmidrule{2-7}
 & $\Delta$ GSM8K acc.
 & $0.064 \pm 0.007$ & $0.065 \pm 0.004$ & $\underline{0.072 \pm 0.008}$
 & $0.069 \pm 0.015$ & $\mathbf{0.084 \pm 0.004}$ \\
\multirow{-3}{*}{3 \ (10\%)} & $\Delta$ MMLU acc.
 & $0.030 \pm 0.007$ & $0.025 \pm 0.008$ & $\underline{0.039 \pm 0.004}$
 & $0.029 \pm 0.006$ & $\mathbf{0.044 \pm 0.010}$ \\
\midrule
 & \cellcolor[HTML]{EFEFEF}$\Delta$ IFEval prompt acc.
 & \cellcolor[HTML]{EFEFEF}$\underline{3.530 \pm 2.435}$
 & \cellcolor[HTML]{EFEFEF}$-0.407 \pm 0.665$
 & \cellcolor[HTML]{EFEFEF}$2.172 \pm 1.545$
 & \cellcolor[HTML]{EFEFEF}$3.259 \pm 1.090$
 & \cellcolor[HTML]{EFEFEF}$\mathbf{4.141 \pm 2.145}$ \\
 & \cellcolor[HTML]{EFEFEF}$\Delta$ IFEval inst. acc.
 & \cellcolor[HTML]{EFEFEF}$\underline{3.728 \pm 1.886}$
 & \cellcolor[HTML]{EFEFEF}$0.877 \pm 1.236$
 & \cellcolor[HTML]{EFEFEF}$2.632 \pm 1.162$
 & \cellcolor[HTML]{EFEFEF}$3.158 \pm 1.058$
 & \cellcolor[HTML]{EFEFEF}$\mathbf{4.649 \pm 1.911}$ \\
\cmidrule{2-7}
\multirow{-3}{*}{4 \ (5\%)} & $\Delta$ MMLU acc.
 & $-0.833 \pm 1.434$ & $-0.667 \pm 0.850$ & $\underline{0.500 \pm 0.816}$
 & $-0.500 \pm 2.828$ & $\mathbf{0.500 \pm 0.408}$ \\
\bottomrule
\end{tabular}%
}
\caption{\textit{Target and reference performance changes across all settings. All values are fine-tuned minus base; higher is better for task scores, whereas smaller absolute StereoSet change means closer preservation. Reference-score changes therefore have the opposite sign to degradation $\Dtask$ for higher-is-better metrics. For each setting, target metrics (shaded) are separated by a horizontal rule from reference metrics (details in Appendix~\ref{appsec:experiments}). Bold denotes the best result per row; underline the second best. Scores are means $\pm$ standard deviation over three random seeds.}}
\label{tab:main_results}
\end{table}

\paragraph{Case 1: fine-tuning degrades reference behavior (Settings 1, 2, and 4).}
In Setting 1, fine-tuning on CodeAlpaca improves Python programming ability but increases stereotype bias: all baselines increase the StereoSet score relative to the base model, where a smaller change indicates better preservation of the base model's stereotype bias. \safe{} achieves competitive pass@1 and pass@10 rates on HumanEval while producing the smallest $\Delta$StereoSet score among all methods. 

Setting 2 illustrates the trade-off more sharply. Although fine-tuning improves MedQA performance, reference metrics
deteriorate for most baselines: TruthfulQA drops for all baselines, and BBQ drops for Random and Prismatic. \safe{} sacrifices some MedQA accuracy in exchange for substantially smaller losses on off-target drift. The high entanglement score $\pi_{Q_{\mathrm{ent}}} = 0.77$ indicates that most target-improving candidates are entangled with reference-sensitive
directions, explaining why controlling drift necessarily constrains target performance in this setting.

Setting 4 presents a complementary picture. With $\pi_{Q_{\mathrm{ent}}}
= 0.26$ and $\pi_{Q_{\mathrm{clean}}} = 0.22$, a larger share of candidates carry target signal in directions decoupled from the reference geometry. \safe{} achieves the best performance on both target metrics and improves MMLU accuracy, while all baselines except LESS degrade it, confirming that clean task-relevant signal allows the method to improve the target without incurring significant off-target drift.

\paragraph{Case 2: fine-tuning improves reference behavior (Setting 3).}
Setting 3 has by far the lowest entanglement of the four, $\pi_{Q_{\mathrm{ent}}} = 0.08$ (Table~\ref{tab:datasets}): almost no candidate carries its target signal through reference-sensitive directions, so \safe{} can place greater emphasis on target alignment. The reference metrics improve---emotional support dialogue draws on broad world knowledge and social reasoning, the same general capabilities probed by MMLU and GSM8K, so gains transfer rather than trade-off. \safe{} matches DSIR and Prismatic on the target task while achieving the largest gains on both GSM8K and MMLU. Accordingly, selection is driven almost entirely by target alignment and the selected set includes points whose residual Fisher cost is not small.

\begin{figure}[t]
    \centering
    \includegraphics[width=\linewidth]{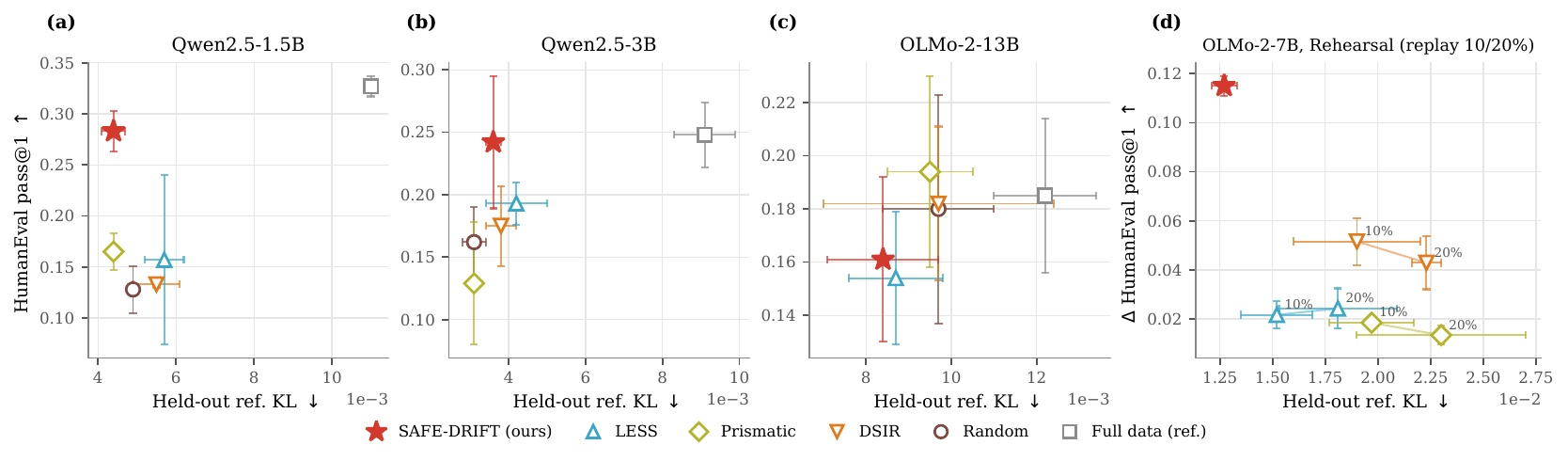}
    \caption{\textit{Realized drift $\Dkl$ vs.\ target gain (Setting~1). \textbf{(a--c)} Ablation across Qwen2.5 and OLMo-2 base models. \textbf{(d)} Comparison against rehearsal baselines on OLMo-2-7B: each baseline is fine-tuned with 10\% or 20\% of replayed reference data, whereas \textsc{Safe-Drift} uses no replay. We plot the mean and standard deviation across three seeds; upper left is better on the Pareto plots.}}
\label{fig:costgain_kl}
\vspace{-3mm}
\end{figure}

\subsection{Robustness and ablation results}
\paragraph{\safe{} outperforms behavior-preservation methods without training on reference data. } We compare against rehearsal, which mixes held-out reference examples into training at 10\% and 20\% of the dataset, to the baselines in Setting 1 (Figure~\ref{fig:costgain_kl}d). To the best our knowledge, \safe{} is the first to formulate drift control from a reference Fisher perspective and to use it to select data under an explicit budget. \safe{} attains the largest target gain compared to any rehearsal-based baselines, with the lowest realized drift, and without ever training on the reference set. We also augment LESS and Prismatic with the reference KL penalty in the SFT loss. The penalty reduces drift as intended, but generally at the cost of target accuracy, so the regularized baselines and \safe{} occupy different points on the same trade-off. 
Full numbers are in Appendix~\ref{app:preservation}.

\begin{figure}[t]
    \centering
    \includegraphics[width=\linewidth]{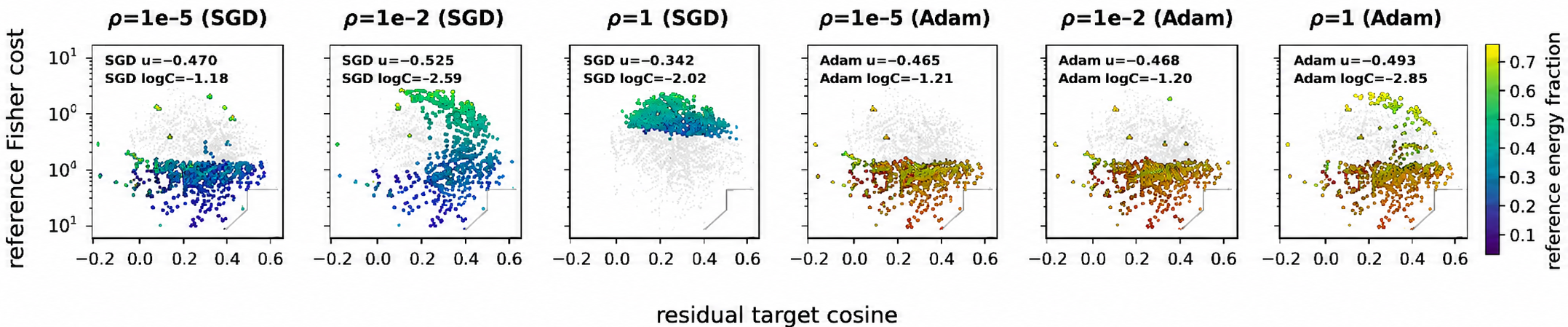}
    \caption{\textit{Effect of the drift budget $\rho$ on target accuracy and off-target drift under SGD and Adam preconditioners. Colored markers denote the examples selected by \textsc{Safe-Drift} in Setting~2. }}\label{fig:ablationrho_experiment_2} 
\end{figure}

\paragraph{The drift budget $\rho$ trades target gain against drift, and its cost depends on the gradient update method and entanglement.} We sweep $\rho$ over $\{10^{-5},10^{-2},10^{0}\}$ under the SGD and Adam preconditioners in Setting 2 (Figure~\ref{fig:ablationrho_experiment_2}). Under the SGD preconditioner ($\mathbf{P}=\I$), increasing $\rho$ moves the selected set from mostly low-cost candidates toward a high-cost, high-entanglement region, consistent with $\pi_{Q_{\mathrm{ent}}}=0.77$; under Adam, the selected set is nearly unchanged between $\rho=10^{-5}$ and $10^{-2}$ and enters that region only at $\rho=1$, so the preconditioner keeps the effective update in a low-cost region for longer. Across a larger sweep (Table~\ref{tab:rho_correlations}), larger $\rho$ consistently improves target accuracy (Spearman $0.753$ and $0.672$ in Settings 2 and 3), but its association with reference drift reverses sign: $0.483$ in the entangled Setting 2 against $-0.735$ in Setting 3, where improving the target also improves reference behavior. Throughout the sweep, the predicted drift $\Dfish$, which \safe{} controls, tracks the realized drift $\Dkl$ (Spearman $0.967$ and $0.670$ in Settings 2 and 3; Table~\ref{tab:rho_correlations}). In practice, we selecting $\rho$ on a validation drift set (Appendix~\ref{app:rho}).

\begin{figure}[t]
    \centering
    \includegraphics[width=0.9\linewidth]{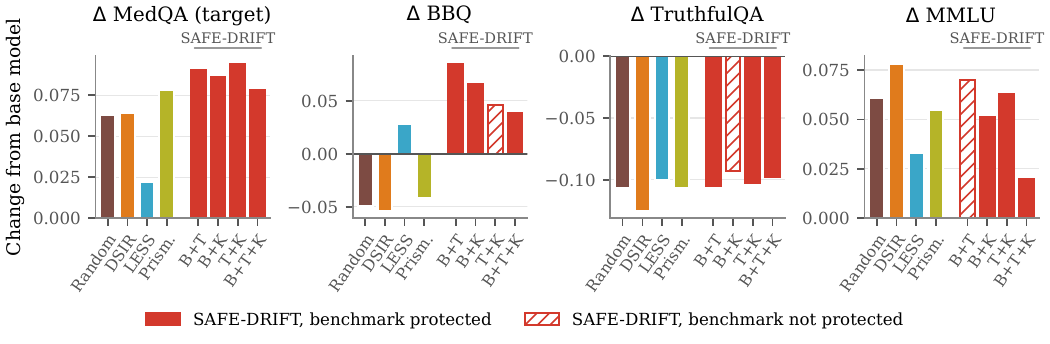}
    \caption{\textit{Composite reference distributions (Setting~2). Each panel reports the change from the base model on the target (MedQA) and on each reference benchmark ($\mathrm{B}$ = BBQ, $\mathrm{T}$ = TruthfulQA, $\mathrm{K}$ = MMLU). Results of \safe{} are shown for each reference composition. Full results, including held-out KL, are in Table~\ref{tab:offtarget_mixtures}.}}
    \label{fig:offtarget_mixtures}
\end{figure}

\paragraph{The drift control generalizes across models, mixtures of references, and full parameter fine-tuning. }
We repeat Setting 1 across Qwen2.5 at 1.5B/3B/7B and OLMo-2 at 7B/13B, at a 15\% budget with three seeds (Figure~\ref{fig:costgain_kl}a--c; full table in Appendix~\ref{app:model_ablation}). \safe{} attains the lowest predicted drift $\Dfish$ in every configuration, reducing $\Dkl$ by 31.1--60\% relative to full data at competitive target gain. Its realized drift $\Dkl$ is lowest at Qwen2.5-1.5B and OLMo-2-13B and its target gain is largest on the smaller Qwen models.  When the reference combines several datasets, \safe{} matches or exceeds the best baseline on the target and keeps BBQ above the base model, even when BBQ is not protected; on TruthfulQA and MMLU, it performs comparably to the baselines (Figure~\ref{fig:offtarget_mixtures}). This suggests that a reference Fisher estimated from a subset of benchmarks can capture directions shared with the others (Appendix~\ref{app:mixtures}).
Also, running \safe{} with full SFT instead of LoRA trades some target accuracy for better reference preservation (Appendix~\ref{app:lora_full}). 

\paragraph{Design choices and computational cost.} Using reference-Fisher-induced norm in Equation~\eqref{eq:selector-objective} makes the data selection more responsive to the budget $\rho$; Euclidean norm objective selects more reference-sensitive gradients regardless of $\rho$ (Appendix~\ref{app:ablation:norm}). The best preconditioner
varies by setting, and we recommend matching the downstream optimizer (Appendix~\ref{app:preconditioner_ablation}). Results are stable across ranks $(K_R, K_T)$, with the largest basis giving the smallest drift (Appendix~\ref{app:K_ablation}). Finally, \safe{} shares its dominant cost---gradient construction over the candidate pool---with LESS and Prismatic, and its reported selection FLOPs match those of other gradient-based selectors at the displayed precision (Appendix~\ref{app:compute}).  

\section{Conclusion}

Data selection methods are typically evaluated on target task alone, without accounting for their impact on model behavior outside the target domain. In this paper, we explicitly formalize this trade-off between target improvement and off-target drift. Our method, \safe, enables controllable off-target drift during targeted data selection. Through systematic experiments across various settings and domains, we demonstrate the effectiveness of our method. However, our method relies on two assumptions. First, it requires a well-approximated reference distribution. If the reference set does not fully cover all the behaviors we want to preserve, such as reasoning ability or domain knowledge, the observed reference distribution may underestimate the true off-target drift. We believe a future direction is to study how reference distribution should be constructed, expanded, and adapted rather than treated as fixed inputs, for example, following related work on broad-coverage evaluation, red-teaming evaluation, and trustworthiness \citep{liang2022holistic,wang2023decodingtrust,mazeika2024harmbench,chang2025retain}. Second, our analysis is based on a one-step or local update approximation which is a reasonable assumption for supervised fine-tuning, but it might not hold for fine-tuning based on reinforcement learning, where parameter updates can grow large as the policy diverges from initialization. We encourage future work to handle the trade-off in such settings where drift accumulates across updates.  

\section*{Acknowledgments}

C.D. would like to acknowledge support for this project
from the National Science Foundation (NSF grant IIS-2238616). This work was completed in part with resources provided by the University of Chicago’s Research Computing Center.

\bibliography{references} 
\newpage
\appendix

\section{Related Work \& Notations}
\subsection{Related Works}
\paragraph{Data Selection}
Data selection aims to find a subset of training examples
that induces performance similar to training on the full
dataset.  In particular, \cite{albalak2024survey} divide data selection methods in two main classes: Distribution Matching methods, and Distribution Diversification methods.

{\it Distribution matching methods} aim to find samples that closely match the target distribution. In this setting, a crucial component is to assess the similarity of a given data point with the target data.
A plethora of methods have introduced non-model specific notions of useful data ~\citep{chen2023skillit} or $n$-gram features ~\citep{xie2023dataselectionlanguagemodels} to perform data selection for pre-training. Because these methods lack the finesse of detecting ``high-quality'' data points by solely relying on surface attributes, the general trend in the literature -- particularly in the context of fine-tuning or instruction tuning --- seems to have turned towards using model-specific metrics based on scores like perplexity ~\citep{marion2023moreinvestigatingdatapruning}, model embedding ~\citep{ivison-etal-2023-data}. Recent methods have for instance relied on formulating the data selection problem as a
multi-armed bandit, where the utility function is an aggregated reward function of the cosine similarity and
magnitude similarity between gradients of batches from the auxiliary and target datasets.
Another line of recent work relies on model gradients ~\citep{xia2024less, choe2026what} for data selection, and calculate the pointwise data influence between pair of training and validation instances.

{\it Distribution diversification methods}, by contrast, seek to emphasize diversity of instances.  These
algorithms aim to efficiently approximate various geometric extent measures over the training set using a small subset. These include coreset methods~\citep{sener2018activelearningconvolutionalneural, mirzasoleiman2020coresetsdataefficienttrainingmachine} and submodular algorithms~\citep{yehuda2022activelearningcoveringlens, bhatt-etal-2024-experimental}.
\paragraph{Catastrophic Forgetting}
Catastrophic forgetting ~\citep{wang2024comprehensive, luo2025empiricalstudycatastrophicforgetting} happens when adaptation to a new distribution generally results in a largely reduced ability to capture the old ones. Existing approaches to remedy this issue can be typically divided into two categories: data-aware and data-oblivious. Data-aware methods ~\citep{wang-etal-2023-orthogonal, ritter2018onlinestructuredlaplaceapproximations} assume access to earlier data distributions to some extent.  In contrast, data-oblivious methods operate solely with the pre-trained model, without revisiting any prior data.  Data-oblivious methods assume access to the pretrained model without revisiting any  prior data. Previous works ~\citep{panda2024lotteryticketadaptationmitigating, biderman2024loralearnsforgets}
selectively update a subset of parameters in a pretrained model to limit interference with previously knowledge.

\subsection{Notations.}
Throughout the paper, we use bold uppercase letters, such as $\mathbf{A}$, to denote matrices. For a matrix $\mathbf{A}\in\mathbb{R}^{d\times n}$, we write $\mathbf{A}(i)$ for its $i$-th column and $\mathbf{A}_{ij}$ for its $(i,j)$-th entry. For a vector $a\in\mathbb{R}^d$, we write $a_i$ for its $i$-th coordinate.
Let ${\diag}(\mathbf{A},\mathbf{B})$ denote the block-diagonal matrix
$\begin{bmatrix}\mathbf{A}&0\\0&\mathbf{B}\end{bmatrix}$ given matrices $\mathbf{A}$ and $\mathbf{B}$. Let $\mathbf 0_{K}\in\mathbb{R}^{K\times K}$ denote the zero matrix,
$\I_K\in\mathbb{R}^{K\times K}$ denote the identity matrix for an integer $K\ge0$, with empty matrices when $K=0$.

\section{Technical Appendices}
\subsection{Proof of Proposition~\ref{prop:local-kl}}
\label{app:prop_1}
\begin{proof}

Assume there exists $r_0>0$ such that the following hold in the neighborhood $\mathcal N_{r_0}(\theta_0):=\{\theta\in\mathbb R^d:\|\theta-\theta_0\|_2<r_0\}$:
\begin{enumerate}
    \item[(i)] For $\Rdist$-almost every $x$, the distributions $p_\theta(\cdot\mid x)$ have a common support $\mathcal Y_x$ throughout $\mathcal N_{r_0}(\theta_0)$ and are strictly positive on that support. For every $y\in\mathcal Y_x$, the map $\theta \mapsto \log p_\theta(y\mid x)$ is three times continuously differentiable there.
    \item[(ii)] Differentiating $\int_{\mathcal Y_x}p_\theta(y\mid x)\,dy=1$ twice under the integral is valid; for example, the first two derivatives of $p_\theta$ admit integrable dominating envelopes on this neighborhood. The first two log derivatives at $\theta_0$ and the squared score have finite joint expectations under $x\sim\Rdist$, $y\sim p_{\theta_0}(\cdot\mid x)$.
    \item[(iii)] There exists an integrable envelope $G(x,y)$ such that
    $$
    \sup_{\theta\in\mathcal N_{r_0}(\theta_0)}
    \bigl\|\nabla_\theta^3 \log p_\theta(y\mid x)\bigr\|
    \le G(x,y), \quad \E_{x\sim \Rdist}\E_{y\sim p_{\theta_0}(\cdot\mid x)}[G(x,y)] < \infty.
    $$
\end{enumerate}

For a fixed $x$, define
$$d_x(\theta)
=
\KL\!\left(
p_{\theta_0}(\cdot\mid x)
\,\|\, 
p_\theta(\cdot\mid x)
\right).$$
Then,
\[
D_\Rdist(\Delta\theta)=\E_{x\sim \Rdist}[\,d_x(\theta_0+\Delta\theta)].
\]
From the definition of KL divergence,
\begin{equation*}
    d_x(\theta)=\E_{y\sim p_{\theta_0}(\cdot\mid x)}[\log p_{\theta_0}(y\mid x)-\log p_\theta(y\mid x)]
\end{equation*}
and
\begin{equation*}
    \nabla_\theta d_x(\theta) = \E_{p_{\theta_0}(\cdot\mid x)}[-\nabla_\theta \log p_{\theta}(\cdot|x)], \qquad \nabla^2_\theta d_x(\theta) = \E_{p_{\theta_0}(\cdot\mid x)}[-\nabla^2_\theta \log p_{\theta}(\cdot|x)].
\end{equation*}

At $\theta=\theta_0$,
\begin{align*}
    \nabla_\theta d_x(\theta) |_{\theta=\theta_0}
    &= \E_{p_{\theta_0}(\cdot\mid x)}[-\nabla_\theta \log p_{\theta_0}(\cdot|x)]\\
    &= -\int_{\mathcal Y_x} p_{\theta_0}(y|x)\frac{\nabla_\theta p_{\theta_0}(y|x)}{p_{\theta_0}(y|x)}\,dy\\
    &=-\nabla_\theta \int_{\mathcal Y_x} p_{\theta_0}(y|x)\,dy\\
    &= \mathbf{0}_d
\end{align*}
and 
\begin{equation*}
    \nabla^2_\theta d_x(\theta)|_{\theta=\theta_0}=\E_{p_{\theta_0}(\cdot\mid x)}[-\nabla^2_\theta \log p_{\theta_0}(\cdot|x)].
\end{equation*}
By a second-order Taylor expansion around $\theta=\theta_0$,
\begin{align*}
    d_x(\theta)
    &=d_x(\theta_0)+\nabla_\theta d_x(\theta)|_{\theta=\theta_0}^{\top}(\theta-\theta_0) + \frac{1}{2}(\theta-\theta_0)^{\top}\nabla^2_\theta d_x(\theta)|_{\theta=\theta_0}(\theta-\theta_0)+ R_x(\Delta\theta)\\
    &=\frac{1}{2}\Delta \theta^{\top}\E_{p_{\theta_0}(\cdot\mid x)}[-\nabla^2_\theta \log p_{\theta_0}(\cdot|x)]\Delta \theta+ R_x(\Delta\theta),
\end{align*}
where $R_x(\Delta\theta)
$ is Taylor remainder.
The common-support and differentiation assumptions give the Fisher identity \citep{kunstner2019limitations},
\begin{equation*}
    \mathbf{F}_\Rdist=\E_{x\sim \Rdist}\E_{y\sim p_{\theta_0}(\cdot\mid x)}\big [-\nabla^2_\theta \log p_{\theta_0}(\cdot|x)\big],
\end{equation*}

which leads to,
\begin{equation*}
    D_\Rdist(\Delta\theta) = \frac{1}{2}\Delta\theta^{\top}\mathbf{F}_\Rdist\Delta\theta + \E_{x\sim \Rdist}[R_x(\Delta\theta)].
\end{equation*}

For $\|\Delta\theta\|_2<r_0$, Taylor's theorem applied to each log density, followed by expectation over $y$, gives
\begin{align*}
\left|R_x(\Delta\theta)\right|
&\le\frac{\|\Delta\theta\|_2^3}{6}
\E_{y\sim p_{\theta_0}(\cdot\mid x)}\!\left[
\sup_{\theta\in\mathcal N_{r_0}(\theta_0)}\bigl\|\nabla_\theta^3 \log p_\theta(y\mid x)\bigr\|\right]\\
&\le\frac{\|\Delta\theta\|_2^3}{6}\E_{y\sim p_{\theta_0}(\cdot\mid x)}[G(x,y)].
\end{align*}
Taking expectation over $x\sim \Rdist$ gives
$$\left|D_\Rdist(\Delta\theta)-\frac12 \Delta\theta^\top \F_\Rdist \Delta\theta
\right|\le\frac{\|\Delta\theta\|_2^3}{6}\,
\E_{x\sim \Rdist}\E_{y\sim p_{\theta_0}(\cdot\mid x)}[G(x,y)].
$$
Since the envelope is integrable, 
$$D_\Rdist(\Delta\theta)=\frac12 \Delta\theta^\top \F_\Rdist \Delta\theta
+o(\|\Delta\theta\|_2^2).
$$
This proves the claim.
\end{proof}


\subsection{Proof of Proposition~\ref{prop:oracle-safe-direction}}
\label{app:safe_drift_opt}
\begin{theorem}[Oracle safe update (full)]
\label{thm:oracle-safe-update-appendix}
Let $\F_\Rdist \succeq 0$ and let $g_\Tdist \neq 0$. For $\rho>0$ and
$\epsilon>0$, consider
\begin{align*}
    \Delta^\star
    \in
    \argmin_{\Delta \in \mathbb{R}^d}
    g_\Tdist^\top \Delta
    \quad \text{subject to} \quad
    \frac12 \Delta^\top \F_\Rdist \Delta \leq \rho,
    \qquad
    \|\Delta\|_2 \leq \epsilon .
\end{align*}
The optimal solutions are characterized by the following cases, checked in order. Here a constraint is \emph{tight} when it holds with equality; tightness alone does not imply a strictly positive KKT multiplier.

\begin{enumerate}
    \item \textbf{Feasible norm-only optimum.}
    If
    \[
        \frac{\epsilon^2}{2\|g_{\Tdist}\|_2^2}
        g_{\Tdist}^\top \F_{\Rdist} g_{\Tdist}
        \leq \rho ,
    \]
    then a zero drift multiplier is valid, and the unique solution is
    \[
        \Delta^\star
        =
        -\epsilon \frac{g_{\Tdist}}{\|g_{\Tdist}\|_2}.
    \]

    \item \textbf{Feasible drift-only optimum.}
    Suppose $g_{\Tdist} \in \operatorname{Range}(\F_{\Rdist})$ and define
    \[
        \Delta_R
        =
        -
        \sqrt{
            \frac{2\rho}{g_{\Tdist}^\top \F_{\Rdist}^\dagger g_{\Tdist}}
        }
        \F_{\Rdist}^\dagger g_{\Tdist} ,
    \]
    where $\F_{\Rdist}^\dagger$ denotes the Moore--Penrose pseudoinverse. If
    \[
        \|\Delta_R\|_2 \leq \epsilon ,
    \]
    then a zero norm multiplier is valid, and the complete optimal set is
    \[
        \left\{\Delta_R+v:\ v\in\operatorname{Null}(\F_{\Rdist}),\quad
        \|v\|_2^2\leq\epsilon^2-\|\Delta_R\|_2^2\right\}.
    \]
    In particular, $\Delta_R$ is the minimum-norm optimum. The norm constraint can still be tight for some of these optima.

    \item \textbf{Both multipliers strictly positive.}
    In all remaining cases, both constraints are tight and both KKT multipliers are strictly positive. There
    exists a unique damping parameter $\alpha>0$ such that
    \[
        \frac{\epsilon^2}{2}
        \frac{
            g_{\Tdist}^\top (\F_{\Rdist}+\alpha I)^{-1}
            \F_{\Rdist}
            (\F_{\Rdist}+\alpha I)^{-1}g_{\Tdist}
        }{
            g_{\Tdist}^\top (\F_{\Rdist}+\alpha I)^{-2}g_{\Tdist}
        }
        =
        \rho ,
    \]
    and the solution is
    \[
        \Delta^\star
        =
        -
        \epsilon
        \frac{
            (\F_{\Rdist}+\alpha I)^{-1}g_{\Tdist}
        }{
            \|(\F_{\Rdist}+\alpha I)^{-1}g_{\Tdist}\|_2
        } .
    \]
\end{enumerate}

In case~3, the unique solution has the KKT representation
\[
    \Delta^\star
    =
    -c\,(\F_{\Rdist}+\alpha I)^{-1}g_{\Tdist}
\]
for some $c>0$ and $\alpha>0$. The endpoint $\alpha\to\infty$ yields the normalized Euclidean direction in case~1. In case~2, the pseudoinverse gives the minimum-norm optimum with its drift-determined scale; arbitrary feasible nullspace components must be included to recover every optimum.
\end{theorem}

\begin{proof}
Rewrite~\eqref{eq:oracle-primal} as a minimization:
\begin{equation}
\label{eq:min_form}
    \min_{\Delta\theta \in \mathbb{R}^d} \quad g_{\mathcal T}^\top \Delta\theta
    \qquad
    \text{subject to}
    \qquad
    \frac{1}{2}\Delta\theta^\top \mathbf{F}_{\mathcal R} \Delta\theta \leq \rho,
    \qquad
    \|\Delta\theta\|^2_2 \leq \epsilon^2.
\end{equation}
The objective is linear and both constraints are convex since $\mathbf{F}_{\mathcal R} \succeq 0$ and $\|\cdot\|_2$ is convex. Note here that $\mathbf{F}_{\mathcal R} \succeq 0$ always holds since for any $v\in\R^d$, $v^\top \F_{\mathcal R} v=\E\!\left[(v^\top s_{\theta_0}(x,y))^2\right]\ge 0.$
Taking $\Delta\theta = 0$ gives $0 < \rho$ and $0 < \epsilon^2$, which is strictly
feasible for any $\rho, \epsilon > 0$. The feasible set is compact and nonempty, so an optimum exists. Slater's condition and strong duality hold, and the KKT conditions are necessary and sufficient for a global optimum.

Form the Lagrangian with multipliers $\sigma \geq 0$ and $\mu \geq 0$:
\[
\mathcal{L}(\Delta\theta, \sigma, \mu)
= g_{\mathcal T}^\top \Delta\theta
+ \sigma\!\left(\frac{1}{2}\Delta\theta^\top \mathbf{F}_{\mathcal R} \Delta\theta - \rho\right)
+ \mu\!\left(\|\Delta\theta\|^2_2 - \epsilon^2\right).
\]

\textbf{Stationarity.} Setting $\nabla_{\Delta\theta}\mathcal{L} = 0$:
\begin{equation}
\label{eq:stationarity2}
g_{\mathcal T} + \sigma \mathbf{F}_{\mathcal R} \Delta\theta + 2\mu \Delta\theta = 0
\implies
(\sigma \mathbf{F}_{\mathcal R} + 2\mu \I)\Delta\theta = -g_{\mathcal T}.
\end{equation}
For any $\mu > 0$, the matrix $\sigma \mathbf{F}_{\mathcal R} + 2\mu \I$ has an inverse 
regardless of whether $\mathbf{F}_{\mathcal R}$ is singular, and:
\begin{equation}
\label{eq:solution2}
\Delta\theta^\star = -(\sigma \mathbf{F}_{\mathcal R} + 2\mu \I)^{-1} g_{\mathcal T}.
\end{equation}

\textbf{Complementary slackness.}
\begin{align}
\sigma\!\left(\frac{1}{2}(\Delta\theta^\star)^\top 
    \mathbf{F}_{\mathcal R} \Delta\theta^\star - \rho\right) &= 0, 
    \label{eq:cs_fisher}\\
\mu\!\left(\|\Delta\theta^\star\|_2^2 - \epsilon^2\right) &= 0.
    \label{eq:cs_norm}
\end{align}
We consider the three non-trivial cases.

\textbf{Case 1: $\sigma = 0$, $\mu = 0$.}
Stationarity gives $g_{\mathcal T} = 0$, which is trivial. Assume 
$g_{\mathcal T} \neq 0$, so this case does not arise.

\textbf{Case 2: $\sigma > 0$, $\mu = 0$ (zero norm multiplier).}
Stationarity requires $g_{\mathcal T}\in\operatorname{Range}(\F_{\mathcal R})$ and gives
$\Delta\theta^\star=-\sigma^{-1}\F_{\mathcal R}^{\dagger}g_{\mathcal T}+v$ with $v\in\operatorname{Null}(\F_{\mathcal R})$.
A zero norm multiplier allows either a slack or a tight norm constraint. By~\eqref{eq:cs_fisher}, the quadratic drift constraint is tight:
\[
\frac{1}{2}(\Delta\theta^\star)^\top \mathbf{F}_{\mathcal R} \Delta\theta^\star = \rho.
\]
Substituting this stationarity solution and using $\F_{\mathcal R}v=0$:
\[
\frac{1}{2\sigma^2}g_{\mathcal T}^\top \mathbf{F}_{\mathcal R}^{\dagger} 
\mathbf{F}_{\mathcal R} \mathbf{F}_{\mathcal R}^{\dagger} g_{\mathcal T} 
= \frac{1}{2\sigma^2}g_{\mathcal T}^\top \mathbf{F}_{\mathcal R}^{\dagger} 
g_{\mathcal T} = \rho,
\]
giving $\sigma^\star = \sqrt{g_{\mathcal T}^\top \mathbf{F}_{\mathcal R}^{\dagger} 
g_{\mathcal T} / (2\rho)}$, and every solution in this case is:
\begin{equation}
\label{eq:case2}
   \Delta\theta^\star
= -\sqrt\frac{2\rho}{g_{\mathcal T}^\top \mathbf{F}_{\mathcal R}^{\dagger} g_{\mathcal T}}\,
\mathbf{F}_{\mathcal R}^{\dagger} g_{\mathcal T}+v,\qquad
v\in\operatorname{Null}(\F_{\mathcal R}),\quad
\|v\|_2^2\leq\epsilon^2-\|\Delta_R\|_2^2.
\end{equation}

where \(\F_{\mathcal R}^\dagger\) is the Moore--Penrose pseudoinverse. Let the decomposition of $g_{\mathcal T}$
$$g_{\mathcal T} = g_{\mathrm{range}} + g_{\mathrm{null}},
\qquad
g_{\mathrm{null}} \in \mathrm{Null}(\F_{\mathcal R}),$$

The range and nullspace components in~\eqref{eq:case2} are orthogonal, so this set is nonempty exactly when $\|\Delta_R\|_2\leq\epsilon$. Every such point satisfies KKT and is optimal; equality in the drift-only Cauchy--Schwarz bound forces this range component, so the set is complete. If $g_{\mathrm{null}}\neq0$, the problem without the norm constraint is unbounded below along $-g_{\mathrm{null}}$. The original norm-constrained problem remains bounded, and projecting stationarity onto the nullspace shows that its norm multiplier must satisfy $\mu>0$.





\textbf{Case 3: $\sigma = 0$, $\mu > 0$ (norm constraint active). }
By the slackness condition \eqref{eq:cs_norm}, 
\begin{equation*}
    \|\Delta \theta^*\|_2^2 = \frac{1}{4\mu^2}\|g_{\mathcal T}\|_2^2=\epsilon^2
\end{equation*}
leading to $\mu^*=\|g_{\mathcal T}\|_2/2\epsilon$. Thus,
\begin{equation*}
    \Delta\theta^*=-\frac{\epsilon}{\|g_{\mathcal T}\|_2}g_{\mathcal T}
\end{equation*}

This point is feasible exactly under the condition in case~1 of the theorem and is the unique minimizer over the Euclidean ball. Its drift constraint may be slack or tight, while the drift multiplier can be zero.

\textbf{Case 4: $\sigma > 0$, $\mu > 0$ (both multipliers positive).}
If neither of the preceding feasible cases applies, both multipliers must be positive. In this case the solution takes the form~\eqref{eq:solution2}, which
we write compactly:
\[
\Delta\theta^\star 
= -\frac{1}{\sigma^*}
  \left(\mathbf{F}_{\mathcal R} +\alpha I\right)^{-1}g_{\mathcal T},
\]
where $\alpha=2\mu^\star/\sigma^\star>0$. Both constraints are tight by complementary slackness, and the Lagrangian has positive-definite Hessian $\sigma^\star\F_{\mathcal R}+2\mu^\star I$, so this optimum is unique. The update is a damped natural gradient with damping determined by the ratio of the two multipliers.

Using the active norm constraint ($\|\Delta\theta^*\|_2=\epsilon$, we let $\sigma^* = \frac{\|(\F_{\mathcal R}+\alpha \I)^{-1}g_{\mathcal T}\|_2}{\epsilon}$), and $\Delta\theta^\star $ becomes
\begin{equation}
\label{eq:both_active_raw_fisher}
\Delta\theta^\star
=
-\epsilon\,
\frac{(\F_{\mathcal R}+\alpha \I)^{-1}g_{\mathcal T}}
{\|(\F_{\mathcal R}+\alpha \I)^{-1}g_{\mathcal T}\|_2},
\end{equation}

Then, $\alpha$ is chosen to satisfy \eqref{eq:cs_fisher}
\begin{equation}
\label{eq:case4}
    \frac{1}{2(\sigma^*)^2}g_{\mathcal T}^\top(\mathbf{F}_{\mathcal R} + \alpha I)^{-1}\F_{\mathcal R} (\mathbf{F}_{\mathcal R} + \alpha I)^{-1}g_{\mathcal T}=\rho
\end{equation}

The existence of positive multipliers gives a positive root of this scalar equation. In this remaining case the target gradient must have components on at least two distinct eigenvalues: for a single supported eigenvalue, either the norm-only optimum is feasible or the drift-only optimum is feasible. Proposition~\ref{prop:alpha_mono} below therefore gives uniqueness of $\alpha$.



\end{proof}

\subsection[Characterization of alpha]{Characterization of $\alpha$}
\label{app:alpha_choice}
Let $(\lambda_i,v_i)_{i=1}^d$ be a complete orthonormal eigensystem of $\mathbf F_{\mathcal R}$, including zero eigenvalues, and write $g_{\mathcal T}=\sum_i c_i v_i$. Proposition~\ref{prop:oracle-safe-direction} gives the oracle direction when both multipliers are positive.
When both multipliers are positive, $\alpha$ is a positive root of the scalar equation $\phi(\alpha)=2\rho/\epsilon^2$, where the sums include all eigenvectors, including those with $\lambda_i=0$:
\begin{align*}
\phi(\alpha)
:=
\frac{\sum_i \frac{\lambda_i c_i^2}{(\lambda_i+\alpha)^2}}
{\sum_i \frac{c_i^2}{(\lambda_i+\alpha)^2}}.
\end{align*}

\begin{proposition}\label{prop:alpha_mono}
Let $\phi$ be defined as above with $g_{\mathcal T}\neq0$. Then $\phi$ is nondecreasing on $(0,\infty)$. If there exist $i\neq j$ such that $c_i c_j\neq 0$ and $\lambda_i\neq \lambda_j$, then $\phi'(\alpha)>0$ for all $\alpha>0$; under this condition, any positive solution is unique. Otherwise, $\phi$ is constant at the single eigenvalue supported by $g_{\mathcal T}$, so a matching level has every $\alpha>0$ as a solution.
\end{proposition}

\begin{proof}
\begin{align*}
N(\alpha)
:=
\sum_i \frac{\lambda_i c_i^2}{(\lambda_i+\alpha)^2},
\qquad
D(\alpha)
:=
\sum_i \frac{c_i^2}{(\lambda_i+\alpha)^2},
\qquad
\phi(\alpha)=\frac{N(\alpha)}{D(\alpha)}.
\end{align*}
Since $D(\alpha)>0$ for all $\alpha>0$, the quotient is well defined. 
We then differentiate both $N(a)$ and $D(a)$. 
\begin{align*}
\phi'(\alpha)
=
\frac{
N'(\alpha)D(\alpha)-N(\alpha)D'(\alpha)
}{
 D(\alpha)^2
}.
\end{align*}
Moreover,
\begin{align*}
N'(\alpha)
&=
-2\sum_i \frac{\lambda_i c_i^2}{(\lambda_i+\alpha)^3},
&
D'(\alpha)
&=
-2\sum_i \frac{c_i^2}{(\lambda_i+\alpha)^3},
\end{align*}
so
\begin{align*}
N'(\alpha)D(\alpha)-N(\alpha)D'(\alpha)
&=
2\sum_{i,j}
\frac{\lambda_i c_i^2 c_j^2(\lambda_i-\lambda_j)}
{(\lambda_i+\alpha)^3(\lambda_j+\alpha)^3}.
\end{align*}
Pairing the $(i,j)$ and $(j,i)$ terms gives
\begin{align*}
N'(\alpha)D(\alpha)-N(\alpha)D'(\alpha)
&=
\sum_{i<j}
\frac{2c_i^2 c_j^2(\lambda_i-\lambda_j)^2}
{(\lambda_i+\alpha)^3(\lambda_j+\alpha)^3}
\;\ge\;0.
\end{align*}
Hence $\phi'(\alpha)\ge 0$ for all $\alpha>0$. If there exist $i\neq j$ such that $c_i c_j\neq 0$ and $\lambda_i\neq \lambda_j$, then the corresponding pairwise term is strictly positive for every $\alpha>0$, so $\phi'(\alpha)>0$.
\end{proof}

The endpoint values are
\[
\phi_{\infty}=\frac{g_{\mathcal T}^{\top}\F_{\mathcal R}g_{\mathcal T}}{\|g_{\mathcal T}\|_2^2},\qquad
\phi_{0}=\lim_{\alpha\downarrow0}\phi(\alpha)=
\begin{cases}
0,&g_{\mathrm{null}}\neq0,\\[2pt]
\displaystyle\frac{g_{\mathcal T}^{\top}\F_{\mathcal R}^{\dagger}g_{\mathcal T}}
{g_{\mathcal T}^{\top}(\F_{\mathcal R}^{\dagger})^2g_{\mathcal T}},&g_{\mathrm{null}}=0.
\end{cases}
\]
Under the strict-monotonicity condition, a finite positive root exists exactly when $\phi_0<2\rho/\epsilon^2<\phi_\infty$. A level at or above $\phi_\infty$ uses the norm-only optimum; a level at or below $\phi_0$ uses the drift-only optimum when $g_{\mathrm{null}}=0$, with its own scale and feasible nullspace components. When $g_{\mathrm{null}}\neq0$, the positive budget excludes the lower endpoint $\phi_0=0$. These endpoint cases are checked before a one-dimensional monotone solve. See Algorithm~\ref{alg:solve_alpha_short} for the reduced-coordinate implementation.

\subsection[Choices of the Preconditioner P]{Choices of the Preconditioner $\mathbf{P}$}
\label{app:preconditioner}

Influence functions describe the derivative of a re-optimized solution under infinitesimal data perturbations. Explicit fine-tuning steps instead depend on the optimizer and its state. We compare their local update models through
\[
\Delta\theta(w) = -\eta \mathbf{P}\mathbf{G}w,
\]
where $\mathbf{P}\in \mathbb{R}^{d \times d}$ is a fixed \emph{preconditioner}, $w \in \mathbb{R}_{\geq 0}^{N_C}$ satisfies $\sum_i w_i=1$, and $\eta > 0$ is a step size. For a size-$S$ subset, this appendix uses normalized weights $w_i=\mathbf{1}\{i\in\mathcal S\}/S$, equivalent to the binary indicator divided by $S$ in the main text. The \emph{linearized target gain} is
\[
\mathrm{Gain}(w) 
:= -g_{\Tdist}^\top \Delta\theta(w) 
= \eta\,g_{\Tdist}^\top \mathbf{P}\mathbf{G}w
= \eta \sum_{i=1}^{N_C} w_i\,g_{\Tdist}^\top \mathbf{P}g_i.
\]
For a target risk with locally bounded Hessian, the actual risk decrease equals this gain plus $O(\|\Delta\theta(w)\|_2^2)$. Different choices of $\mathbf{P}$ recover exact one-step models or useful local proxies, as specified below; they do not identify a full multi-step training trajectory.

\subsubsection{Special Cases of the Preconditioner}
\label{subsec:special_cases}

\paragraph{SGD: $\mathbf{P} =  \I$.}
A single stochastic gradient step on example $z_i$ gives:
\[
\Delta\theta(e_i) = -\eta\,g_i,
\]
corresponding to $\mathbf{P} =  \I$. The gain is:
\[
\mathrm{Gain}_{\mathrm{SGD}}(w) = \eta\,g_{\Tdist}^\top \mathbf{G}w 
= \eta\sum_i w_i\,g_{\Tdist}^\top g_i,
\]
which is Euclidean gradient alignment. It does not generally equal influence alignment, which additionally depends on the training Hessian.

\paragraph{Influence / Newton: $\mathbf{P} = \mathbf{H}^{-1}$.}
Let $R(\theta)=N_C^{-1}\sum_i\ell(z_i;\theta)$ and $\mathbf H=\nabla^2R(\theta_0)$. Assume twice-continuously differentiable losses, $\nabla R(\theta_0)=0$, and $\mathbf H\succ0$. For the local minimizer $\theta_{a,w}$ of $R(\theta)+a\sum_iw_i\ell(z_i;\theta)$ that continues from $\theta_0$, implicit differentiation gives \citep{koh2017influence}
\[
\left.\frac{d\theta_{a,w}}{da}\right|_{a=0}=-\mathbf H^{-1}\mathbf G w,
\qquad
\theta_{a,w}-\theta_0=-a\mathbf H^{-1}\mathbf G w+o(a).
\]
Thus $\mathbf P=\mathbf H^{-1}$ matches the infinitesimal influence direction, with linearized gain
\[
\Delta\theta(w) = -\eta \mathbf{H}^{-1}\mathbf G w,
\qquad
\mathrm{Gain}_{\mathrm{inf}}(w) 
= \eta\sum_i w_i\,g_{\Tdist}^\top \mathbf{H}^{-1}g_i.
\]
Finite re-optimization generally differs from this first-order update. Away from a stationary point, $\mathbf H^{-1}$ may still define a Newton-style curvature model when it exists, but the influence derivative above requires the stated stationary-solution assumptions.

\paragraph{Adam: $\mathbf{P} = \mathbf{P}_{\mathrm{Adam}}$.}
Let $g_t := \nabla_\theta \ell(z_t;\theta^t)$ denote the stochastic gradient used at optimization step $t$.
The Adam optimizer maintains uncorrected moment accumulators $m^t$ and $v^t$, together with their bias-corrected versions $\widehat m^t$ and $\widehat v^t$. With steps indexed from $t=0$, its coordinate-wise update is \citep{kingma2014adam}:
\begin{equation*}
\begin{split}
     \theta^{t+1}-\theta^t
     &=
     -\eta_t \frac{\widehat m^{t+1}}{\sqrt{\widehat v^{t+1}}+\epsilon_{\mathrm{Adam}}},
     \\
     m^{t+1}
     &=
     \beta_1m^t+(1-\beta_1)g_t,
     \qquad \widehat m^{t+1}=\frac{m^{t+1}}{1-\beta_1^{t+1}},
     \\
     v^{t+1}
     &=
     \beta_2v^t+(1-\beta_2)(g_t\odot g_t),
     \qquad \widehat v^{t+1}=\frac{v^{t+1}}{1-\beta_2^{t+1}}.
\end{split}
\end{equation*}
Here $\beta_1,\beta_2\in[0,1)$, $\epsilon_{\mathrm{Adam}}>0$, and $m^0=v^0=0$. Division and square roots are elementwise.

To obtain a fixed preconditioned-gradient proxy for selection, we freeze the coordinate scales and omit first-moment history. Let $s_j=\mathbb E[g_j^2]<\infty$ be a finite second moment under a fixed warm-up gradient distribution, shared across candidates, and recall $g_i=\nabla_\theta\ell(z_i;\theta_0)$. The proxy update is
\begin{equation*}
(\Delta\theta)_j
=
-\eta\,
\frac{(g_i)_j}
{\sqrt{s_j}+\epsilon_{\mathrm{Adam}}}.
\end{equation*}
This defines the frozen diagonal preconditioner
\begin{equation*}
\mathbf{P}_{\mathrm{Adam}}
=
\diag\!\left(
    \frac{1}{\sqrt{s_1}+\epsilon_{\mathrm{Adam}}},
    \ldots,
    \frac{1}{\sqrt{s_d}+\epsilon_{\mathrm{Adam}}}
  \right)
\in \mathbb{R}^{d \times d}.
\end{equation*}
This is a second-moment scaling proxy, not a general approximation to $\mathbf H^{-1}$: large second moments reduce the coordinate scale, and cross-coordinate interactions are omitted. Second moments need not equal gradient variances. Because actual Adam also updates its moments and uses first-moment history, a fixed linear map of the current candidate gradients does not reproduce the complete Adam trajectory.

The gain under Adam preconditioning is:
\begin{equation*}
\begin{split}
\mathrm{Gain}_{\mathrm{Adam}}(w)
&=
\eta\,g_{\Tdist}^\top
\mathbf{P}_{\mathrm{Adam}}
\mathbf{G}w
\\
&=
\eta\sum_{i=1}^{N_C}
w_i\,g_{\Tdist}^\top
\mathbf{P}_{\mathrm{Adam}}
g_i
\\
&=
\eta\sum_{i=1}^{N_C}
w_i \sum_{j=1}^d
\frac{(g_{\Tdist})_j(g_i)_j}
{\sqrt{s_j}+\epsilon_{\mathrm{Adam}}}.
\end{split}
\end{equation*}
This is the coordinate-wise inner product between the target gradient and each training gradient, normalized by the Adam scaling factor.

The frozen scales can be estimated during a warm-up phase, following the use of optimizer information for selection in LESS~\citep{xia2024less}. For example, averaging the bias-corrected second moments over $T$ warm-up steps gives
\begin{equation*}
    \mathbf{P}_{\mathrm{Adam}}
    =
    \diag\!\left(
    \frac{1}{\sqrt{\bar{v}_j}+\epsilon_{\mathrm{Adam}}}
    \right),
    \qquad
    \bar{v}_j = \frac{1}{T}\sum_{t=1}^T \widehat v_j^t,
\end{equation*}
making the score efficient to compute.

\paragraph{Natural gradient: $\mathbf{P} = \mathbf{F}^{-1}$.}
For a chosen prompt distribution $\mathcal Q$, assume the common-support, smoothness, and integrability conditions of Appendix~\ref{app:prop_1} with $\Rdist$ replaced by $\mathcal Q$, and let $\mathbf F=\mathbb E_{x\sim\mathcal Q}\mathbb E_{y\sim p_{\theta_0}(\cdot|x)}[s_{\theta_0}(x,y)s_{\theta_0}(x,y)^\top]$. If $\mathbf F\succ0$, choosing $\mathbf P=\mathbf F^{-1}$ gives the natural-gradient update $\Delta\theta=-\eta\mathbf F^{-1}g_i$, whose local geometry is the quadratic KL metric under these regularity conditions. For singular $\mathbf F$, a damped inverse is well defined; a pseudoinverse instead requires separate treatment of nullspace directions, as in Theorem~\ref{thm:oracle-safe-update-appendix}.

\section{Further details on \safe{} data selector in Section~\ref{sec:real}}
\label{app:fisher_shrinkage}

\subsection{Low-Rank Working Basis}
\label{app:working_basis}

All downstream computation is carried out in a $K$-dimensional working basis
that combines the retained reference Fisher eigenspace with a residualized
target-side subspace.

\paragraph{Estimating $g_\Tdist$}
Given the target sample $\mathcal{D}_{\mathrm{tar}} = \{z_j\}_{j=1}^{N_T}$, we estimate the target gradient by the sample mean
\begin{equation}
\label{eq:raw-target-gradient}
\widehat g_{\mathrm{tar}}^{\mathrm{raw}}
:=
\frac{1}{N_T}\sum_{j=1}^{N_T}\nabla_\theta \ell(z_j;\theta_0) \in \mathbb{R}^d.
\end{equation}
This is the direct plug-in estimator of the oracle target gradient.

\paragraph{Reference basis}
Retain the top $K_R$ strictly positive eigenpairs $(\widehat\V_{K_R},\widehat{\mathbf\Lambda}_{K_R})$ of the dual Gram matrix $\frac{1}{N_R}\mathbf S_R\mathbf S_R^\top$, with $K_R$ no larger than its rank. The parameter-space reference basis and truncated Fisher are
\[
\widehat\U_{K_R}=\mathbf S_R^\top\widehat\V_{K_R}(N_R\widehat{\mathbf\Lambda}_{K_R})^{-1/2},
\qquad
\widehat\F_R^{(K_R)}=\widehat\U_{K_R}\widehat{\mathbf\Lambda}_{K_R}\widehat\U_{K_R}^\top.
\]
The recovered basis $\widehat\U_{K_R}\in\mathbb R^{d\times K_R}$ satisfies
$\widehat{\U}_{K_R}^\top \widehat{\U}_{K_R} = \I_{K_R}$ and
$\widehat{\F}_R^{(K_R)}\widehat{\U}_{K_R} = \widehat{\U}_{K_R}\widehat{\mathbf{\Lambda}}_{K_R}$.

\paragraph{Residualized task basis}
 Collect the per-example gradients $\widehat g^{\mathrm{raw}}_i$ for all $i\in[N_T+N_C]$ into
\[
\widehat{\mathbf{G}}_{TC}
:=
\begin{bmatrix}
\widehat g^{\mathrm{raw}}_1 & \cdots & \widehat g^{\mathrm{raw}}_{N_C+N_T}
\end{bmatrix}
\in\mathbb{R}^{d\times (N_C+N_T)}.
\]
A low-rank approximation of the reference Fisher alone is not sufficient to represent target-side structure outside $\mathrm{span}(\widehat \U_{K_R})$. We therefore build an additional task block by first residualizing against the reference subspace:
\begin{equation*}
\widetilde{\mathbf{G}}_{TC}
:=
(\I_{d}-\widehat \U_{K_R}\widehat \U_{K_R}^\top)\widehat{\mathbf{G}}_{TC}.
\end{equation*}
On the residualized matrix, we form the dual Gram matrix $ \frac{1}{N_T+N_C}\widetilde{\mathbf{G}}_{TC}^\top \widetilde{\mathbf{G}}_{TC}$ and
compute its top-$K_T$ strictly positive eigenpairs
$(\widehat{\V}_{K_T}, \widehat{\mathbf{ \Omega}}_{K_T})$, with $K_T$ no larger than the residual rank. We then define the added task basis by
\[
    \widehat{\U}_{K_T}
    := \widetilde{\mathbf{G}}_{TC}\,\widehat{\V}_{K_T}
       \bigl((N_C+N_T)\widehat{\mathbf{\Omega}}_{K_T}\bigr)^{-1/2}
    \in \mathbb{R}^{d \times K_{T}}.
\]
By construction, $\widehat{\U}_{K_R}^\top \widehat{\U}_{K_T} = 0$ and
$\widehat{\U}_{K_T}^\top \widehat{\U}_{K_T} = \I_{K_{T}}$. If either matrix has rank zero, its retained basis is an empty block and its retained rank is zero; no inverse of a zero eigenvalue is taken.

The motivation for this residualization is as follows. If we were to project all quantities onto $\widehat{U}_{K_R}$ alone, the quality of the approximation would depend entirely on how much of the target and candidate gradient signal lies within the reference eigenspace. When the target task and the reference distribution are largely disentangled in parameter space, the target and candidate gradients may be nearly orthogonal to
$\widehat{\mathbf U}_{K_R}$. Projecting onto the reference basis alone would then collapse all gradient coordinates to near zero, making every candidate look equally uninformative. The residual basis retains dominant gradient directions outside that subspace. Truncation can nevertheless discard target-specific directions, especially when the candidate pool dominates the residual spectrum; a nonzero raw target gradient may have zero projection into the working basis.

\paragraph{Common working basis }
Stacking the two orthonormal bases gives \eqref{eq:common_basis}. Define the reduced surrogate Fisher by
\[
\widehat\F_{R,K}:=\U_K^\top\widehat\F_R^{(K_R)}\U_K
=\diag(\widehat{\mathbf\Lambda}_{K_R},\mathbf0_{K_T})\in\mathbb R^{K\times K}.
\]
The reduced ridge-regularized matching metric is
\begin{align}
    \label{eqdef:M_alpha_K}
    \mathbf{M}_{\alpha,K}
    :=\widehat\F_{R,K}+\alpha\I_K=\begin{bmatrix}
         \widehat{\mathbf{\Lambda}}_{K_R} + \alpha \I_{K_R} & 0 \\
         0 & \alpha \I_{K_{T}}
       \end{bmatrix}.
\end{align}
The zero residual block belongs to the truncated surrogate: orthogonality to the retained reference eigenspace does not imply zero cost under the full empirical or population Fisher. In fact, exact spectral truncation gives
\[
\widehat\F_R^{\mathrm{raw}}=\widehat\F_R^{(K_R)}+E_R,
\qquad E_R\succeq0,\qquad
\|E_R\|_{\mathrm{op}}=\widehat\lambda_{K_R+1},
\]
where $\widehat\lambda_{K_R+1}=0$ if all positive eigenvalues are retained. Consequently, for any ambient update $\delta$ satisfying $\|\delta\|_2\le\epsilon$ and $\tfrac12\delta^\top\widehat\F_R^{(K_R)}\delta\le\rho$,
\[
\frac12\delta^\top\widehat\F_R^{\mathrm{raw}}\delta
\le\rho+\frac{\epsilon^2}{2}\widehat\lambda_{K_R+1}.
\]
This is an empirical quadratic bound. Transferring it to the population Fisher requires additional estimation-error control, and transferring it to KL drift also requires control of the local Taylor remainder.

\paragraph{Choosing $K_R$ and $K_{T}$}
One optional rank heuristic measures how much each eigendirection modifies $(\widehat{\F}_R^{\mathrm{raw}} + \alpha \I)^{-1}$ relative to $\alpha^{-1}\I$ at a fixed positive $\alpha$.
Let the nonzero eigenvalues of the empirical reference Fisher $\widehat\F_R^{\mathrm{raw}}$ be
$
\widehat\lambda_1 \ge \widehat\lambda_2 \ge \cdots \ge \widehat\lambda_{r_R} > 0,
$
with corresponding orthonormal eigenvectors $\widehat u_1,\dots,\widehat u_{r_R}$ when $r_R>0$, where $0\le r_R\leq \min\{N_R,d\}$.
Then,
\begin{align*}
(\widehat \F_R^{\mathrm{raw}}+\alpha \I)^{-1}
=
\alpha^{-1}\I
-
\sum_{j=1}^{r_R}
\frac{\widehat\lambda_j}{\alpha(\widehat\lambda_j+\alpha)}
\widehat u_j \widehat u_j^\top. 
\end{align*}
The j-th direction contributes through the relative correction factor
\begin{equation}
\label{eq:reference_ridge_relevance}
r_j(\alpha)
:=
\frac{\widehat\lambda_j}{\widehat\lambda_j+\alpha}.
\end{equation}
When $\widehat\lambda_j\ll\alpha$, the correction to the isotropic inverse is small in that eigendirection. This observation alone does not guarantee that discarding it preserves a particular target direction or selection.

For $r_R>0$ and a chosen tolerance $\delta_R\in(0,1)$, the heuristic is
\begin{align*}
K_R^{\mathrm{auto}}
:=
\min\left\{
K\in\{1,\ldots,r_R\}:
\frac{\sum_{j=1}^{K} r_j(\alpha)}
{\sum_{j=1}^{r_R} r_j(\alpha)}
\ge 1-\delta_R
\right\}.
\end{align*}

Similarly, when the residual rank $r_T>0$, let the positive eigenvalues of $\frac{1}{N_T+N_C}\widetilde G_{TC}^\top \widetilde G_{TC}$ be $\widehat\omega_1\ge\cdots\ge\widehat\omega_{r_T}>0$. An optional residual-energy heuristic is
\begin{align*}
K_{T}^{\mathrm{auto}}
:=
\min\left\{
K\in\{1,\ldots,r_T\}:
\frac{\sum_{j=1}^{K}\widehat\omega_j}
{\sum_{j=1}^{r_T}\widehat\omega_j}
\ge 1-\delta_T
\right\},
\end{align*}
for $\delta_T\in(0,1)$. Set $K_R^{\mathrm{auto}}=0$ when $r_R=0$ and $K_T^{\mathrm{auto}}=0$ when $r_T=0$. These are aggregate spectral criteria, not uniform guarantees on target-gradient or drift preservation. The reference criterion depends on $\alpha$: if it is used, choose a pilot $\alpha$ and fix the resulting basis before solving for damping, or explicitly repeat rank selection and damping as a heuristic. The fixed-basis monotonicity used in Algorithm~\ref{alg:solve_alpha_short} does not itself justify a joint rank-and-damping search.

\paragraph{Choosing $\alpha$ based on $(\rho,\epsilon)$.}

We replace the population Fisher and target gradient by their reduced empirical counterparts from Section~\ref{sec:real}. For a fixed working basis, the resulting Rayleigh quotient $\widehat\phi(\alpha)$ is nondecreasing, as in Proposition~\ref{prop:alpha_mono}. All $K$ target coordinates must be included, including the residual task coordinates whose surrogate Fisher eigenvalues are zero. Algorithm~\ref{alg:solve_alpha_short} detects an endpoint root or bisects when a root is strictly bracketed within the prescribed positive interval; its status distinguishes a root found to tolerance from an interval or iteration limit. Subsequent downscaling enforces the reduced surrogate budgets even when no root is returned. A zero projected target gives the zero continuous update and carries no target-alignment information in this basis.

\begin{algorithm}[th]
\caption{Finite-interval choice of $\widehat\alpha$ from $(\rho,\epsilon)$}
\label{alg:solve_alpha_short}
\begin{algorithmic}[1]
\Statex \textbf{Input:} Target feature $b=\widehat b_{\mathrm{tar}}^{(K)}$, diagonal reduced Fisher $\widehat\F_{R,K}=\diag(d_1,\ldots,d_K)\succeq0$ from \eqref{eqdef:M_alpha_K}, budgets $\rho,\epsilon>0$, interval $0<\alpha_{\min}<\alpha_{\max}$, tolerance $\tau>0$, integer iteration limit $T\ge1$.
\Statex \textbf{Output:} Damping parameter $\widehat\alpha$ and status.
\If{$b=0$}
    \State \Return $(\alpha_{\min},\texttt{zero\_target})$ \Comment{continuous update is zero}
\EndIf
\State Set $d_j\gets[\widehat\F_{R,K}]_{jj}$ and $c_j\gets b_j$ for $j=1,\ldots,K$; set $r\gets2\rho/\epsilon^2$.
\State Define
\[
\widehat\phi(a):=
\frac{\sum_{j=1}^{K}d_jc_j^2/(d_j+a)^2}
{\sum_{j=1}^{K}c_j^2/(d_j+a)^2},\qquad
\widehat\phi_\infty:=\frac{\sum_{j=1}^{K}d_jc_j^2}{\sum_{j=1}^{K}c_j^2}.
\]
\If{$r=\widehat\phi(\alpha_{\max})$}
    \State \Return $(\alpha_{\max},\texttt{endpoint\_root})$
\ElsIf{$r=\widehat\phi(\alpha_{\min})$}
    \State \Return $(\alpha_{\min},\texttt{endpoint\_root})$
\ElsIf{$r\ge\widehat\phi_\infty$}
    \State \Return $(\alpha_{\max},\texttt{inactive\_approx})$
\ElsIf{$r<\widehat\phi(\alpha_{\min})$}
    \State \Return $(\alpha_{\min},\texttt{floor\_limited})$
\ElsIf{$r>\widehat\phi(\alpha_{\max})$}
    \State \Return $(\alpha_{\max},\texttt{cap\_limited})$
\EndIf
\State $\ell\gets\alpha_{\min}$; $h\gets\alpha_{\max}$ \Comment{$\widehat\phi(\ell)<r<\widehat\phi(h)$}
\For{$t=1,\ldots,T$}
    \State $a\gets\exp((\log\ell+\log h)/2)$; $r_a\gets\widehat\phi(a)$.
    \If{$|r_a-r|\le\tau r$}
        \State \Return $(a,\texttt{root\_tolerance})$
    \ElsIf{$r_a<r$}
        \State $\ell\gets a$
    \Else
        \State $h\gets a$
    \EndIf
\EndFor
\State \Return $(\ell,\texttt{iteration\_limit})$

\end{algorithmic}
\end{algorithm}

\paragraph{Endpoint and termination guarantees.}
The pseudocode is stated in exact arithmetic. If $b$ is supported on one eigenvalue $d_\star$, then $\widehat\phi(a)\equiv d_\star$: a matching level returns an endpoint root, but this root is not unique. If all supported eigenvalues are zero, then $\widehat\phi\equiv0$ and the positive level $r$ takes the \texttt{inactive\_approx} branch. Otherwise, support on two distinct eigenvalues gives strict monotonicity, including when one is zero. The \texttt{floor\_limited} status means either that a root lies below the floor or that no positive root exists. The \texttt{cap\_limited} branch has $\widehat\phi(\alpha_{\max})<r<\widehat\phi_\infty$ and therefore a unique root above the cap. The \texttt{inactive\_approx} status means that the norm-only optimum of the reduced surrogate problem is feasible, including equality at $r=\widehat\phi_\infty$; it does not imply strictly slack drift.

For a strictly bracketed root $\alpha_\star$, every unsuccessful iteration preserves $\widehat\phi(\ell)<r<\widehat\phi(h)$ and halves $\log(h/\ell)$. Thus, after $T$ unsuccessful iterations,
\[
0\le\log(\alpha_\star/\ell)
\le 2^{-T}\log(\alpha_{\max}/\alpha_{\min}),
\qquad
0\le\frac{r-\widehat\phi(\ell)}{r}
\le 2^{1-T}\log(\alpha_{\max}/\alpha_{\min}).
\]
For the second bound, the positive numerator and denominator defining $\widehat\phi$ satisfy $0\le a\widehat\phi'(a)/\widehat\phi(a)\le2$: differentiating their logarithms gives twice the difference of two weighted means of $a/(d_j+a)\in[0,1]$, and monotonicity gives the lower bound. Integrating this inequality and using $1-e^{-x}\le x$ proves the result. The \texttt{root\_tolerance} status certifies only the stated relative quotient residual; an arbitrary iteration limit need not meet that tolerance. These guarantees assume exact spectral data and arithmetic; a numerical implementation must also handle rounding and overflow.

\subsubsection[Scaling value for Delta theta star]{Scaling value for $\Delta\theta^*$}

For fixed $\alpha>0$, use the reduced matrices $\widehat\F_{R,K}$ and $\mathbf M_{\alpha,K}$ from \eqref{eqdef:M_alpha_K}. Define the unscaled reduced direction and practical update by
\[
h_{\alpha,K}:=\mathbf M_{\alpha,K}^{-1}\widehat b_{\mathrm{tar}}^{(K)}\in\mathbb R^K,
\qquad
\widehat\Delta\theta_K^\star:=-h_{\alpha,K}/\widehat\mu_K\in\mathbb R^K.
\]
Their ambient lifts are $\U_Kh_{\alpha,K}$ and $\U_K\widehat\Delta\theta_K^\star\in\mathbb R^d$, respectively. We use the smallest downscaling factor subject to $\widehat\mu_K\ge1$ that satisfies the imposed surrogate reference and norm budgets:
\begin{align}
\label{eqdef:hat_normalized_value_K}
\widehat\mu_K
:=
\max\left\{
1,\;
\sqrt{\frac{h_{\alpha,K}^\top\widehat\F_{R,K}h_{\alpha,K}}{2\rho}},\;
\frac{\|h_{\alpha,K}\|_2}{\epsilon}
\right\}.
\end{align}
The second and third terms are included only when their respective budgets are imposed. When both budgets are imposed, direct substitution gives $\tfrac12(\widehat\Delta\theta_K^\star)^\top\widehat\F_{R,K}\widehat\Delta\theta_K^\star\le\rho$ and $\|\widehat\Delta\theta_K^\star\|_2\le\epsilon$. If either budget is omitted, only the retained budget is guaranteed. The corresponding bounds also hold for the ambient lift with the truncated Fisher, because $\U_K$ has orthonormal columns. If $b=0$, then $h_{\alpha,K}=0$, $\widehat\mu_K=1$, and the update is zero. The budgets need not be saturated: the restriction $\widehat\mu_K\ge1$ deliberately permits only shrinking the unscaled direction. In particular, a finite cap used in the inactive case is an approximation to a direction, not an oracle-optimal step length. Indeed, $\|\widehat\Delta\theta_K^\star\|_2\le\|b\|_2/\alpha$, so with fixed nonzero $b$ the practical step tends to zero as the cap tends to infinity, whereas the norm-only oracle step has norm $\epsilon$. Even at an exact finite root, both budgets are saturated only if $\|h_{\alpha,K}\|_2/\epsilon\ge1$; otherwise $\widehat\mu_K=1$ and both are slack. A root residual within $\tau r$ likewise controls the ratio of the two unscaled costs, not optimality or budget saturation.

This practical construction should be contrasted with the oracle formulation
$
\Delta\theta^\star
=
-\epsilon\,
\frac{(\mathbf F_{\mathcal R}+\alpha \I)^{-1}g_{\mathcal T}}
{\|(\mathbf F_{\mathcal R}+\alpha \I)^{-1}g_{\mathcal T}\|_2}
$
in the interior regime with positive multipliers for both constraints, where \(\alpha\) is chosen so that the normalized update satisfies the reference-cost constraint
$
\frac12(\Delta\theta^\star)^\top \mathbf F_{\mathcal R}\Delta\theta^\star = \rho
$. The practical construction is a feasible surrogate heuristic; it does not inherit oracle optimality, a population drift budget, or a guarantee for subsequent fine-tuning.

\subsection{Selection Algorithm}
\label{app:selection}

Recall the practical reduced surrogate matching target
$\widehat{\Delta}\theta_K^\star(\alpha)$ and the subset-induced reduced
update $\widehat\Delta\theta_K(w)$ from
\eqref{eq:real_updated_direction_tar} and
\eqref{eq:adam_updated_direction_cand}, respectively. Throughout this section, $\alpha>0$ is finite. Define the whitened candidate-update matrix and target update by
\begin{align*}
\widetilde{\mathbf{\Delta}}_K
&:=
\mathbf{M}_{\alpha,K}^{1/2}
\left(-\frac{\eta}{S}\mathbf{P}_K\mathbf{B}_K\right)
\in\mathbb{R}^{K\times N_C},
\\
\widetilde{\Delta}\theta_K^\star
&:=
\mathbf{M}_{\alpha,K}^{1/2}
\widehat{\Delta}\theta_K^\star(\alpha).
\end{align*}
Then the matching objective
\begin{align*}
w_K^{\mathrm{opt}}
\in
\arg\min_{\substack{w\in\{0,1\}^{N_C}\\ \mathbf 1^\top w=S}}
\frac12
\left\|
\widehat\Delta\theta_K(w)-\widehat\Delta\theta_K^\star(\alpha)
\right\|_{\mathbf{M}_{\alpha,K}}^2
\end{align*}
can be rewritten as the whitened objective
\begin{align*}
w_K^{\mathrm{opt}}
\in
\arg\min_{\substack{w\in\{0,1\}^{N_C}\\ \mathbf 1^\top w=S}}
\frac12\left\|
\widetilde{\mathbf{\Delta}}_K w-\widetilde{\Delta}\theta_K^\star
\right\|_2^2 .
\end{align*}
The forward greedy procedure below approximates this cardinality-constrained optimum; no exact solution or approximation factor is claimed. For a singleton contribution $w=e_i$, keeping the final normalization $1/S$ fixed even when $S>1$, we have
\begin{align*}    
\frac12\left\|
\widetilde{\Delta}_{K}(i)-\widetilde{\Delta}\theta_K^\star
\right\|_2^2
=
\frac12\|\widetilde{\Delta}\theta_K^\star\|_2^2
-
\left\langle
\widetilde{\Delta}_{K}(i),\widetilde{\Delta}\theta_K^\star
\right\rangle
+
\frac12\|\widetilde{\Delta}_{K}(i)\|_2^2.
\end{align*}
Since the first term is independent of \(i\), ranking singleton subsets by smaller objective value is equivalent to ranking them by
\begin{align*}
s_i
:=
\left\langle
\widetilde{\Delta}_{K}(i),\widetilde{\Delta}\theta_K^\star
\right\rangle
-
\frac12\|\widetilde{\Delta}_{K}(i)\|_2^2.
\end{align*}
Thus, $s_i$ is exactly the signed decrease in the whitened quadratic objective from the empty subset to candidate $i$ alone. A singleton is not a feasible final subset when $S>1$.
Compared with LESS~\citep{xia2024less}, which ranks examples by alignment alone, our selector adds a quadratic term that penalizes costly components under the empirical reference geometry. We provide the complete selection algorithm in Algorithm~\ref{alg:greedy_safe_selector}.

\begin{algorithm}[ht]
\caption{Forward greedy approximation to the whitened reduced objective}
\label{alg:greedy_safe_selector}
\begin{algorithmic}[1]
\Statex \textbf{Input:} Whitened candidate-update matrix
$\widetilde{\mathbf{\Delta}}_K
= [\widetilde{\Delta}_{K}(1)\ \cdots\ \widetilde{\Delta}_{K}(N_C)]$,
whitened target update $\widetilde{\Delta}\theta_K^\star$,
integer subset budget $1\le S\le N_C$, integer shortlist size $S\le m\le N_C$.
\Statex \textbf{Output:} Selected subset
$\widehat{\mathcal S}\subseteq[N_C]$ with $|\widehat{\mathcal S}|=S$,
and selection vector $\widehat w$.

\State For each $i \in [N_C]$, compute the singleton score
\begin{align*}
s_i \gets
\left\langle
\widetilde{\Delta}_{K}(i),\widetilde{\Delta}\theta_K^\star
\right\rangle
-\frac12\|\widetilde{\Delta}_{K}(i)\|_2^2.
\end{align*}
\State Let $ \mathcal{M}_m \subseteq [N_C]$ be the indices of the top-$m$ scores
$\{s_i\}_{i=1}^{N_C}$, using any fixed rule to break ties.
\State Initialize $\widehat{\mathcal S} \gets \emptyset$ and
$\tilde s \gets 0 \in \mathbb{R}^K$.
\For{$t=1,\dots,S$}
    \State Choose
    \begin{align*}
    j^\star \in
    \arg\min_{j \in \mathcal M_m \setminus \widehat{\mathcal S}}
    \left(
    \frac12\|\widetilde{\Delta}_{K}(j)\|_2^2
    + \left\langle \tilde s,\widetilde{\Delta}_{K}(j)\right\rangle
    - \left\langle
    \widetilde{\Delta}\theta_K^\star,\widetilde{\Delta}_{K}(j)
    \right\rangle
    \right).
    \end{align*}
    \State Update
    \begin{align*}
    \widehat{\mathcal S} \gets \widehat{\mathcal S} \cup \{j^\star\},
    \qquad
    \tilde s \gets \tilde s + \widetilde{\Delta}_{K}(j^\star).
    \end{align*}
\EndFor
\State Set $\widehat w_i \gets \mathbf 1\{i\in\widehat{\mathcal S}\}$ for
all $i\in[N_C]$.
\State \Return $\widehat{\mathcal S},\widehat w$.
\end{algorithmic}
\end{algorithm}

The singleton scores construct the shortlist. Each greedy choice uses the current sum $\tilde s$ and hence includes interactions with already selected examples; ties may be resolved arbitrarily. For $m>S$, these interactions can change the final subset relative to static top-$S$ ranking. When $m=S$, the final subset is necessarily the entire shortlist, irrespective of the greedy order. The procedure takes all $S$ steps even when the best remaining addition increases the objective, since the cardinality constraint is exact. Both shortlisting and greedy optimization can introduce matching error.

\paragraph{What matching error implies for drift.}
For any population-oracle update $\delta^\star$ satisfying $\tfrac12(\delta^\star)^\top\F_{\Rdist}\delta^\star\le\rho$ and any candidate-induced ambient update $\delta_{\mathcal S}$, let $e:=\delta_{\mathcal S}-\delta^\star$. The triangle inequality for the Fisher seminorm gives
\[
\frac12\delta_{\mathcal S}^\top\F_{\Rdist}\delta_{\mathcal S}
\le\left(\sqrt\rho+\frac{\|e\|_{\F_{\Rdist}}}{\sqrt2}\right)^2,
\qquad
\|e\|_{\F_{\Rdist}}\le\|e\|_{\F_{\Rdist}+\alpha\I}.
\]
Thus matching error can increase the quadratic cost beyond $\rho$; the matching objective does not impose a hard feasibility constraint on the selected subset. The same inequality holds within the empirical surrogate geometry, but a small reduced matching error alone does not control omitted directions, Fisher estimation error, or the difference between the surrogate and actual optimizer update. Under Proposition~\ref{prop:local-kl}, actual KL drift has the additional local remainder $o(\|\delta_{\mathcal S}\|_2^2)$. These local statements do not guarantee the drift of a subsequent multi-step fine-tuning trajectory.

\subsection{Entanglement diagnostics in the working subspace}\label{appsec:entanglement_analysis}

\subsubsection{Entanglement quadrant}\label{sec:entanglement_main}
A key insight from Proposition~\ref{prop:oracle-safe-direction} is that the target gradient's orientation in the reference Fisher geometry affects the local gain--drift trade-off. As a descriptive diagnostic, we use the low-rank basis
$
    \U_K = [\,\widehat \U_{K_R}\;\; \widehat \U_{K_T}\,]
$ to describe where candidate alignment with the reduced target direction occurs. Decompose the candidate- and target-gradient coordinates in this basis as,
\begin{align*}
    \widehat b_i^{(K)}
    =
    \U_K^\top \widehat g_i^{\mathrm{raw}}
    =
    \begin{bmatrix}
       \hat{\U}_{K_R}^\top \widehat g_i^{\mathrm{raw}}\\
   \hat{ \U}_{K_T}^\top \widehat g_i^{\mathrm{raw}}
    \end{bmatrix}=
    \begin{bmatrix}
        \widehat b_{i}^R\\
        \widehat b_{i}^T
    \end{bmatrix},
    \qquad
    \widehat b_{\mathrm{tar}}^{(K)}    =
    \U_K^\top \widehat g_{\mathrm{tar}}^{\mathrm{raw}}
    =
    \begin{bmatrix}
       \widehat\U_{K_R}^\top \widehat g_{\mathrm{tar}}^{\mathrm{raw}}\\
    \widehat\U_{K_T}^\top \widehat g_{\mathrm{tar}}^{\mathrm{raw}}
    \end{bmatrix}
    =
    \begin{bmatrix}
        \widehat b_{\mathrm{tar}}^R\\
        \widehat b_{\mathrm{tar}}^T
    \end{bmatrix}.
\end{align*}
Here, $ \widehat b_{i}^R$ refers to the projection of the candidate gradients along the reference space, while $ \widehat b_{i}^T$ refers to their projection in the residualized target space. We define two target-alignment scores under the reference subspace and the residual task subspace, respectively,
\begin{align*}
      d_i^R
    =
   ( \widehat b_{i}^R)^{\top}
    ( \mathbf{\widehat\Lambda}_{K_R}+\alpha \I_{K_R})^{-1}
    \widehat b_{\mathrm{tar}}^R,
    \qquad
    d_i^T
    =
    \alpha^{-1}
    (\widehat b_{i}^T)^{\top}
    \widehat b_{\mathrm{tar}}^T.
\end{align*}
Here, $d_i^R>0$ indicates positive damped target alignment within the retained reference block, while $d_i^T>0$ indicates positive alignment in the residual task block. The latter block is orthogonal to the retained reference subspace, but need not have zero cost under the full Fisher. An empty block contributes zero to its alignment score. Their sum is $(\widehat b_i^{(K)})^\top\mathbf M_{\alpha,K}^{-1}\widehat b_{\mathrm{tar}}^{(K)}$, which in general differs from the linear term in the whitened matching score. Indeed, substituting~\eqref{eq:real_updated_direction_tar} gives
\[
\left\langle\widetilde\Delta_K(i),\widetilde\Delta\theta_K^\star\right\rangle
=\frac{\eta}{S\widehat\mu_K}
(\widehat b_i^{(K)})^\top\mathbf P_K^\top\widehat b_{\mathrm{tar}}^{(K)},
\]
because the matching metric cancels the inverse metric in the target update. We partition candidates by the signs of $(d_i^R,d_i^T)$:
\begin{equation*}
\begin{aligned}
 Q_{\mathrm{ref}}
&:= \{\, z_i \in \mathcal D_{\mathrm{train}}  : d_i^T \le 0,\; d_i^R > 0 \,\}, \qquad Q_{\mathrm{clean}}
:= \{\, z_i \in \mathcal D_{\mathrm{train}} : d_i^T > 0,\; d_i^R \le 0 \,\}, 
 \\
Q_{\mathrm{ent}}
&:= \{\, z_i \in \mathcal D_{\mathrm{train}} : d_i^T > 0,\; d_i^R > 0 \,\}, 
\qquad Q_{\mathrm{mis}}
:= \{\, z_i \in \mathcal D_{\mathrm{train}}  : d_i^T \le 0,\; d_i^R \le 0 \,\}.
\end{aligned}
\end{equation*}
We then measure the proportions of the candidate examples in each quadrant by,
\begin{align}\label{eqdef:quadrant_entanglment}
    \pi_Q
    :=
    \frac{1}{N_C}
    \sum_{z_i\in \mathcal D_{\mathrm{train}}}
    \mathbf 1\{z_i\in Q\},
    \qquad
    Q\in
    \{ Q_{\mathrm{clean}},
       Q_{\mathrm{ent}},
       Q_{\mathrm{ref}},
       Q_{\mathrm{mis}}\}.
\end{align}
These four proportions summarize signs of blockwise alignment in the candidate pool. $Q_{\mathrm{ent}}$ has positive alignment in both blocks; $Q_{\mathrm{clean}}$ has positive residual alignment and non-positive reference alignment; $Q_{\mathrm{ref}}$ has the reverse pattern; and $Q_{\mathrm{mis}}$ has non-positive alignment in both blocks. The signs alone give neither the magnitudes of the two contributions nor a reference-cost bound. In particular, membership in $Q_{\mathrm{clean}}$ need not imply positive total target alignment: a negative reference contribution can dominate the positive residual contribution. Total gain for a preconditioned optimizer also depends on $\mathbf P$. We provide $\pi_{Q_{\mathrm{clean}}}$ and $\pi_{Q_{\mathrm{ent}}}$ of our experiments in Table~\ref{tab:datasets}.

The quadrant percentages in \eqref{eqdef:quadrant_entanglment} are descriptive diagnostics, not selection rules or guarantees of target improvement or drift control. \textsc{Safe-Drift} does not simply select all examples from $Q_{\mathrm{clean}}$. These pool-level percentages describe the available blockwise alignment patterns; the selected subset is determined by the matching procedure.

\subsubsection{Reference-entanglement diagnostics}
\label{sec:reference-ent}
The same decomposition also permits diagnostics of blockwise gradient energy, complementing the alignment signs above. These energy ratios describe the retained subspace, without establishing gain or drift for an update.

We use two complementary \emph{reference-entanglement diagnostics} for a reduced
gradient, when $K_R,K_T>0$ and the corresponding denominators are positive. Here $\mathbf\Lambda_R:=\widehat{\mathbf\Lambda}_{K_R}$ is the diagonal matrix of retained reference eigenvalues:
\begin{align}\label{eq:reference-entanglement diagnostics}
\rho_E(b)
=
\frac{\|b^R\|_2^2/K_R}
{\|b^R\|_2^2/K_R+\|b^T\|_2^2/K_T},\qquad \rho_{\mathrm{safe}}(b)
=
\frac{(b^R)^\top \mathbf{\Lambda}_R b^R / K_R}
{(b^R)^\top \mathbf{\Lambda}_R b^R / K_R+\alpha\|b^T\|_2^2/K_T}.
\end{align}
The quantity $\rho_E(b)$ compares mean squared gradient coordinates in the two blocks; it is dimension-normalized rather than the fraction of total gradient energy. We retain the numerical definition of $\rho_{\mathrm{safe}}(b)$ as a hybrid diagnostic, weighting the reference block by $\mathbf\Lambda_R$ and the residual block by $\alpha$. It is not the energy allocation under $\mathbf M_{\alpha,K}$, whose reference block is $\mathbf\Lambda_R+\alpha\I$, nor the allocation of the preconditioned update.

For equal, strictly positive squared magnitude in every coordinate, the baselines are
\[
\rho_E=\frac12,\qquad
\rho_{\mathrm{safe}}=\frac{\bar\lambda_R}{\bar\lambda_R+\alpha},
\qquad \bar\lambda_R:=\frac{\tr(\mathbf\Lambda_R)}{K_R}.
\]
Thus $K_R/K$ is not a general baseline for equal energy per coordinate in these diagnostics. When a block dimension or a denominator is zero, the corresponding ratio is undefined and should not be interpreted or plotted as an observed value.

We apply these diagnostics to target and candidate gradients together with the alignment signs from Section~\ref{sec:entanglement_main}. A large value indicates greater relative energy in the retained reference block under the specified weighting; it does not measure absolute update magnitude, establish target gain, or bound reference drift. Their role is to describe the structured working subspace used by the selector.

\section{Experiments}\label{appsec:experiments}
\subsection{Synthetic linear experiments}\label{app:exp_linear}

To build better intuition on the performance of our method in simple settings, we begin with a family of synthetic linear models designed to isolate the two statistical ingredients highlighted by our theory: the estimation problem (estimating the target gradient $g_{\Tdist}$ from few target examples, and estimating the reference geometry $\mathbf{F}_{\Rdist}$ from limited reference data) and the optimization problem (maximizing first-order target gain under an off-target budget). 

\paragraph{Linear-Gaussian setup.}
We first consider a linear Gaussian model with $\sigma^2>0$ and noise independent of the input in each data-generating distribution. For an input $x \in \R^d$, the base model is
\begin{align*}
p_\theta(y \mid x) = \mathcal{N}(x^\top \theta, \sigma^2),
\qquad
\ell((x,y);\theta) = \frac{1}{2\sigma^2}(y - x^\top \theta)^2.
\end{align*}
Target examples are generated by
\begin{align*}
x \sim \mathcal{N}(0,\mathbf{\Sigma}_{\Tdist}),
\qquad
y = x^\top \beta_{\Tdist} + \xi,
\qquad
\xi \sim \mathcal{N}(0,\sigma^2),
\end{align*}
while reference inputs are drawn from
\begin{align*}
x \sim \mathcal{N}(0,\mathbf{\Sigma}_{\Rdist}).
\end{align*}
We set the base point to $\theta_0 = 0$. In this model, the per-example gradient and target gradient at the base point are
\begin{align*}
g_i= \nabla_\theta \ell(z_i;\theta_0)
= -\frac{1}{\sigma^2} y_i x_i, \qquad
g_{\Tdist}
= \nabla_\theta L_{\Tdist}(\theta_0)
= -\frac{1}{\sigma^2}\mathbf{\Sigma}_{\Tdist} \beta_{\Tdist},
\end{align*}
and the reference Fisher is
\begin{align*}
\mathbf{F}_{\Rdist}
= \E_{x \sim \Rdist}\,
\E_{y \sim p_{\theta_0}(\cdot \mid x)}
\!\left[
\nabla_\theta \log p_{\theta_0}(y \mid x)
\nabla_\theta \log p_{\theta_0}(y \mid x)^\top
\right]
= \frac{1}{\sigma^2}\mathbf{\Sigma}_{\Rdist}.
\end{align*}
In this synthetic setting, the raw per-example gradient
$g_i$ is available in closed form; it plays the same role
as $\widehat g_i^{\mathrm{raw}}$ in the empirical algorithm.
Moreover, unlike the general nonlinear setting where the quadratic expansion is local, the reference drift is \emph{exact} here:
\begin{align*}
D_{\Rdist}(\Delta\theta)
=
\E_{x \sim \Rdist}
\big[
\KL(p_{\theta_0}(\cdot \mid x)\,\|\,p_{\theta_0+\Delta\theta}(\cdot \mid x))
\big]
=
\frac{1}{2\sigma^2}\Delta\theta^\top \mathbf{\Sigma}_{\Rdist} \Delta\theta
=
\frac{1}{2}\Delta\theta^\top \mathbf{F}_{\Rdist} \Delta\theta.
\end{align*}
The exact target improvement also includes curvature:
\[
L_\Tdist(0)-L_\Tdist(\Delta\theta)
=-g_\Tdist^\top\Delta\theta-\frac{1}{2\sigma^2}\Delta\theta^\top\mathbf\Sigma_\Tdist\Delta\theta.
\]
Thus first-order alignment is not identical to realized target improvement, even though reference drift is exactly quadratic here.

\paragraph{Updates and evaluation.}
Given a selected subset $\mathcal S \subset \Dtrain$ of size $S$, we form the update
\begin{align*}
\Delta\theta(\mathcal S)
=
-\frac{\eta}{S} \mathbf{P} \sum_{i \in \mathcal S} g_i,
\end{align*}
where $\mathbf{P}$ is the preconditioner associated with the update rule. In particular, we use $\mathbf{P} = \mathbf{I}_d$ for stochastic gradient descent.
We evaluate each selected update using
\begin{align*}
\mathrm{Gain}_{\mathrm{held}}(\mathcal S)
=
L_{\Tdist}(\theta_0)
-
L_{\Tdist}(\theta_0 + \Delta\theta(\mathcal S)),
\end{align*}
\begin{align*}
\mathrm{Drift}_{\mathrm{held}}(\mathcal S)
=
D_{\Rdist}(\Delta\theta(\mathcal S)),
\end{align*}
and, when the population quantities are known and both updates are nonzero, the directional error relative to the oracle update in the positive-multiplier regime
\begin{align*}
\Delta\theta^\star
\propto
-(\mathbf{F}_{\Rdist} + \alpha \mathbf{I}_d)^{-1} g_{\Tdist},
\qquad
\mathrm{DirErr}(\mathcal S)
=
1 - \cos\!\big(\Delta\theta(\mathcal S), \Delta\theta^\star\big).
\end{align*}

\paragraph{Experiment 1: Low-dimensional geometry ($d=2$).}

\begin{figure}[t]
    \centering
    \begin{subfigure}[t]{0.48\textwidth}
        \centering
        $\displaystyle v_\alpha=\frac{u_1}{25+\alpha}+\frac{u_2}{1+\alpha}$
        \caption{$\mathbf U=\mathbf I_2$: $u_1$ is $45^\circ$ from $-g_\Tdist$.}
        \label{fig:geometry_aligned}
    \end{subfigure}\hfill
    \begin{subfigure}[t]{0.48\textwidth}
        \centering
        $\displaystyle v_\alpha=\frac{(\sqrt3+1)u_1}{2(25+\alpha)}+\frac{(\sqrt3-1)u_2}{2(1+\alpha)}$
        \caption{$\mathbf U$ rotated by $\pi/6$: $u_1$ is $15^\circ$ from $-g_\Tdist$.}
        \label{fig:geometry_rotated}
    \end{subfigure}
    \caption{Analytic damped directions $v_\alpha/\|v_\alpha\|_2$, $\alpha\ge0$, for the two geometries in \eqref{eq:exp1_linear}. The reference spectrum is $(25,1)$, so $u_2$ has positive curvature. Normalization describes direction only; the feasible step length depends on both budgets. In particular, $\rho=0$ permits only the zero update in this positive-definite example.}
    \label{fig:geometry}
\end{figure}

In the first experiment, we ask how the reference geometry reshapes $\Delta\theta^\star$ as the damping $\alpha$ varies, and also how this reshaping depends on the orientation of the expensive direction $u_1$ relative to the target direction $-g_{\Tdist}$. We choose
\begin{equation}\label{eq:exp1_linear}
\mathbf{\Sigma}_{\Rdist} = \mathbf{U} \diag(25,1) \mathbf{U}^\top,
\qquad
\mathbf{\Sigma}_{\Tdist} = \mathbf{I}_2,
\qquad
\beta_{\Tdist} = (1,1)^\top,
\qquad
\sigma^2 = 1,
\end{equation}
with $\mathbf{U}=\begin{pmatrix}u_1 & u_2\end{pmatrix}$ a rotation, so $-g_{\Tdist}=\beta_{\Tdist}$ points at $45^\circ$ and $u_1$ costs $25\times$ more drift than $u_2$. We compare (a) $\mathbf{U}=\mathbf{I}_2$, where $u_1$ is $45^\circ$ from $-g_{\Tdist}$, and (b) $\mathbf{U}$ a rotation by $\pi/6$, where $u_1$ is only $15^\circ$ away (used in Experiments~2--3). 

In both cases reference weighting turns the direction toward $u_2$, and increasing damping rotates it back toward $-g_{\Tdist}$. The rotation depends on both the initial angle and $\alpha$; neither orientation has a uniformly larger rotation for all damping values.

\paragraph{Experiment 2: Finite-pool subset selection in $d=2$.}
\begin{figure}[th]
    \centering
\includegraphics[width=\textwidth]{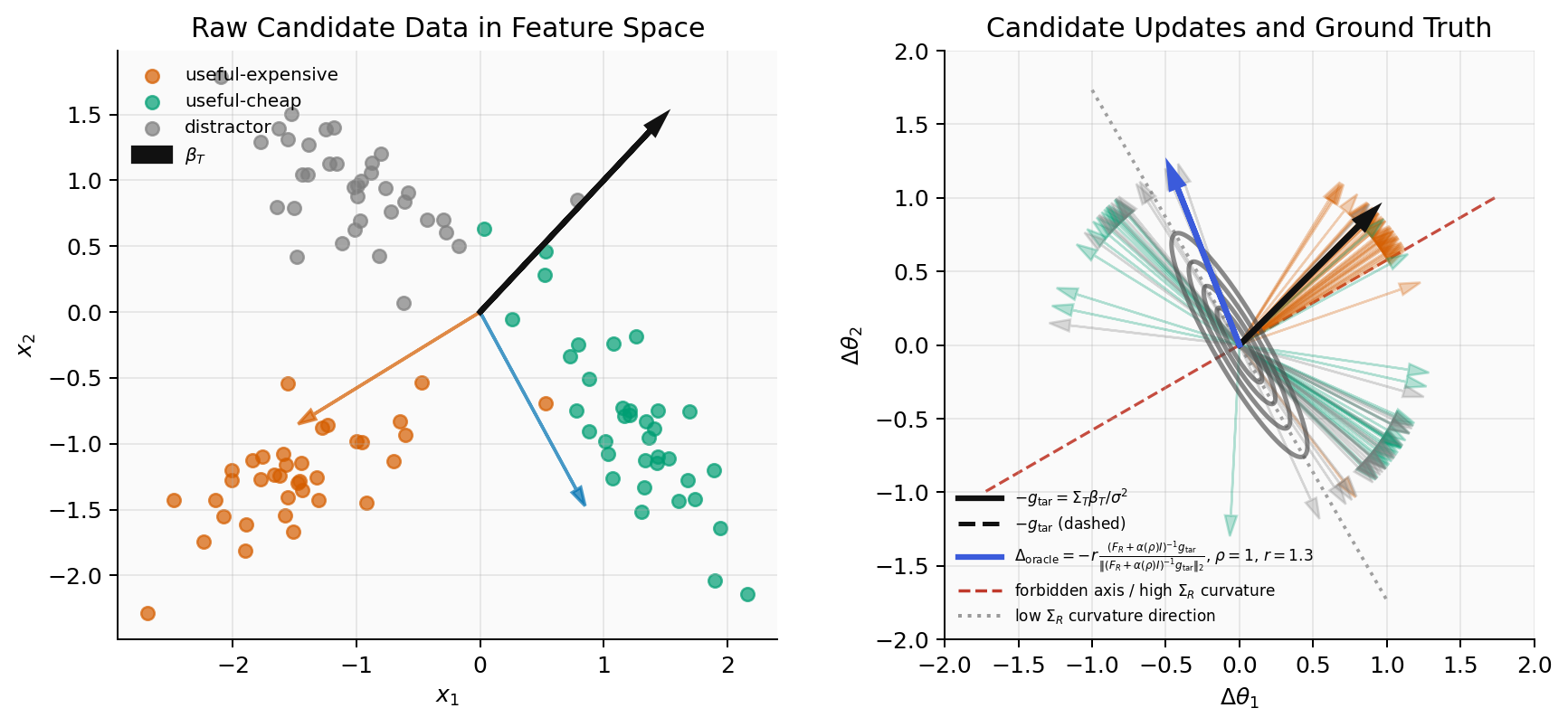}
\caption{The candidate pool in the setting of Experiment~2 (Equation~\eqref{eq:exp2}). \emph{Left:} the inputs $x_i$. \emph{Right:} the candidate descent directions, proportional to $-g_i$, displayed in update coordinates. Here $-g_{\Tdist} = \sigma^{-2}\mathbf{\Sigma}_{\Tdist}\beta_{\Tdist} = \beta_{\Tdist}$, so the black arrow is the target direction $-g_{\Tdist}$.}\label{fig:lin-exp-data-selection-raw-data}
\end{figure}

\begin{figure}[th]
    \centering
    \includegraphics[width=\textwidth]{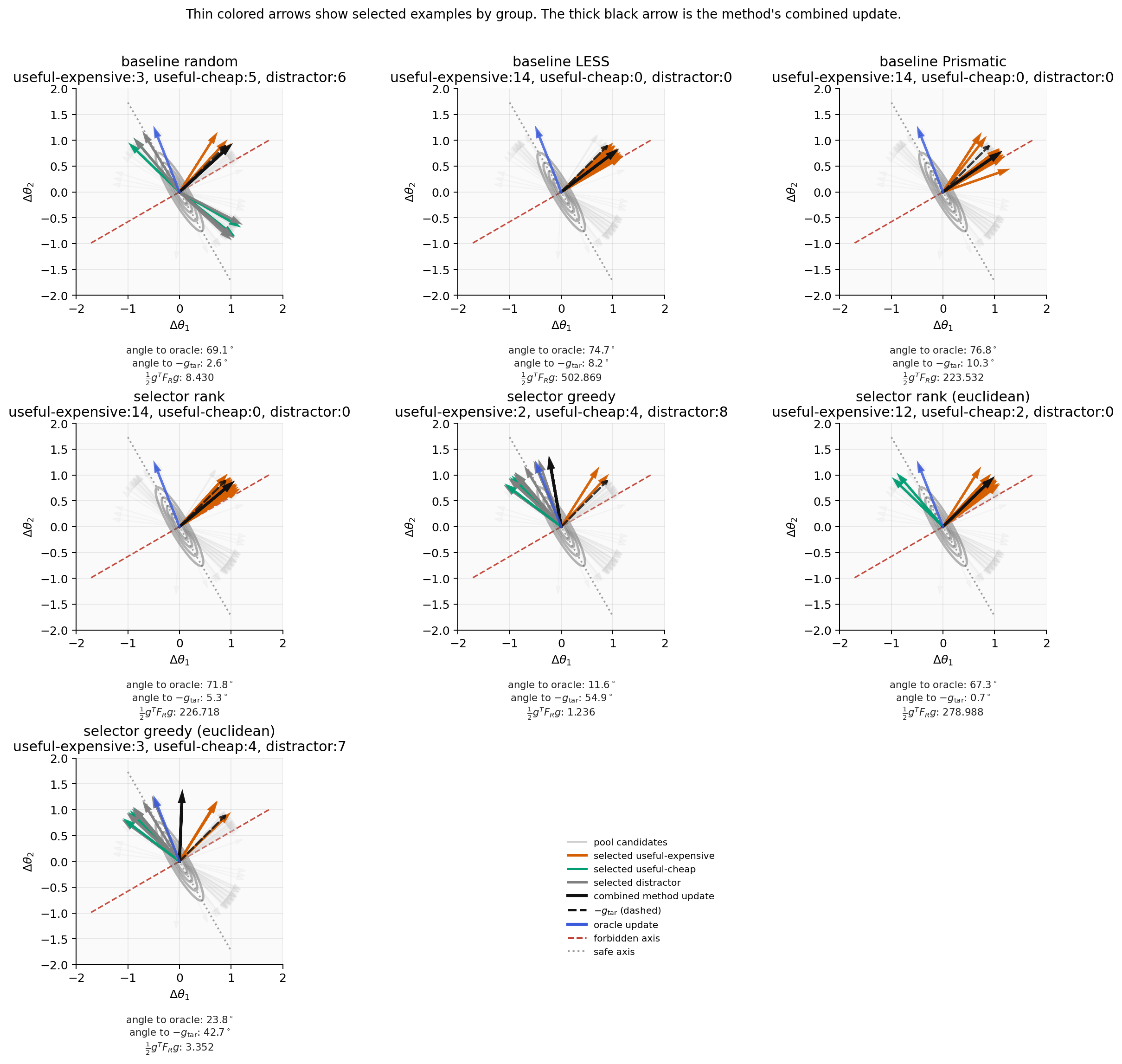}
    \caption{Selection methods on the pool of Figure~\ref{fig:lin-exp-data-selection-raw-data}. For each method: candidate descent directions proportional to $-g_i$, the selected subset (highlighted), and the resulting update direction $\Delta\theta(\mathcal S)$. The graphic's ``forbidden'' and ``safe'' labels denote the high- and low-curvature axes; neither is prohibited or has zero reference cost under the spectrum $(25,1)$.}
    \label{fig:lin-exp-data-selection}
\end{figure}
To visualize which points each method would preferentially select, we consider the setting of Experiment 1 (Equation~\ref{eq:exp1_linear}) and introduce a finite candidate pool, since Experiment~1 contains no data to select from. We draw it from a three-component Gaussian mixture with different input-variance orientations:
\begin{equation}\label{eq:exp2}
x_i \sim \frac{1}{3}\mathcal{N}(0,\mathbf{\Sigma}_1)
+ \frac{1}{3}\mathcal{N}(0,\mathbf{\Sigma}_2)
+ \frac{1}{3}\mathcal{N}(0,\mathbf{\Sigma}_3),
\qquad
y_i = x_i^\top \beta_{\Tdist} + \xi_i,
\qquad
\xi_i \sim \mathcal{N}(0,\sigma^2),    
\end{equation}

with
\begin{align*}
\mathbf{\Sigma}_1 = \mathbf{U} \diag(9,0.25) \mathbf{U}^\top,
\qquad
\mathbf{\Sigma}_2 = \mathbf{U} \diag(0.25,9) \mathbf{U}^\top,
\qquad
\mathbf{\Sigma}_3 = \mathbf{I}_2.
\end{align*}
The first two components concentrate input variance along high- and low-curvature reference directions, respectively, and the third supplies isotropic inputs. The induced gradients also depend on labels and noise and need not be isotropic; component membership alone does not guarantee usefulness or a particular update cost. The raw data---and the corresponding descent directions---are shown in Figure~\ref{fig:lin-exp-data-selection-raw-data}. For each method, we visualize the candidate descent directions, highlight the selected subset, and plot the resulting update direction (Figure~\ref{fig:lin-exp-data-selection}). 

\paragraph{Experiment 3: Trade-offs in Alignment/ off-target drift in $d=2$.}


Experiments~1 and~2 hold the target direction fixed. We now compare two target specifications at fixed reference geometry to illustrate how the target--drift trade-off depends on \emph{entanglement}: the extent to which $-g_{\Tdist}$ overlaps the expensive directions of $\mathbf{F}_{\Rdist}$. Results are reported in Figure~\ref{fig:lin-exp-data-selection-tradeoffs}. 

We compare the two instances:
\begin{itemize}
    \item \textbf{entangled:} $\beta_{\Tdist}=(1,1)^\top$, which lies $15^\circ$ from $u_1$ (the rotated case of Experiment~1);
    \item \textbf{disentangled:} $\beta_{\Tdist}=u_2=(-\tfrac12,\tfrac{\sqrt3}{2})^\top$, exactly orthogonal to $u_1$.
\end{itemize}
The two target gradients also differ in norm ($\sqrt2$ versus $1$), so this comparison does not isolate orientation alone. For each method we measure, for its selected update $\Delta\theta$:
\begin{align*}
\text{alignment: } -g_{\Tdist}^\top\Delta\theta,
\qquad
\text{drift: } \tfrac12\Delta\theta^\top\mathbf{F}_{\Rdist}\Delta\theta,
\end{align*}
The plotted drift axis in Figure~\ref{fig:lin-exp-data-selection-tradeoffs} is labeled $\Delta\theta^\top\F_\Rdist\Delta\theta$, twice the theoretical quadratic cost above; the displayed red line is a nominal reference level and is not used here to certify feasibility. Results are averaged over $100$ random seeds.

In the entangled panel, \safe{} reduces mean plotted drift relative to the other selectors while giving up some target alignment. In the disentangled panel, the selectors attain different alignment--drift trade-offs; \safe{} is not the minimum-drift method. A target direction lying in a lower-curvature eigenspace does not by itself make the budget inactive, and approximate subset matching does not certify every selected update as feasible.

\paragraph{Experiment 4: Estimation error and spectral decay in $d=200$.}
\begin{figure}[th]
    \centering
    \includegraphics[width=0.8\linewidth]{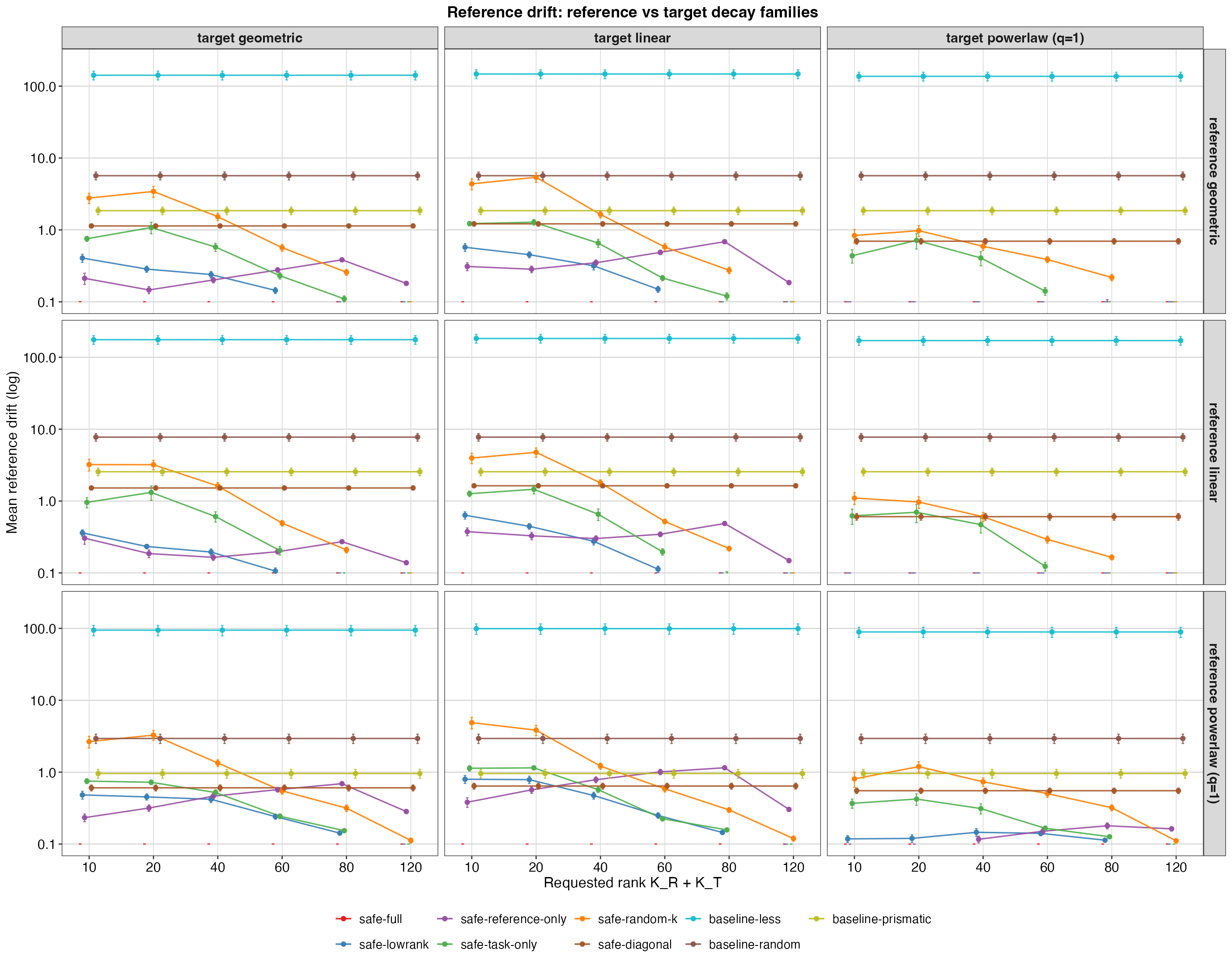}
    \caption{Off-target drift for the baselines and different variants of \safe{} (colors), per requested rank (x-axis) as a function of the spectra of $\mathbf\Sigma_{\Tdist}$ and $\F_{\Rdist}$. }
    \label{fig:tails_all_drift}
\end{figure}




Experiments~1--3 assume the population quantities are known. We now reintroduce the estimation problem: $g_{\Tdist}$ and $\mathbf{F}_{\Rdist}$ must be recovered from finite samples and represented in a shared low-rank basis. We use a synthetic high-dimensional teacher--student model in which the reference and target geometries can have different spectral structure.  In all runs, the ambient dimension is $d=200$, the true reference rank is $r=40$, the candidate pool contains $N_C=800$ examples, the target gradient is estimated from $N_T=32$ samples, the reference Fisher is estimated from $N_R=300$ samples, and the subset budget is $S=40$.  We generate the reference Fisher as $\mathbf{F}_{\Rdist}=\sigma^{-2}\mathbf{U}_R \diag(\lambda^R) \mathbf{U}_R^\top$ and the target covariance as $\mathbf{\Sigma}_{\Tdist} = \mathbf{U}_T \diag(\lambda^T) \mathbf{U}_T^\top$, where $\mathbf{U}_R$ and $\mathbf{U}_T$ are independent random orthonormal bases.  We sweep both the reference and target spectra over geometric, linear, and power-law decay, and in a denser variant we sweep power-law exponents $(q_R,q_T)\in\{0.5,0.75,1.0,1.5,2.0\}^2$.  For each spectrum pair, we compare \safe-full, \safe-low-rank, \safe-reference-only, \safe-task-only, \safe-random-$K$, \safe-diagonal, LESS, random, and Prismatic over requested working ranks $K_R+K_T\in\{10,20,40,60,80,120\}$.  All \safe{} methods are run under a fixed reference-drift budget $\rho=1$, and the realized subset update is always the SGD-style average $\Delta\theta(\mathcal S)=-(\eta/S)\sum_{i\in \mathcal S} g_i$.  We report the exact realized reference drift $\frac{1}{2}\Delta\theta^\top \mathbf{F}_{\Rdist} \Delta\theta$, the exact target gain, and the absolute regret relative to \safe{} full, averaged over $10$ random seeds and $3$ target regimes (for a total of $30$ instances per spectrum pair), with standard-error bars shown in all plots.

\subsection{Additional details on experiments}\label{appsec:real_exp_app}
Unless a setting specifies a separate evaluation set, experiments use the following configuration. The candidate pool $\mathcal{D}_{\mathrm{train}}$ is split into a $3\%$ warmup subset used
to estimate the Adam preconditioner $\mathbf{P}$ and a remaining $97\%$ from which training points are selected. The target set $\mathcal{D}_{\mathrm{tar}}$ is divided into three parts: one for computing the target gradient $\widehat{g}_T$, one for validation, and one as a held-out test set. The reference set $\mathcal{D}_{\mathrm{ref}}$ is split evenly between estimating $\widehat{\F}_R$ and evaluating reference drift. Each selected subset is then used for fine-tuning with the LoRA training recipe in Table~\ref{tab:datasets} using dropout $0.1$, target modules \texttt{\{q,k,v,o\}}, Adam update with learning rate $2\times 10^{-4}$, $1$ epoch, effective batch size $32$, and sequence length $4096$. Gradients for selection and model updates are computed over LoRA adapters~\citep{hu2022lora} applied to the attention modules. 

Throughout this section, we write \textit{reference Fisher cost} to denote the predicted drift $\Dfish$ and \textit{held-out KL} to denote the realized drift $\Dkl$.

\paragraph{Compute Resources}
For model fine-tuning, we use Huggingface Accelerate ~\citep{accelerate} and utilize Zero3 optimizer in DeepSpeed on two A100 GPUs. 

\subsubsection{Baseline selectors}
\label{app:baselines}
We compare \safe{} against five baselines. Full fine-tunes on the entire candidate pool without selection, and Random uniformly subsamples the pool at the target budget. 

All gradient-based methods---LESS, Prismatic Synthesis, and \safe{}---operate on the same set of low-rank projected gradients computed in the shared working basis $\U_K$, ensuring that differences in selection performance reflect the selection criterion rather than gradient featurization. LESS~\citep{xia2024less} scores each candidate by the cosine similarity between its projected gradient and the projected target gradient, using the low-rank Adam-style preconditioner as the inner product metric. We adapt Prismatic Synthesis~\citep{jung2025prismatic} to promote diversity by
clustering candidate gradients via $k$-means and sampling inversely proportional to cluster size, with the number of clusters set to $10\%$ of the pool and $20$ Lloyd iterations. 

DSIR~\citep{xie2023dataselectionlanguagemodels} operates on hashed $n$-gram features rather than gradients, scoring candidates by the dot product between their hashed $n$-gram count vectors and the log importance ratio $\log p - \log q$, estimated from a $30\%$ subsample of the pool with $10{,}000$ hash buckets. 

\subsubsection{Setting 1: Code generation}
\label{app:setting1}
The goal is to improve Python programming ability while leaving the base model's behavior on stereotypes unchanged. The candidate pool is CodeAlpaca~\citep{codealpaca}, a 20K dataset of general-purpose programming instructions generated via self-instruct~\citep{selfinstruct}. The target set is HumanEval~\citep{chen2021codex}, a benchmark of 164 hand-written Python programming problems; we split these into 109 test, 27 validation, and 28 examples for computing the target gradient $\widehat{g}_T$. The reference distribution is defined by the intrasentence split of StereoSet~\citep{nadeem2020stereoset}, which presents each prompt with a stereotypical, an anti-stereotypical, and an unrelated completion. We retain the stereotypical and anti-stereotypical completions in equal proportion to construct a balanced reference set. This results in 4.21K samples which we split into half for reference Fisher estimation and for evaluating off-target drift. Since computing gradients over the full candidate pool is memory-intensive, the task-side basis $\widehat{\mathbf{U}}_{K_T}$ is estimated from $4{,}000$ randomly subsampled candidates. We ablate over reference ranks $K_R\in\{128,256,512\}$, with the residual ranks detailed in Appendix~\ref{app:K_ablation}, at selection budget 15\%; results are reported in Figure~\ref{fig:codealpaca_metrics_15} and \ref{fig:training_curves}. 

\paragraph{Evaluation metrics.}
For the target task, we evaluate code generation quality via pass@$k$ rates following the protocol of \cite{chen2021codex}: for each problem, $k$ candidate
solutions are sampled and the problem is considered solved if at least one passes all unit tests. We report pass@$1$ and pass@$10$ (mean and standard deviation) for all selectors in Table~\ref{tab:main_results}. For the reference distribution, we use StereoSet score, defined as the proportion of prompts for which the model assigns higher average token probability to the stereotypical completion than to the anti-stereotypical one \citep{nadeem2020stereoset}. A score of $50$ means that the stereotypical completion receives higher average token probability on half the prompts; this aggregate statistic alone does not establish that a model is unbiased. The base model scores $64.9$ on this metric. We additionally record the probability margin, the difference in average token probability between the stereotypical and anti-stereotypical
completions. All reference metrics are reported as changes from the base model.

\subsubsection{Setting 2: Improving medical knowledge}
\label{app:setting2}

The goal of this experiment is to improve medical question-answering ability of a model while leaving the base model's broader non-medical behavior unchanged. The candidate pool is MedMCQA~\cite{pmlr-v174-pal22a}, a large collection of multiple-choice medical exam questions. For the target set, we subsample 256 samples from MedQA~\cite{jin2021disease}, which consists of medical licensing-style questions. The reference set combines BBQ~\cite{parrish-etal-2022-bbq} and TruthfulQA~\cite{lin2022truthfulqameasuringmodelsmimic}, covering social bias and factual reliability respectively, forming a non-medical alignment distribution.

\paragraph{Evaluation metrics.}
We measure fine-tuning performance on the target task using MedQA multiple-choice accuracy, reporting development accuracy for model selection and held-out test accuracy for final evaluation. To assess off-target drift, we evaluate the fine-tuned model on several non-medical benchmarks drawn from the reference evaluation suite. On BBQ~\cite{parrish-etal-2022-bbq}, we report multiple-choice accuracy as a measure of robustness to social bias. On TruthfulQA~\cite{lin2022truthfulqameasuringmodelsmimic}, we report  token-level F1 to capture factual reliability under free-form generation. On IFEval~\cite{zhou2023instructionfollowingevaluationlargelanguage}, we report prompt accuracy and instruction accuracy, which measure whether the model satisfies explicit formatting and content constraints.  

\begin{figure}[t]
    \centering
    \includegraphics[width=\linewidth]{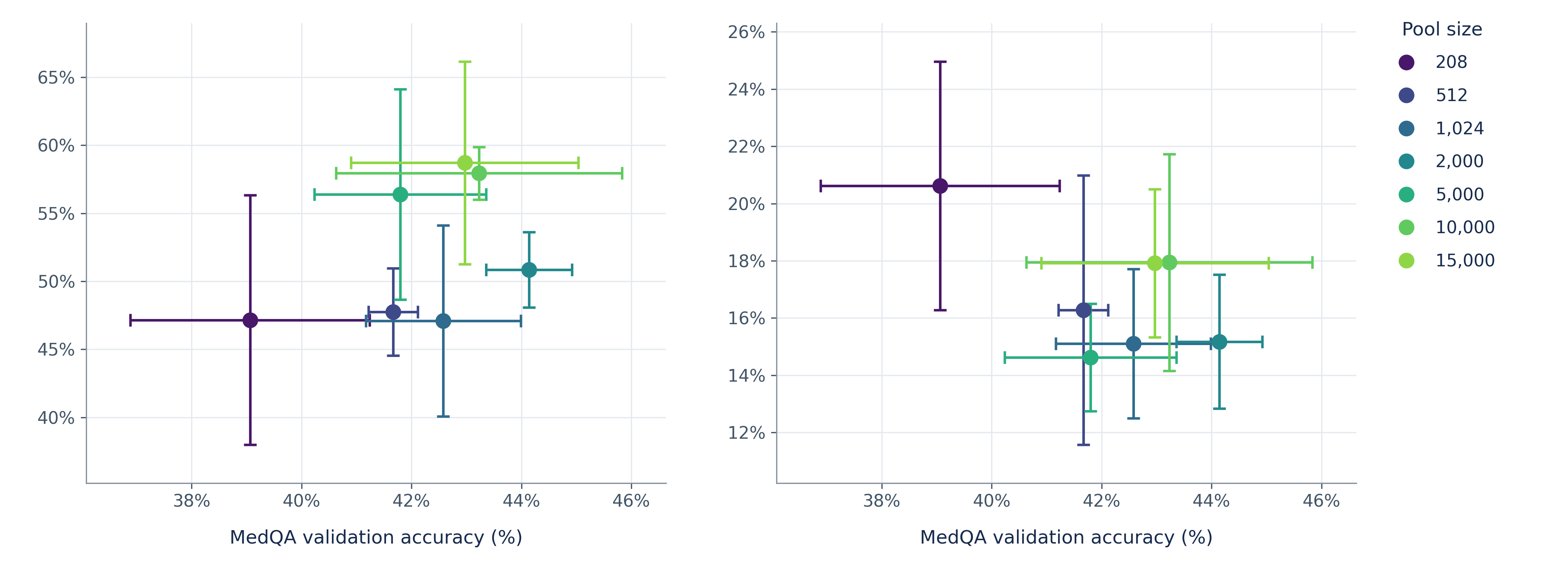}
    \caption{Pool-size effects on MedQA validation accuracy versus BBQ validation accuracy (left) and TruthfulQA validation token F1 (right) Training uses LoRA rank 16, $\alpha = 32$.}
    \label{fig:random_pool_res_set2}
\end{figure}

\paragraph{Additional experiments} To further show the importance of finetuning in this context, we compare the performance of random selection across various pool sizes. The results are shown in Figure~\ref{fig:random_pool_res_set2}. This figure highlights, in particular, that increasing the pool size does not automatically lead to better performance on the target task (here, MedQA), thereby making a case for the use of data selection in this context.

\subsubsection{Setting 3: Improving emotional support}
\label{app:setting3}
We seek to improve emotional-support dialogue performance while leaving the base model's behavior on mathematical reasoning unchanged. The candidate pool is EmpatheticDialogues~\cite{Rashkin2018TowardsEO}, a dataset of emotion-grounded open-domain conversations. The target set is ESConv~\cite{liu-etal-2021-towards}, a benchmark of emotional-support conversations. To define the protected reference distribution, we use GSM8K~\cite{cobbe2021gsm8k}, a dataset of grade-school math word problems that probes multi-step mathematical reasoning. 

\paragraph{Evaluation metrics.} We evaluate target-task performance on ESConv using  ROUGE-L F1, which is better suited than exact match for this open-ended generation setting. To measure drift away from the reference capability, we report accuracy on GSM8K and additionally on MMLU (zero-shot) as a broader out-of-domain measure of reasoning and knowledge preservation.

\begin{figure}[t]
    \centering
    \includegraphics[width=0.5\linewidth]{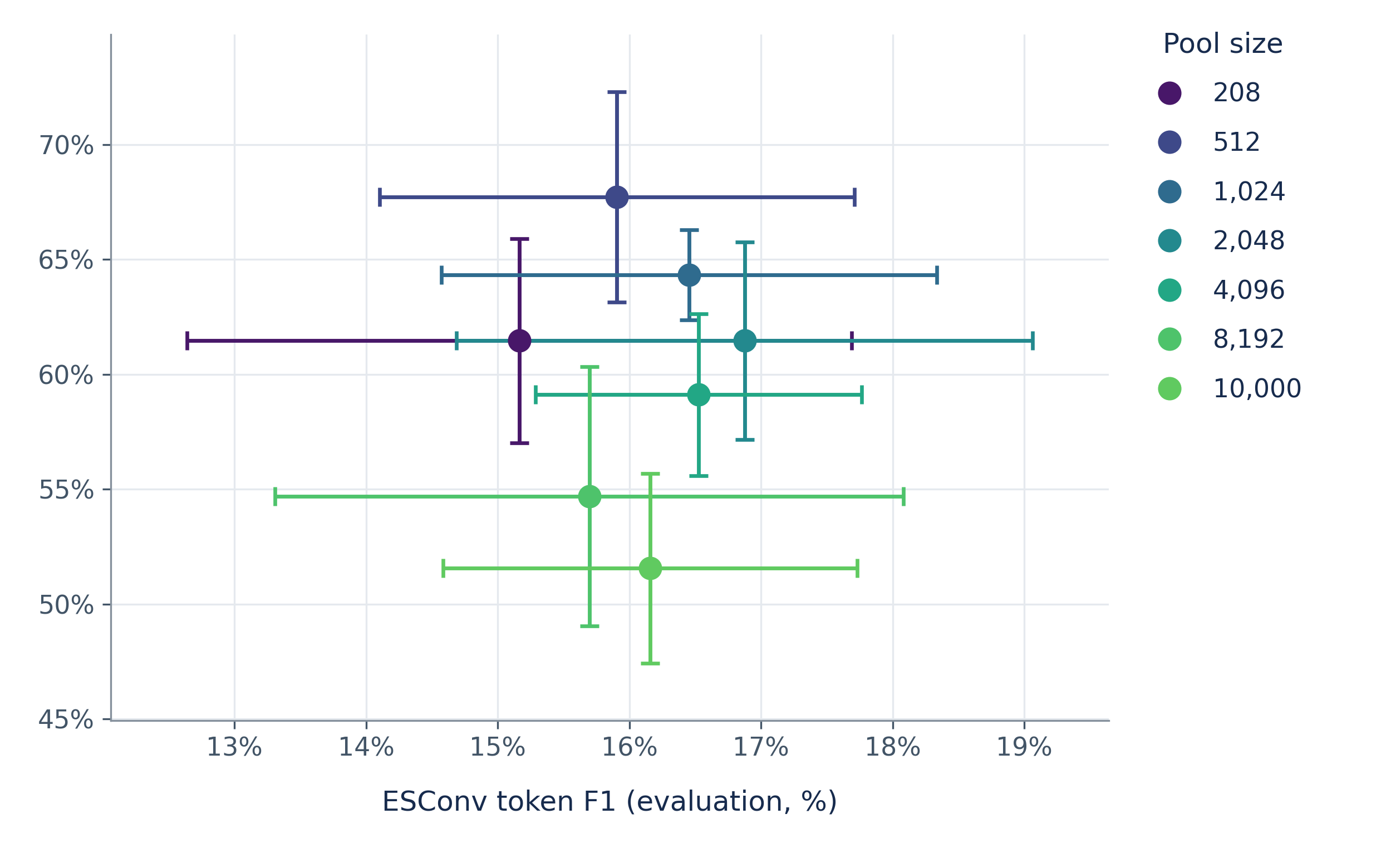}
    \caption{Pool-size effects on ESConv validation score versus GSM8K validation accuracy. Training uses LoRA rank 64, $\alpha = 128$.}
    \label{fig:random_pool_res_set3}
\end{figure}

\paragraph{Additional experiments and diagnostics.} As for setting 2, we compare the performance of random selection across various pool sizes. The results are shown in Figure~\ref{fig:random_pool_res_set3}. Again, we observe that increasing the pool size does not automatically lead to better performance on the target task.

\subsubsection{Setting 4: FLAN instruction tuning}
\label{app:setting4}
The goal is to improve instruction-following behavior while preserving the base model's performance on an unrelated knowledge-and-reasoning reference task. This setting is deliberately off-target: the selector is asked to identify useful instruction-tuning examples from a broad FLAN-v2~\citep{wang2023far} pool using a small IFEval-derived target set, while the reference geometry is estimated from MMLU. Thus, unlike settings where the target, candidate, and reference distributions are closely related, this experiment tests whether a selector can extract target-relevant instruction-following signal without unnecessarily perturbing models’ factual knowledge. The reported damping value for \safe{} is $\alpha=2\times10^{-4}$.

\paragraph{Evaluation metrics.}
For the target task, we evaluate instruction following on the official IFEval test set \cite{zhou2023instructionfollowingevaluationlargelanguage}, which contains $491$ prompts. Evaluation is performed with using OLMES ~\citep{gu2025olmesstandardlanguagemodel}. We report the four standard IFEval accuracies: prompt-level strict, prompt-level loose, instruction-level strict, and instruction-level loose. Prompt-level accuracy requires every constraint in a prompt to be satisfied simultaneously, whereas instruction-level accuracy scores each individual constraint separately and averages over constraints. These aggregate accuracies use different denominators. When the number of constraints varies across prompts, neither aggregate score must exceed the other; for example, one correct single-constraint prompt and one entirely incorrect three-constraint prompt give prompt accuracy $1/2$ but instruction accuracy $1/4$. Strict scoring applies the original verifier directly, while loose scoring applies light normalization such as lowercasing, whitespace cleanup, and simple punctuation handling before verification. The OLMES primary scores for IFEval is prompt-level strict and instruction strict accuracy, which we use as the main target metric in Table~\ref{tab:main_results}.

For the reference task, we evaluate MMLU on a fixed held-out $200$-item test slice. 
The chat-format mode evaluates MMLU as open-ended generation under the same chat template used during fine-tuning, making it sensitive to whether the fine-tuned model still emits clean single-letter answers such as \texttt{A}, \texttt{B}, \texttt{C}, or \texttt{D}. The loglikelihood mode instead compares
\begin{align*}
P_\theta(\texttt{A}\mid \text{prompt}),\,
P_\theta(\texttt{B}\mid \text{prompt}),\,
P_\theta(\texttt{C}\mid \text{prompt}),\,
P_\theta(\texttt{D}\mid \text{prompt})
\end{align*}
and selects the answer with maximum probability. This mode is less sensitive to chat-template formatting and provides a cleaner estimate of whether the model's underlying multiple-choice knowledge has changed. Since the purpose of the reference task is preservation rather than improvement. Values closest to zero indicate the least off-target movement from the base model.

\subsection{Entanglement Diagnostics}
\begin{figure}[t]
    \centering
    \includegraphics[width=1.0\linewidth]{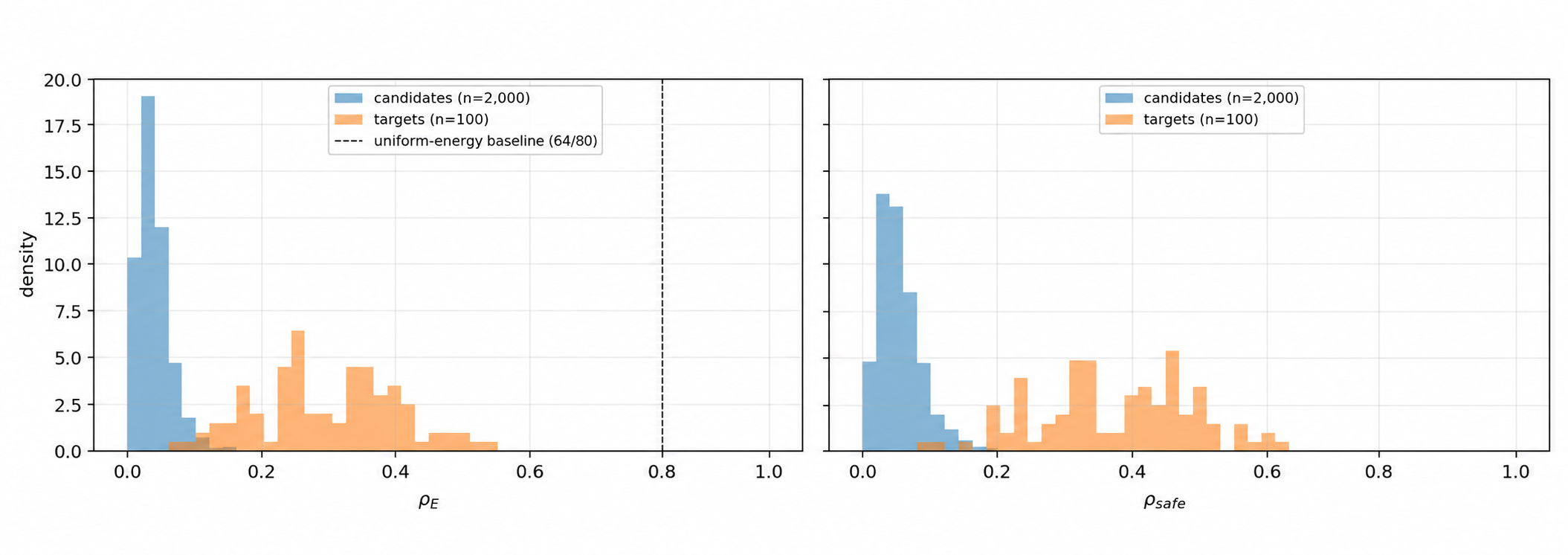}
    \caption{Reference-subspace entanglement diagnostics for Setting~3. We draw the density of $\rho_E$ and $\rho_{\mathrm{safe}}$ from \eqref{eq:reference-entanglement diagnostics} with $K_R=64$, and $K=80$.
    Values close to $0$ indicate smaller reference-block energy per coordinate relative to the residual block, and values close to $1$ indicate the reverse, under the respective weighting in \eqref{eq:reference-entanglement diagnostics}.}
    \label{fig:reference-entanglement diagnostics_setting3}
\end{figure}

We show histograms of the reference-entanglement diagnostics from Section~\ref{sec:reference-ent} for Settings 3 and 4. These ratios normalize each block by its dimension, so they compare energy per coordinate rather than the unnormalized fraction of gradient energy in the reference span. In Setting 3, the concentration near zero indicates weak reference-block energy per coordinate under the respective diagnostic weighting (Figure~\ref{fig:reference-entanglement diagnostics_setting3}). Neither these ratios nor $\pi_{Q_{\mathrm{ent}}}=0.08$ bounds the absolute reference cost of a selected update. Target gradients have larger diagnostic values than candidate gradients; one possible explanation is that a basis estimated from many more candidates than targets underrepresents target-specific residual directions.

\begin{figure}[t]
    \centering
    \includegraphics[width=0.9\linewidth]{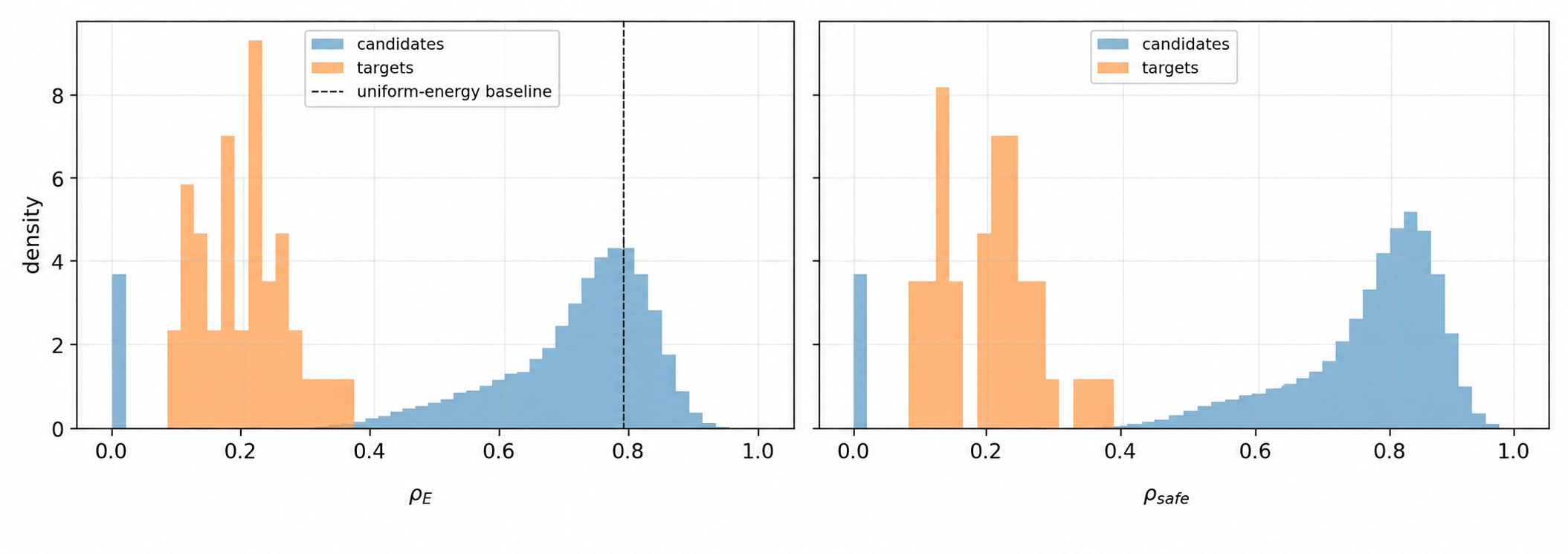}
    \caption{Reference-subspace entanglement diagnostics for Setting~4. We draw the density of $\rho_E$ and $\rho_{\mathrm{safe}}$ from \eqref{eq:reference-entanglement diagnostics} with $K_R=128$, and $K=160$. The displayed dashed line marks $K_R/K=128/160=0.8$, which is not the equal-coordinate-energy baseline for these dimension-normalized diagnostics. Those baselines are $1/2$ for $\rho_E$ and $\bar\lambda_R/(\bar\lambda_R+\alpha)$ for $\rho_{\mathrm{safe}}$, where $\bar\lambda_R$ is the mean retained reference eigenvalue (Section~\ref{sec:reference-ent}).
}
    \label{fig:entanglement_plot_setting4}
\end{figure}

Setting 4 is substantially more entangled, as displayed in Figure~\ref{fig:entanglement_plot_setting4} and the entanglement scores of $\pi_{Q_{\mathrm{ent}}}=0.26$ and $\pi_{Q_{\mathrm{clean}}}=0.22$. The quadrant frequencies indicate positive residual-task alignment for many candidates with positive reference-block alignment. They do not determine total target gain or establish a necessary loss in MMLU accuracy.

\subsection{Comparison Against Behavior-Preservation Baselines}
\label{app:preservation}

\begin{table}[t]
\centering
\small
\resizebox{\textwidth}{!}{%
\begin{tabular}{llcccc}
\toprule
\multicolumn{6}{l}{\textit{(a) Rehearsal (replay) baselines --- Setting 1}} \\
\midrule
Method & Replay \% & $\Delta$ HumanEval@1 $\uparrow$ & $\Delta$ HumanEval@10 $\uparrow$ & \multicolumn{2}{c}{Held-out KL ($\Dkl$) $\downarrow$} \\
\midrule
DSIR       & 10 & $0.0516 \pm 0.0096$ & $0.1362 \pm 0.0040$ & \multicolumn{2}{c}{$0.0190 \pm 0.0030$} \\
DSIR       & 20 & $0.0429 \pm 0.0108$ & $0.1260 \pm 0.0090$ & \multicolumn{2}{c}{$0.0223 \pm 0.0007$} \\
LESS       & 10 & $0.0217 \pm 0.0056$ & $0.0870 \pm 0.0090$ & \multicolumn{2}{c}{$0.0152 \pm 0.0017$} \\
LESS       & 20 & $0.0243 \pm 0.0082$ & $0.0995 \pm 0.0153$ & \multicolumn{2}{c}{$0.0181 \pm 0.0028$} \\
Prismatic  & 10 & $0.0185 \pm 0.0009$ & $0.0793 \pm 0.0060$ & \multicolumn{2}{c}{$0.0197 \pm 0.0020$} \\
Prismatic  & 20 & $0.0134 \pm 0.0038$ & $0.0750 \pm 0.0130$ & \multicolumn{2}{c}{$0.0230 \pm 0.0040$} \\
\textsc{Safe-Drift} & 0 & $\mathbf{0.132 \pm 0.004}$ & $\mathbf{0.183 \pm 0.023}$ & \multicolumn{2}{c}{$\mathbf{0.0127 \pm 0.0006}$} \\
\midrule
\multicolumn{6}{l}{\textit{(b) KL-regularized selection baselines --- Settings 2 and 3}} \\
\midrule
Setting & Method & Selected $\lambda$ & $\Delta$ test target $\uparrow$ & $\Delta$ test reference $\uparrow$ & Held-out KL ($\Dkl$) $\downarrow$ \\
\midrule
\multirow{4}{*}{2}
 & DSIR                 & ---                 & $+3.59 \pm 2.04$ & $-8.10 \pm 3.58$ & $4.210 \pm 0.528$ \\
 & LESS                 & $\lambda = 0.1$ & $+2.08 \pm 0.60$ & $+0.73 \pm 1.26$ & $1.974 \pm 0.224$ \\
 & Prismatic            & $\lambda = 1$ & $+2.99 \pm 1.63$ & $+0.89 \pm 2.07$ & $\mathbf{1.534 \pm 0.025}$ \\
 & \textsc{Safe-Drift}  & ---                 & $+3.36 \pm 1.42$ & $-3.50 \pm 3.24$ & $4.003 \pm 0.467$ \\
\midrule
\multirow{4}{*}{3}
 & DSIR                 & ---                 & $+4.65 \pm 1.18$ & -- & $\mathbf{0.317 \pm 0.025}$ \\
 & LESS                 & $\lambda = 0$       & $\mathbf{+6.77 \pm 1.18}$ & -- & $0.368 \pm 0.039$ \\
 & Prismatic            & $\lambda = 0$       & $+4.55 \pm 0.97$ & -- & $0.364 \pm 0.023$ \\
 & \textsc{Safe-Drift}  & ---                 & $+4.51 \pm 1.10$ & -- & $0.391 \pm 0.011$ \\
\bottomrule
\end{tabular}%
}
\caption{\textit{Comparison against behavior-preservation baselines.} \textbf{(a)} Rehearsal: reference examples, drawn from the Fisher-estimation split and disjoint from the drift-evaluation set, are mixed into training at 10\% and 20\% of the budget. \textsc{Safe-Drift} never trains on the reference set. \textbf{(b)} LESS and Prismatic augmented with a reference KL penalty with $\lambda$ chosen on validation set by the geometric mean of target and reference accuracy. We show means and standard deviations of differences from each setting's base model. For Setting~3, reference-score changes are reported in Table~\ref{tab:main_results}; panel~(b) reports target changes and KL for this comparison. }\label{tab:preservation_baselines}
\end{table}

We provide the complementary results of replay-based methods of DSIR, LESS, and Prismatic of Section~\ref{sec:main} in Table~\ref{tab:preservation_baselines}. We additionally experiment with augmenting the reference KL-divergence penalty to the SFT loss:
$$\mathcal{L}_{\mathrm{SFT}}(\theta)+\lambda \ \mathbb E_{u \sim \widehat{\mathcal R}_{\mathrm{seq}}}\Big[\frac{1}{|u|} \sum_{t=1}^{|u|} D_{\mathrm{KL}}\big(p_{\theta_0}(\cdot | u_{<t}) \Vert p_{\theta}(\cdot | u_{<t})\big) \Big],$$
where $\widehat{\mathcal R}_{\mathrm{seq}}$ denotes the empirical distribution of nonempty observed reference token sequences, distinct from the prompt distribution in \eqref{eq:off-target-drift}. The displayed regularizer averages per-sequence token means; for unequal sequence lengths this differs from a corpus-wide token-weighted mean. A fixed observed-prefix average is not generally sequence-level KL. The sequence-KL chain rule instead requires conditional KL terms averaged under the base model's prefix distribution, without per-sequence length normalization and with termination accounted for when sequence lengths vary.
We sweep over $\lambda \in \{0, 0.01, 0.1, 1\}$ and select the best $\lambda$ on a validation set using the geometric mean of target and reference performance. 

When replay increases from 10\% to 20\%, mean target performance falls for DSIR and Prismatic but rises modestly for LESS; held-out KL rises for all three. For the KL penalty, the reported Spearman correlations between positive $\lambda$ and held-out KL are $-0.949$ and $-0.896$ for LESS and Prismatic in Setting 2, and $-0.949$ for both in Setting 3. In Setting 2, regularized Prismatic has the lowest mean KL, whereas \safe{} has comparable target gain to DSIR with less reference-score degradation; its KL is $4.003$ versus DSIR's $4.210$, a modest reduction. In Setting 3, validation selects $\lambda=0$ for LESS and Prismatic. \safe{} has the largest mean reference-score improvement and smallest reference-score standard deviation, but not the lowest KL. These metrics therefore capture distinct aspects of reference behavior.

\subsection[Choosing the Drift Budget rho]{Choosing the Drift Budget $\rho$}
\label{app:rho}

\begin{table}[t]
\centering
\small
\begin{tabular}{llcccccc}
\toprule
Setting & Optimizer & \begin{tabular}[c]{@{}c@{}}Hyperparam.\\ configs\end{tabular} & \begin{tabular}[c]{@{}c@{}}Seeds\\ per cell\end{tabular} & \begin{tabular}[c]{@{}c@{}}$\rho$ vs.\\ target\end{tabular} & \begin{tabular}[c]{@{}c@{}}$\rho$ vs.\\ $\Dtask$ \end{tabular} & \begin{tabular}[c]{@{}c@{}}Target vs.\\ $\Dtask$ \end{tabular} & \begin{tabular}[c]{@{}c@{}}$\Dfish$ vs.\\ $\Dkl$\end{tabular} \\
\midrule
Setting 2 & Adam & 9  & 5 & $0.753$ & $0.483$  & $0.561$  & $0.967$ \\
Setting 3 & Adam & 18 & 5 & $0.672$ & $-0.735$ & $-0.820$ & $0.670$ \\
\bottomrule
\end{tabular}
\caption{\textit{Spearman rank correlations for the drift budget $\rho$.} $\Dtask$ denotes reference-score degradation, measured as base minus fine-tuned score. The negative Setting~3 correlations indicate that target and reference accuracy increase together---consistent with the low target--reference entanglement of that setting. The final column correlates the predicted drift $\Dfish$ with the realized drift $\Dkl$ over the same runs.
\label{tab:rho_correlations}}
\end{table}
\begin{figure}
    \centering
    \includegraphics[width=\linewidth]{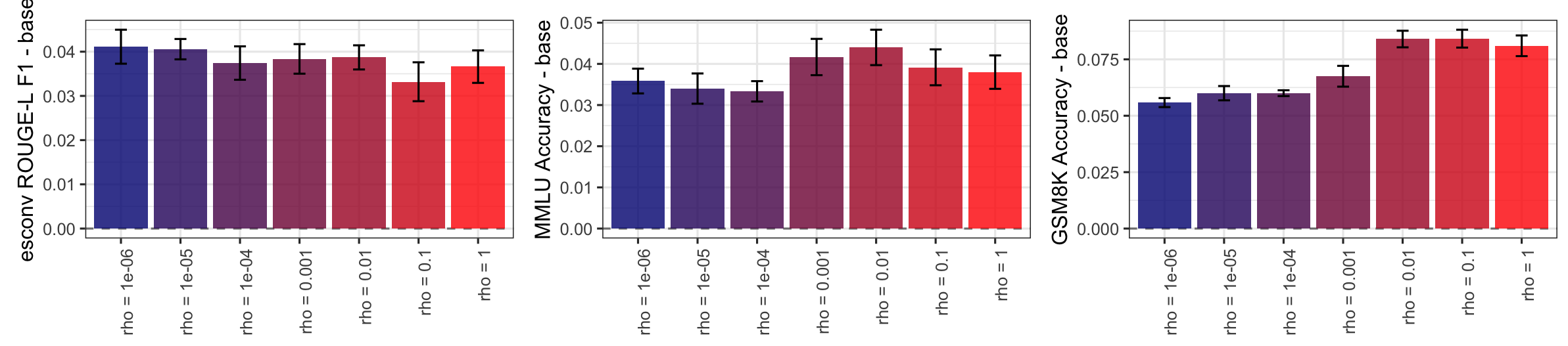}
    \caption{Effect of $\rho$ in ESconv (Setting 3) }
    \label{fig:effect_rho}
\end{figure}

The drift budget $\rho$ is the main user-facing control in \safe: tighter budgets avoid high-cost reference directions, looser ones let the selector pursue target
gain. Section~\ref{sec:main} sweeps $\rho$ over $\{10^{-5},10^{-2},10^{0}\}$ in
Setting 2; Table~\ref{tab:rho_correlations} reports rank correlations over a
larger sweep of nine hyperparameter configurations in Setting 2 and eighteen in Setting 3, five seeds each. Increasing $\rho$ is positively associated with target performance (Spearman correlation $0.753$ and $0.672$), but its association with reference drift
reverses sign between settings: $0.483$ in Setting 2, where a looser budget costs reference accuracy, against $-0.735$ in Setting 3, where it improves both axes at once (Figure~\ref{fig:effect_rho}). $\rho$ is thus a trade-off parameter, not a universally monotone regularizer. The sweep also shows an empirical association between the predicted Fisher cost and held-out KL: $\Dfish$ and $\Dkl$ correlate at $\rho_s = 0.967$ and $0.670$. Figures~\ref{fig:ablation_geometry_medqa_gradients} and~\ref{fig:ablation_geometry_esconv_gradients} provide complementary candidate-level diagnostics; they do not certify the KL of the final model.

We recommend tuning $\rho$ on validation target and reference-drift sets. For a local oracle update satisfying the population quadratic budget and $\|\Delta\theta\|\le\epsilon<r_0$, the assumptions of Proposition~\ref{prop:local-kl} give $D_\Rdist\le\rho+M\epsilon^3/6$. Pinsker's inequality and Jensen's inequality then bound the reference-input-averaged change in any conditional event probability by $\sqrt{(\rho+M\epsilon^3/6)/2}$. This is neither a pointwise guarantee for every prompt nor a guarantee for the selected subset or final SFT model. Empirical Fisher estimation, truncation, subset-matching error, and the training trajectory require separate control; hence validation determines the practical meaning of a chosen $\rho$.

\subsection{Robustness across base models, composite references, and fine-tuning methods}\label{app:ablation}

\subsubsection{Robustness across base models}
\label{app:model_ablation}
\begin{table}[t]
\centering
\small
\setlength{\tabcolsep}{4pt}
\resizebox{\textwidth}{!}{%
\begin{tabular}{llcccccc}
\toprule
Model & Evaluation & Random & DSIR & LESS & Prismatic & \textsc{Safe-Drift} & Full Data \\
\midrule
\multirow{3}{*}{Qwen2.5-1.5B}
 & HumanEval pass@1 $\uparrow$ & $0.128 \pm 0.023$ & $0.133 \pm 0.003$ & $0.157 \pm 0.083$ & $0.165 \pm 0.018$ & $\mathbf{0.283 \pm 0.020}$ & $0.327 \pm 0.010$ \\
 & $\log_{10} \Dfish$ $\downarrow$ & $-5.58 \pm 0.04$ & $-5.06 \pm 0.06$ & $-5.48 \pm 0.06$ & $-5.64 \pm 0.04$ & $\mathbf{-6.29 \pm 0.04}$ & $-5.60 \pm 0.05$ \\
 & $\Dkl$ $\downarrow$ & $0.0049 \pm 0.0001$ & $0.0055 \pm 0.0006$ & $0.0057 \pm 0.0005$ & $\mathbf{0.0044 \pm 0.0001}$ & $\mathbf{0.0044 \pm 0.0003}$ & $0.0110 \pm 0.0001$ \\
\midrule
\multirow{3}{*}{Qwen2.5-3B}
 & HumanEval pass@1 $\uparrow$ & $0.162 \pm 0.028$ & $0.175 \pm 0.032$ & $0.193 \pm 0.017$ & $0.129 \pm 0.049$ & $\mathbf{0.242 \pm 0.053}$ & $0.248 \pm 0.026$ \\
 & $\log_{10} \Dfish$ $\downarrow$ & $-6.16 \pm 0.03$ & $-5.65 \pm 0.06$ & $-6.13 \pm 0.03$ & $-6.22 \pm 0.01$ & $\mathbf{-6.89 \pm 0.03}$ & $-6.25 \pm 0.06$ \\
 & $\Dkl$ $\downarrow$ & $\mathbf{0.0031 \pm 0.0003}$ & $0.0038 \pm 0.0004$ & $0.0042 \pm 0.0008$ & $\mathbf{0.0031 \pm 0.0001}$ & $0.0036 \pm 0.0002$ & $0.0091 \pm 0.0008$ \\
\midrule
\multirow{3}{*}{Qwen2.5-7B}
 & HumanEval pass@1 $\uparrow$ & $0.117 \pm 0.041$ & $0.089 \pm 0.045$ & $\mathbf{0.180 \pm 0.055}$ & $0.103 \pm 0.053$ & $0.159 \pm 0.044$ & $0.269 \pm 0.040$ \\
 & $\log_{10} \Dfish$ $\downarrow$ & $-6.89 \pm 0.02$ & $-6.41 \pm 0.04$ & $-6.81 \pm 0.02$ & $-6.89 \pm 0.02$ & $\mathbf{-7.62 \pm 0.03}$ & $-6.88 \pm 0.13$ \\
 & $\Dkl$ $\downarrow$ & $0.0038 \pm 0.0003$ & $0.0038 \pm 0.0002$ & $0.0039 \pm 0.0006$ & $\mathbf{0.0030 \pm 0.0004}$ & $0.0037 \pm 0.0001$ & $0.0076 \pm 0.0010$ \\
\midrule
\multirow{3}{*}{OLMo-2-7B}
 & HumanEval pass@1 $\uparrow$ & $0.130 \pm 0.023$ & $\mathbf{0.148 \pm 0.021}$ & $0.117 \pm 0.026$ & $0.135 \pm 0.013$ & $0.125 \pm 0.033$ & $0.123 \pm 0.010$ \\
 & $\log_{10} \Dfish$ $\downarrow$ & $-5.98 \pm 0.08$ & $-5.38 \pm 0.06$ & $-5.72 \pm 0.13$ & $-5.90 \pm 0.08$ & $\mathbf{-7.05 \pm 0.15}$ & $-5.95 \pm 0.07$ \\
 & $\Dkl$ $\downarrow$ & $0.0106 \pm 0.0021$ & $0.0110 \pm 0.0005$ & $\mathbf{0.0098 \pm 0.0006}$ & $0.0129 \pm 0.0031$ & $0.0127 \pm 0.0006$ & $0.0209 \pm 0.0021$ \\
\midrule
\multirow{3}{*}{OLMo-2-13B}
 & HumanEval pass@1 $\uparrow$ & $0.180 \pm 0.043$ & $0.182 \pm 0.029$ & $0.154 \pm 0.025$ & $\mathbf{0.194 \pm 0.036}$ & $0.161 \pm 0.031$ & $0.185 \pm 0.029$ \\
 & $\log_{10} \Dfish$ $\downarrow$ & $-6.88 \pm 0.17$ & $-6.23 \pm 0.08$ & $-6.71 \pm 0.06$ & $-6.69 \pm 0.10$ & $\mathbf{-7.95 \pm 0.07}$ & $-6.83 \pm 0.07$ \\
 & $\Dkl$ $\downarrow$ & $0.0097 \pm 0.0013$ & $0.0097 \pm 0.0027$ & $0.0087 \pm 0.0011$ & $0.0095 \pm 0.0010$ & $\mathbf{0.0084 \pm 0.0013}$ & $0.0122 \pm 0.0012$ \\
\bottomrule
\end{tabular}%
}
\caption{\textit{Base model ablation} (Setting~1). Two model families and five configurations; 15\% selection budget and following configurations in Table~\ref{tab:datasets}. We measure reference Fisher cost $\Dfish$ and the held-out reference KL $\Dkl$, each corresponding to predicted and realized off-target drift. Bold denotes the best selection method per row; ``Full Data'' is shown for reference and not ranked.
\label{tab:model_ablation_full}}
\end{table}

We repeat Setting 1 across two model families and five configurations---Qwen2.5 at 1.5B, 3B and 7B, and OLMo-2 at 7B and 13B---at a 15\% selection budget with three seeds. Table~\ref{tab:model_ablation_full} reports the full numbers
summarized in Figure~\ref{fig:costgain_kl}(a--c), including fine-tuning on the
full candidate pool. Safe-Drift attains the lowest reported log Fisher cost in every configuration, reducing $\Dkl$ by 31.1–60.0\% relative to full-data fine-tuning. Realized drift $\Dkl$ is less uniform: \safe{} is lowest at Qwen2.5-1.5B and OLMo-2-13B, and competitive but not best elsewhere. Target performance shows a similar split, with \safe{} leading on the
smaller Qwen models but not the larger ones---trading target gain for drift control where the target directions are entangled with the reference geometry.

\subsubsection{Composite Reference Distributions}
\label{app:mixtures}
\safe{} constructs its local drift surrogate from a reference distribution that the practitioner specifies. A natural question is whether the method degrades when the reference is made more complex. We therefore rerun Setting 2 with the reference constructed from several benchmarks at once: writing $\mathrm{B}$ for BBQ, $\mathrm{T}$ for TruthfulQA and $\mathrm{K}$ for MMLU (non-medical knowledge), we set the composition as $\mathrm{B}\!+\!\mathrm{T}$, $\mathrm{B}\!+\!\mathrm{K}$, $\mathrm{T}\!+\!\mathrm{K}$ and all three jointly. Every composition optimizes the same target (MedQA accuracy).

\begin{table}[H]
\centering
\small
\setlength{\tabcolsep}{5pt}
\resizebox{\textwidth}{!}{%
\begin{tabular}{llccccccc}
\toprule
Composition & Method & $\Delta$ Target & $\Delta$ B & $\Delta$ T & $\Delta$ K & $\Delta$ Ref GM & $\Delta$ All-ref GM & $\Dkl$ $\downarrow$ \\
\midrule
\multirow{5}{*}{$\mathrm{B\!+\!T}$}
 & \textsc{Safe-Drift} & $\mathbf{+0.090}$ & $\bf{+0.085}$ & $-0.105$ & $+0.070$ & $\mathbf{-0.062}$ & $\mathbf{-0.032}$ & $\mathbf{5.17}$ \\
 & Random    & $+0.063$ & $-0.049$ & $-0.107$ & $+0.061$ & $-0.100$ & $-0.065$ & $5.47$ \\
 & DSIR      & $+0.064$ & $-0.054$ & $-0.125$ & $\textbf{+0.078}$ & $-0.117$ & $-0.078$ & $5.49$ \\
 & LESS      & $+0.022$ & $+0.028$ & $\textbf{-0.100}$ & $+0.033$ & $-0.072$ & $-0.049$ & $5.54$ \\
 & Prismatic & $+0.078$ & $-0.042$ & $-0.107$ & $+0.055$ & $-0.098$ & $-0.065$ & $5.46$ \\
\midrule
\multirow{5}{*}{$\mathrm{B\!+\!K}$}
 & \textsc{Safe-Drift} & $\mathbf{+0.086}$ & $\textbf{+0.066}$ & $\textbf{-0.093}$ & $+0.051$ & $\mathbf{+0.058}$ & $\mathbf{-0.031}$ & $\mathbf{5.16}$ \\
 & Random    & $+0.063$ & $-0.049$ & $-0.107$ & $+0.061$ & $+0.004$ & $-0.065$ & $5.47$ \\
 & DSIR      & $+0.064$ & $-0.054$ & $-0.125$ & $\textbf{+0.078}$ & $+0.009$ & $-0.078$ & $5.49$ \\
 & LESS      & $+0.022$ & $+0.028$ & $-0.100$ & $+0.033$ & $+0.030$ & $-0.049$ & $5.54$ \\
 & Prismatic & $+0.078$ & $-0.042$ & $-0.107$ & $+0.055$ & $+0.005$ & $-0.065$ & $5.46$ \\
\midrule
\multirow{5}{*}{$\mathrm{T\!+\!K}$}
 & \textsc{Safe-Drift} & $\mathbf{+0.094}$ & $\textbf{+0.046}$ & $-0.103$ & $+0.063$ & $\mathbf{-0.066}$ & $\mathbf{-0.041}$ & $\mathbf{5.29}$ \\
 & Random    & $+0.063$ & $-0.049$ & $-0.107$ & $+0.061$ & $-0.069$ & $-0.065$ & $5.47$ \\
 & DSIR      & $+0.064$ & $-0.054$ & $-0.125$ & $+0.078$ & $-0.083$ & $-0.078$ & $5.49$ \\
 & LESS      & $+0.022$ & $+0.028$ & $-0.100$ & $+0.033$ & $-0.070$ & $-0.049$ & $5.54$ \\
 & Prismatic & $+0.078$ & $-0.042$ & $-0.107$ & $+0.055$ & $-0.071$ & $-0.065$ & $5.46$ \\
\midrule
\multirow{5}{*}{$\mathrm{B\!+\!T\!+\!K}$}
 & \textsc{Safe-Drift} & $\mathbf{+0.078}$ & $+0.038$ & $-0.098$ & $+0.020$ & $\mathbf{-0.048}$ & $\mathbf{-0.048}$ & $\mathbf{5.13}$ \\
 & Random    & $+0.063$ & $-0.049$ & $-0.107$ & $+0.061$ & $-0.065$ & $-0.065$ & $5.47$ \\
 & DSIR      & $+0.064$ & $-0.054$ & $-0.125$ & $+0.078$ & $-0.078$ & $-0.078$ & $5.49$ \\
 & LESS      & $+0.022$ & $+0.028$ & $-0.100$ & $+0.033$ & $-0.049$ & $-0.049$ & $5.54$ \\
 & Prismatic & $\mathbf{+0.078}$ & $-0.042$ & $-0.107$ & $+0.055$ & $-0.065$ & $-0.065$ & $5.46$ \\
\bottomrule
\end{tabular}%
}
\caption{\textit{Composite reference distributions} (Setting~2). $\mathrm{B}$ = BBQ, $\mathrm{T}$ = TruthfulQA, $\mathrm{K}$ = MMLU (non-medical knowledge). Each composition names the reference tasks used to construct the \textsc{Safe-Drift} constraint; omitted tasks are still evaluated but not protected. All other baselines optimize the same target (MedQA accuracy). $\Delta$ Ref GM is the fine-tuned-minus-base change in the geometric mean of scores over the \emph{protected} reference tasks; $\Delta$ All-ref GM is the corresponding change over all three. The last column reports held-out reference KL $\Dkl$, whose base-model value is zero.
}\label{tab:offtarget_mixtures}
\end{table}

Table~\ref{tab:offtarget_mixtures} reports the results. Across all four compositions, \safe{} has the lowest held-out reference KL and the largest change in the geometric mean of protected reference scores. It matches or exceeds the best baseline target gain, with gains ranging from $+0.078$ to $+0.094$ across compositions. These results do not establish monotonic improvement as more references are added or minimum absolute movement on every reference task. The geometric mean over all three reference tasks also improves relative to the baselines, suggesting some shared preservation benefit, although individual benchmark scores exhibit trade-offs.

\subsubsection{LoRA Versus Full Supervised Fine-Tuning}
\label{app:lora_full}

\begin{table}[t]
\centering
\small
\begin{tabular}{lcc}
\toprule
Update & Test MedQA accuracy & Test reference accuracy \\
\midrule
LoRA     & $44.30\% \pm 1.42\%$ & $36.76\% \pm 3.58\%$ \\
Full SFT & $37.50\% \pm 3.87\%$ & $\mathbf{38.92\% \pm 2.42\%}$ \\
\bottomrule
\end{tabular}
\caption{\textit{LoRA versus full supervised fine-tuning} (Setting~2).
\label{tab:lora_vs_full}}
\end{table}

LoRA defines the trainable parameter space throughout our experiments, because it has become the dominant approach to practical parameter-efficient instruction tuning. There is no conceptual obstacle to running \safe{} over a full SFT parameterization---only a computational one, since per-example gradients must then be formed in the full parameter space.

Table~\ref{tab:lora_vs_full} compares the two on Setting 2. LoRA has the higher reported mean target accuracy, whereas full SFT has the higher reported mean reference accuracy. Higher reference accuracy alone establishes neither smaller distributional drift nor smaller movement from the base model. The table does not isolate the effect of update-subspace size or establish a general ordering between the two parameterizations.

\subsection{Ablations across method design}

\subsubsection{Using reference-induced norm vs Euclidean norm in the selector}\label{app:ablation:norm}

\begin{figure}[t]
    \centering
    \includegraphics[width=\linewidth]{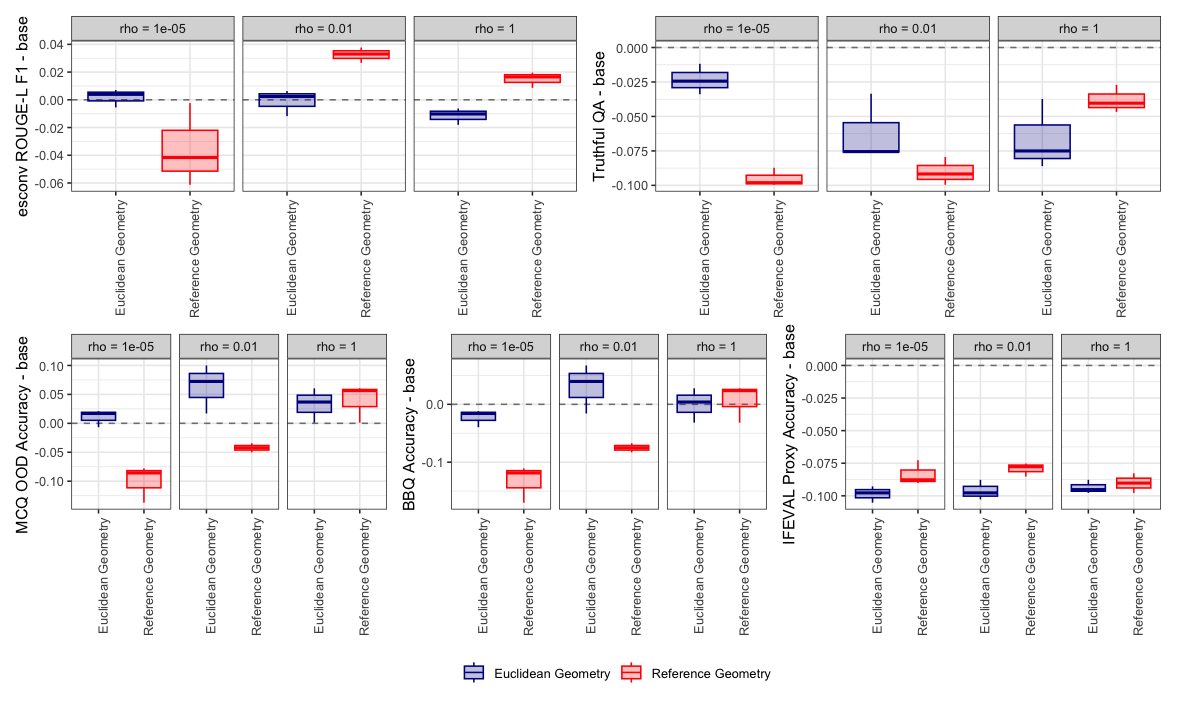}
    \caption{Target gain and off-target drift $\Dtask$ of \safe{} using Euclidean vs reference Fisher-induced norm objective in MedQA experiment (Setting 2). We plot the difference from the base model across drift budget $\rho$. }
    \label{fig:ablation_geometry_medqa}
\end{figure}

\begin{figure}[t]
    \centering
    \includegraphics[width=0.9\linewidth]{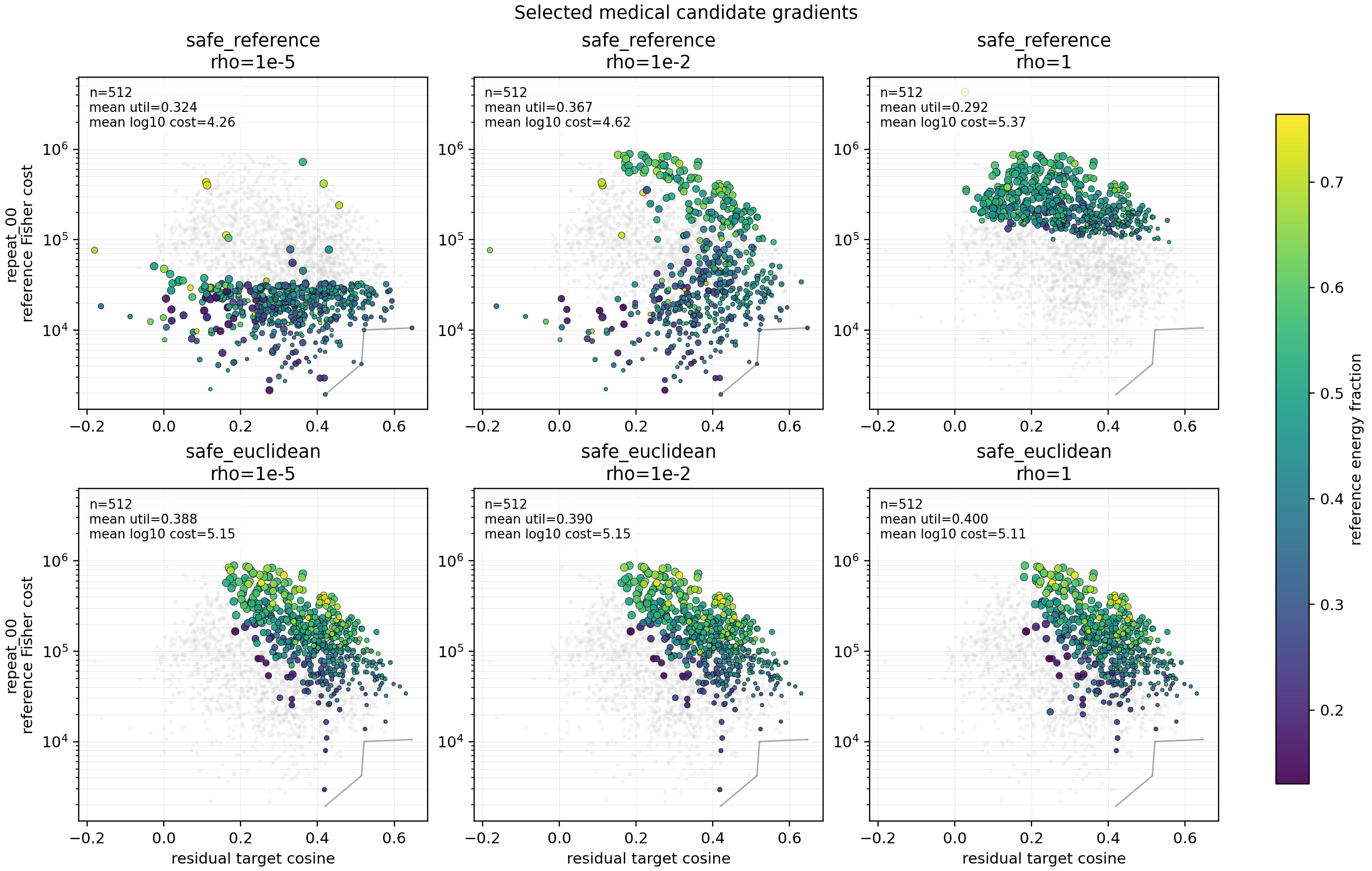}
    \caption{Selected points plotted on target gain versus realized off-target drift $\Dkl$ from \safe{} using Euclidean vs reference Fisher-induced norm objective in MedQA experiment (Setting 2).}\label{fig:ablation_geometry_medqa_gradients}
\end{figure}

\begin{figure}[t]
    \centering
    \includegraphics[width=\linewidth]{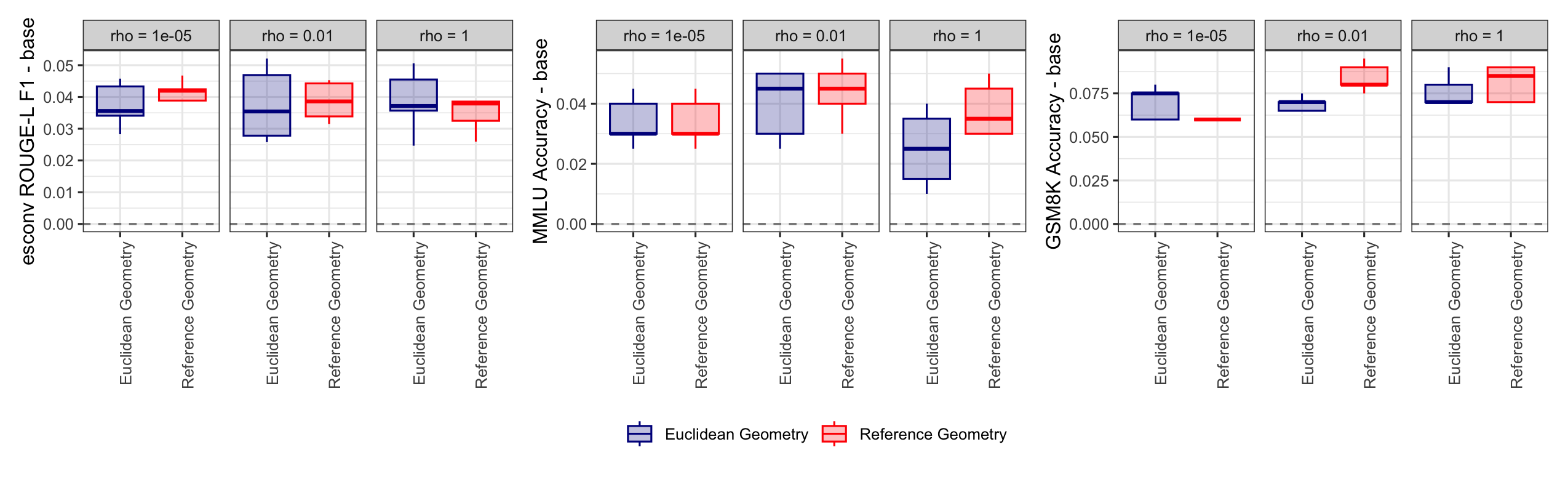}
    \caption{Target gain and off-target drift $\Dtask$ of \safe{} using Euclidean vs reference Fisher-induced norm objective in Esconv experiment (Setting 3). We plot the difference from the base model across drift budget $\rho$.}
    \label{fig:ablation_geometry_esconv}
\end{figure}

\begin{figure}[t]
    \centering
    \includegraphics[width=0.9\linewidth]{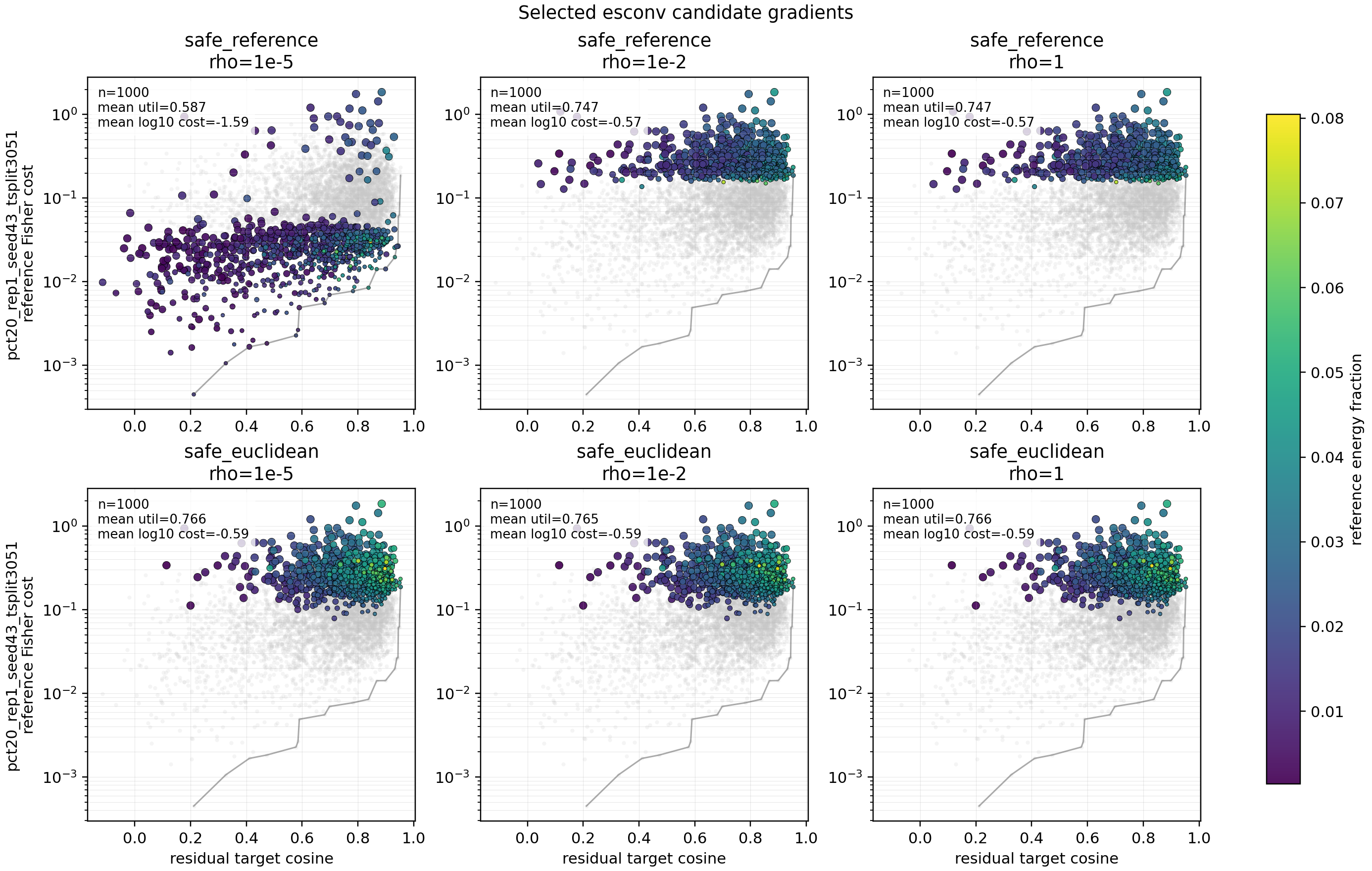}
    \caption{Selected points plotted on target gain versus realized off-target drift $\Dkl$ from \safe{} using Euclidean vs reference Fisher-induced norm objective in Esconv experiment (Setting 3).}\label{fig:ablation_geometry_esconv_gradients}
\end{figure}

We evaluate the effect of using the reference Fisher-induced norm and the Euclidean norm in the data selection objective in Eq.~\eqref{eq:selector-objective}. We present results for the MedQA and Esconv datasets corresponding to Settings 2 and 3.
We report metrics on the target and reference tasks with varying off-target drifts $\rho$ (Figures~\ref{fig:ablation_geometry_esconv} and \ref{fig:ablation_geometry_medqa}), and visualize the selected gradients (Figures~\ref{fig:ablation_geometry_esconv_gradients} and \ref{fig:ablation_geometry_medqa_gradients}). Overall, we see that selecting data points in the Euclidean geometry yields worse performance in controlling the off-target drift: we consistently see that, in the Euclidean geometry, selected gradients have high projections along the reference geometry, irrespective of the value of $\rho$. By contrast, selecting points in the reference geometry yields a stronger sensitivity to higher off-target costs. 

\subsubsection{Effect of the optimizer preconditioner}
\label{app:preconditioner_ablation}
\begin{figure}[t]
    \centering
    \includegraphics[width=\linewidth]{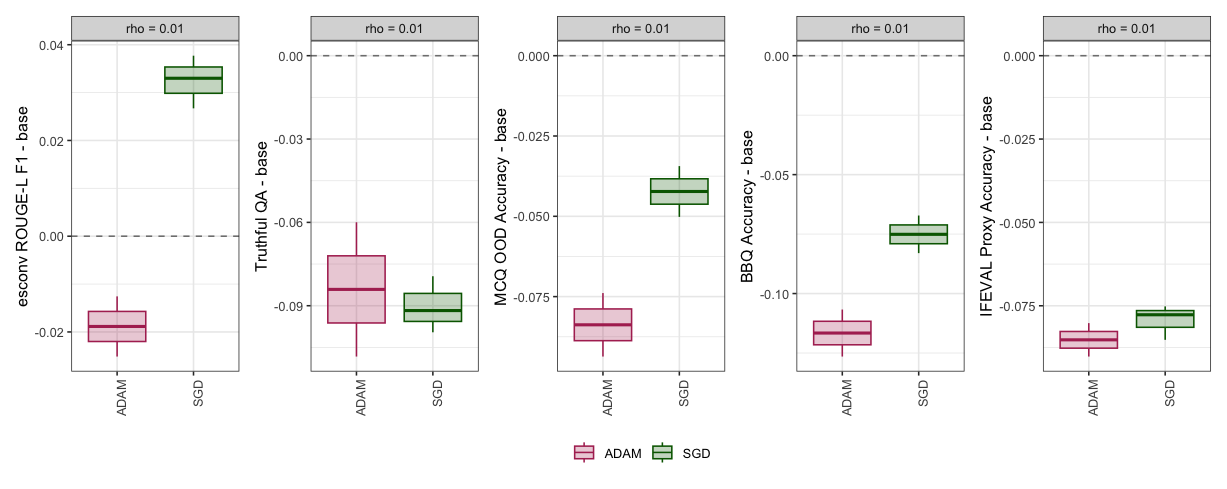}
    \caption{Comparison of using preconditioner $P$ from Adam versus SGD optimizer in MedQA experiment (Setting 2)}\label{fig:ablati_medqa_adam_sgd}
\end{figure}

\begin{figure}[t]
    \centering
    \includegraphics[width=\linewidth]{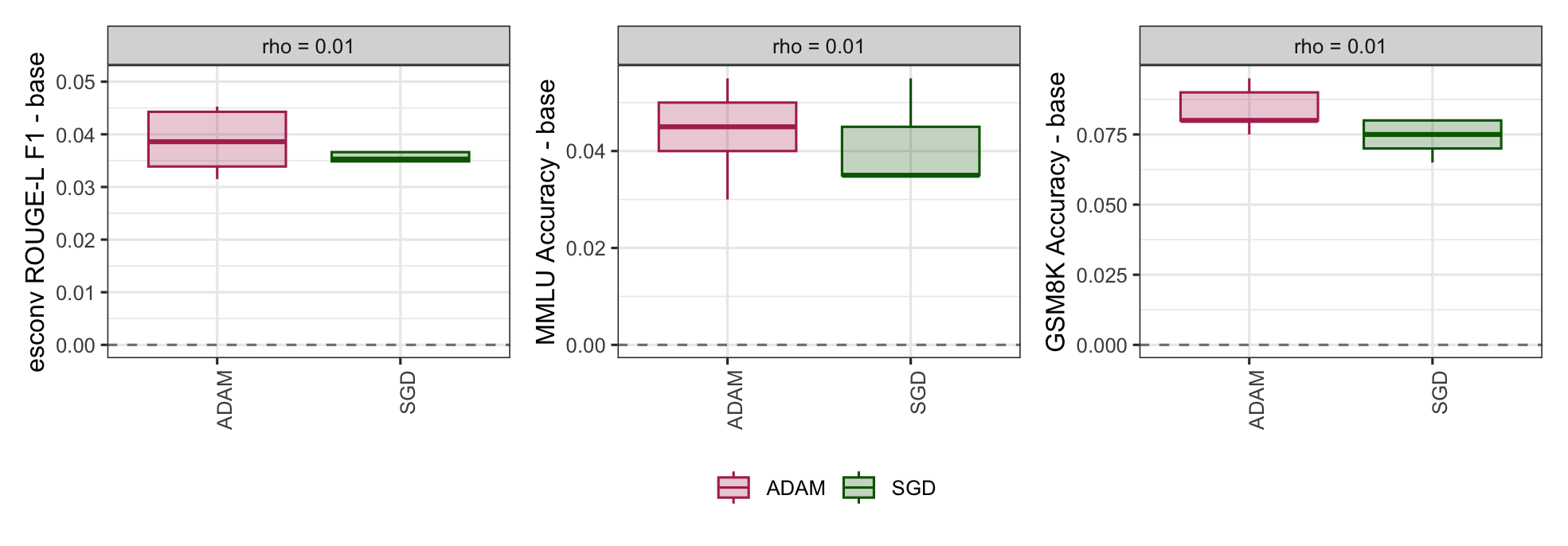}
    \caption{Comparison of using preconditioner $P$ from Adam versus SGD optimizer in Esconv experiment (Setting 3)}\label{fig:ablati_esconv_adam_sgd}
\end{figure}

\begin{figure}[t]
    \centering    \includegraphics[width=0.9\linewidth]{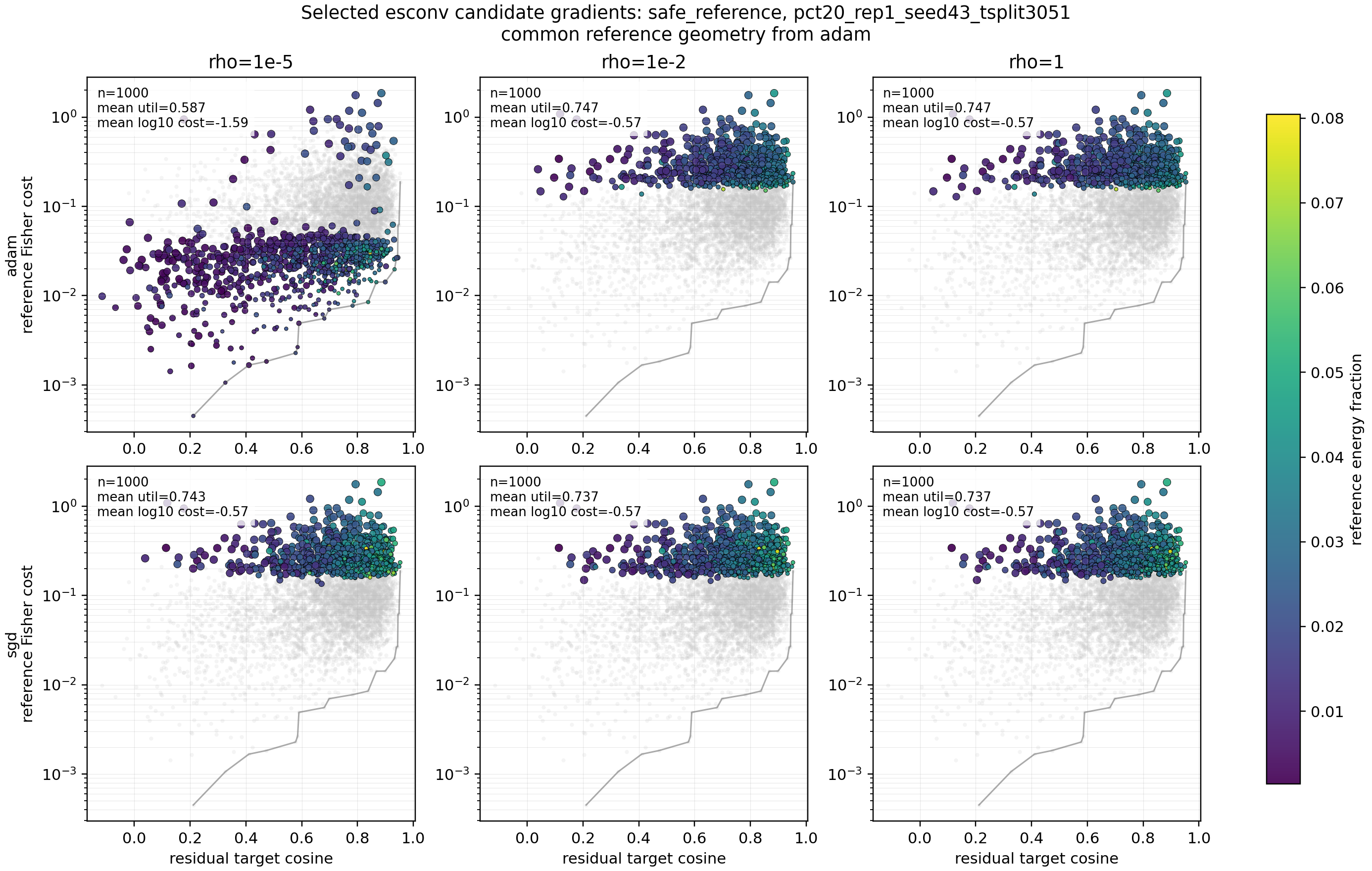}
    \caption{Selected points plotted on target gain versus realized off-target drift $\Dkl$ from \safe{} using preconditioner $P$ from Adam versus SGD optimizer in Esconv experiment (Setting 3)}\label{fig:ablation_geometry_esconv_gradients_adam_sgd}
\end{figure}

We evaluate the effect of the optimizer-induced preconditioner \(\mathbf{P}\). 
Our framework allows different choices of \(\mathbf{P}\), including the identity preconditioner corresponding to SGD-style updates and an Adam-style diagonal preconditioner estimated from warm-up gradients. This ablation tests whether the observed behavior of \safe{} depends critically on using the Adam preconditioner.

Figures ~\ref{fig:ablationrho_experiment_2} and ~\ref{fig:ablati_medqa_adam_sgd}-\ref{fig:ablation_geometry_esconv_gradients_adam_sgd} show that the best preconditioner varies across settings. SGD performs better in Setting 2 while Adam performs better in Setting 3. 
This is expected because the preconditioner changes the effective update induced by each selected example. 
Importantly, both variants retain the same qualitative target-reference trade-off, suggesting that \safe{} is not tied to a single optimizer approximation. 
In practice, we recommend using the preconditioner that matches the downstream fine-tuning optimizer.

\subsubsection[Different choices of rank K\_R and K\_T]{Different choices of rank $K_R$ and $K_T$}
\label{app:K_ablation}\label{app:spectral}

\begin{figure}[t]
    \centering
    \includegraphics[width=1.0\linewidth]{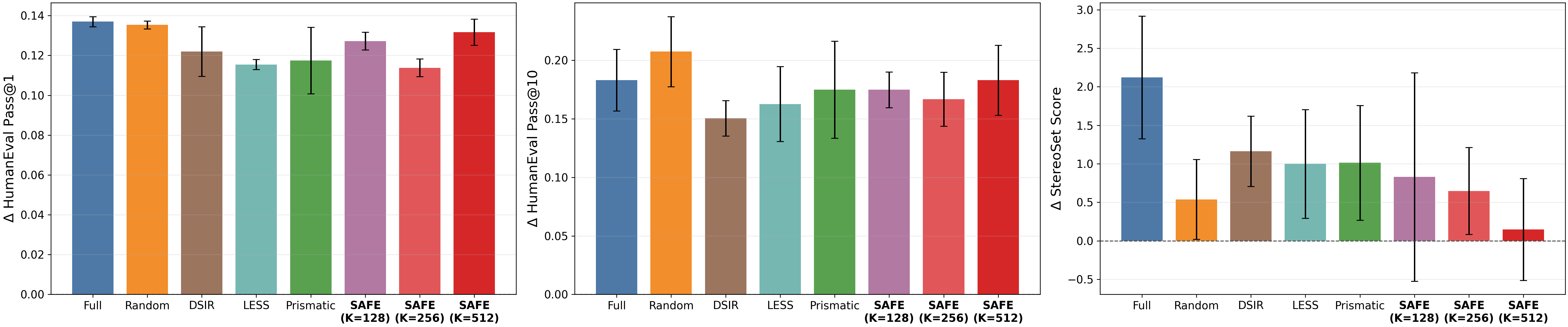}
    \caption{Ablation on the low-rank basis size $(K_R, K_T)$ for \safe{} on Setting~1. Left and center panels show change in HumanEval pass@1 and pass@10 relative to the base model; right panel shows change in StereoSet score relative to the base model.}
    \label{fig:codealpaca_metrics_15}
\end{figure}

\begin{figure}[t]
    \centering
    \includegraphics[width=1.0\linewidth]{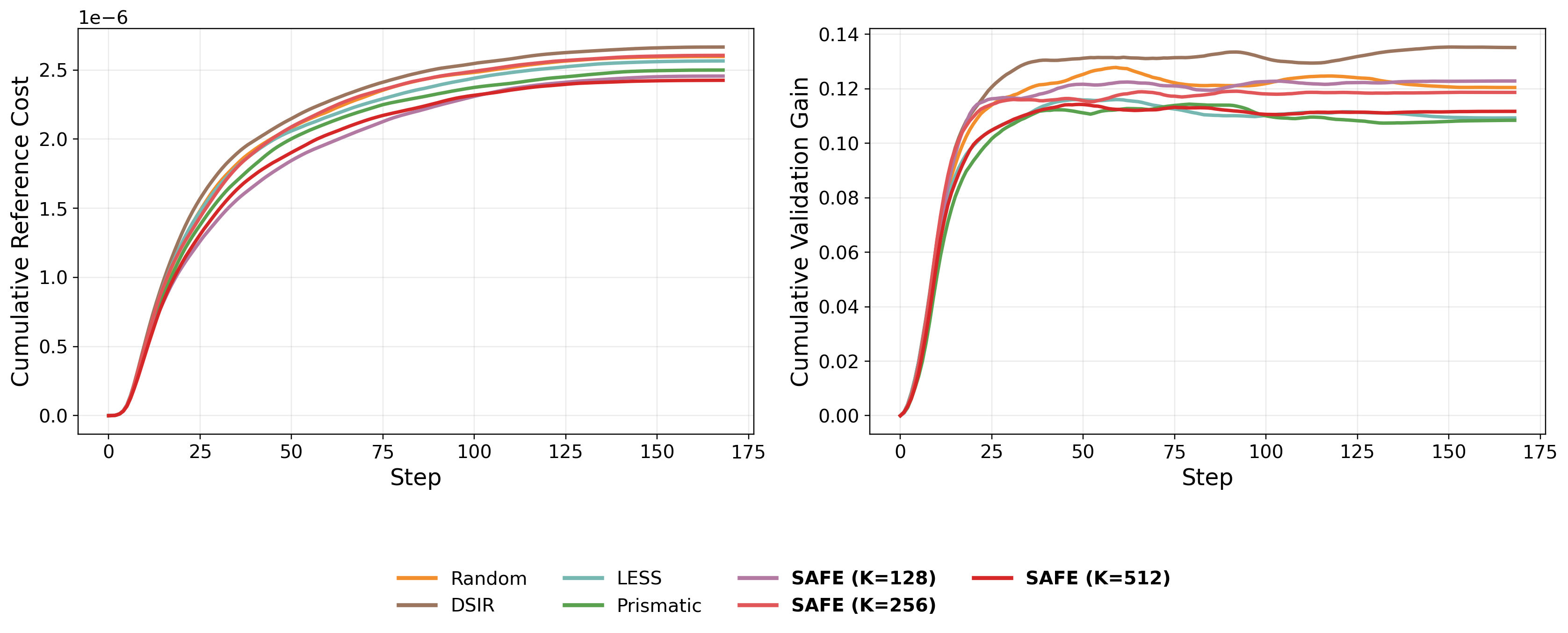}
    \caption{Reference Fisher cost $\Dfish = \Delta\theta_t^\top \hat {\mathbf{F}}_{R}^{(K_R)} \Delta\theta_t$ and projected first-order target-gain proxy $-(\U_K^\top\Delta\theta_t)^\top\hat b_{\mathrm{tar}}^{(K)}$ over training steps, where $\Delta\theta_t=\theta_t-\theta_0$. These use displacement from initialization and fixed initial geometry; they are not a sum of per-step Fisher costs or an exact validation-risk decrease.}
    \label{fig:training_curves}
\end{figure}

We ablate over three low-rank settings $(K_R, K_T) \in \{(128, 32),\, (256, 64),\, (512, 128)\}$ for \safe{} on Setting 1, with results shown in Figure~\ref{fig:codealpaca_metrics_15} for selection budget $15\%$. \safe{} is robust to the choice of $(K_R, K_T)$: all three variants achieve competitive HumanEval pass@1 and pass@10 rates, and all produce smaller changes in StereoSet score than most baselines. One thing to note is that the basis $(K_R, K_T) = (512, 128)$ achieves the smallest off-target drift. In this comparison, the largest basis gives the smallest mean StereoSet change; the smaller bases do not outperform every baseline on that metric.

Figure~\ref{fig:training_curves} further shows the Fisher cost of the displacement from initialization and the projected first-order target-gain proxy over training steps for all methods at the displayed 15\% budget. \safe{} variants have lower plotted reference cost at $(K_R, K_T) = (128, 32), (512, 128)$ than baselines at comparable or higher plotted gain proxy. These are diagnostics of the fixed local model, not guarantees on the fine-tuning trajectory. In particular, the quadratic cost of a total displacement contains cross terms between steps and need not increase with training time; two opposite steps with nonzero Fisher cost have zero final displacement but positive summed step costs.

\begin{table}[t]
\centering
\small
\begin{tabular}{llcccccccc}
\toprule
Setting & Block & $1$ & $8$ & $32$ & $64$ & $128$ & $256$ & $512$ & $1024$ \\
\midrule
\multirow{2}{*}{Setting 1}
 & Reference       & $0.133$ & $0.257$ & $0.414$ & $0.525$ & $0.654$ & $0.794$ & $0.924$ & $1.000$ \\
 & Task/candidate  & $0.074$ & $0.258$ & $0.488$ & $0.627$ & $0.748$ & $0.823$ & $0.885$ & $0.937$ \\
\midrule
\multirow{2}{*}{Setting 3}
 & Reference       & $0.234$ & $0.443$ & $0.685$ & $0.847$ & $1.000$ & ---     & ---     & ---     \\
 & Task/candidate  & $0.118$ & $0.279$ & $0.448$ & $0.564$ & $0.704$ & $0.862$ & $0.9999$ & ---    \\
\bottomrule
\end{tabular}
\caption{\textit{Cumulative spectral energy} (squared singular-value energy) as a function of rank, mean over three seeds. These are energy fractions of the reference-score and task/candidate-gradient matrices. Bounding omitted reference cost additionally requires the spectral scale and update norm; preserving the mean target gradient requires its projection error to be small (Appendix~\ref{app:working_basis}). 
\label{tab:spectral_energy}}
\end{table}

\paragraph{Quality of the Low-Rank Approximation}
The reference Fisher and gradient coordinates are represented in a shared low-rank basis used for selection; this does not restrict the subsequent fine-tuning trajectory to that span. For an update of norm at most $\epsilon$, the omitted empirical reference cost is at most $\epsilon^2\widehat\lambda_{K_R+1}/2$ (Appendix~\ref{app:working_basis}). A retained energy fraction alone cannot supply this bound without the spectral scale and update norm, and retained task/candidate energy need not preserve the mean target-gradient direction.

Table~\ref{tab:spectral_energy} reports cumulative squared singular-value energy, averaged over three seeds. Setting 1 retains 92.4\% of reference energy at $K_R=512$ and 74.8\% of task/candidate energy at $K_T=128$. In Setting 3, the diagnostic matrices use 128 reference and 512 task/candidate examples, explaining their rank limits. Flatter spectra generally require larger ranks for comparable matrix approximation; target alignment additionally depends on the projection error of the target gradient.

\subsubsection{Different selection budgets}
\begin{figure}[t]
    \centering
    \includegraphics[width=0.8\linewidth]{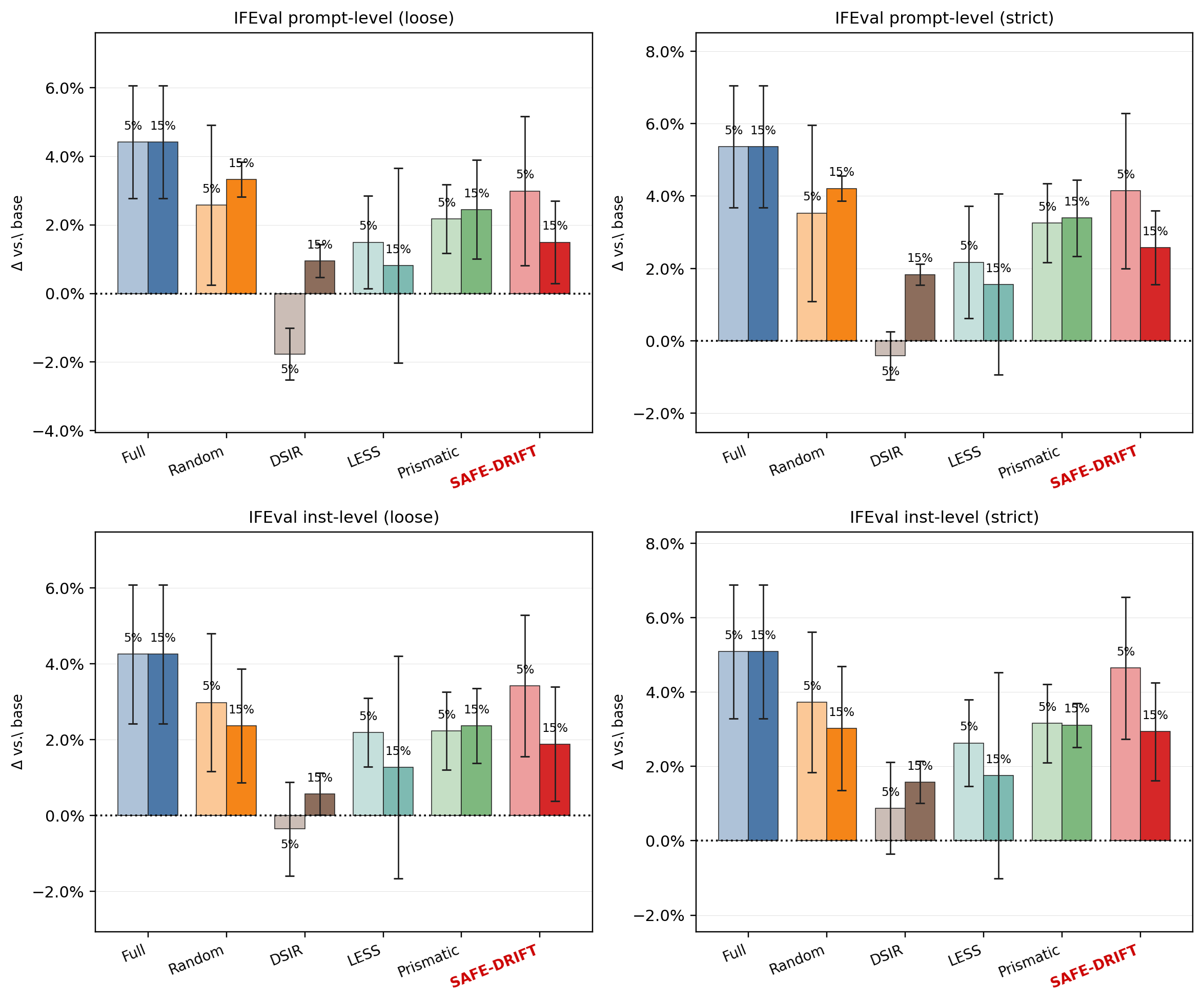}
    \caption{Relative change of IFEval performance compared to the base model. Bar heights are the means and error bars indicate one standard deviation across three seeds. The lighter bar corresponds to a 5\% selection budget and the darker bar to a 15\% budget.}
    \label{fig:ifeval_delta_setting4}
\end{figure}

\begin{figure}[t]
    \centering
    \includegraphics[width=0.9\linewidth]{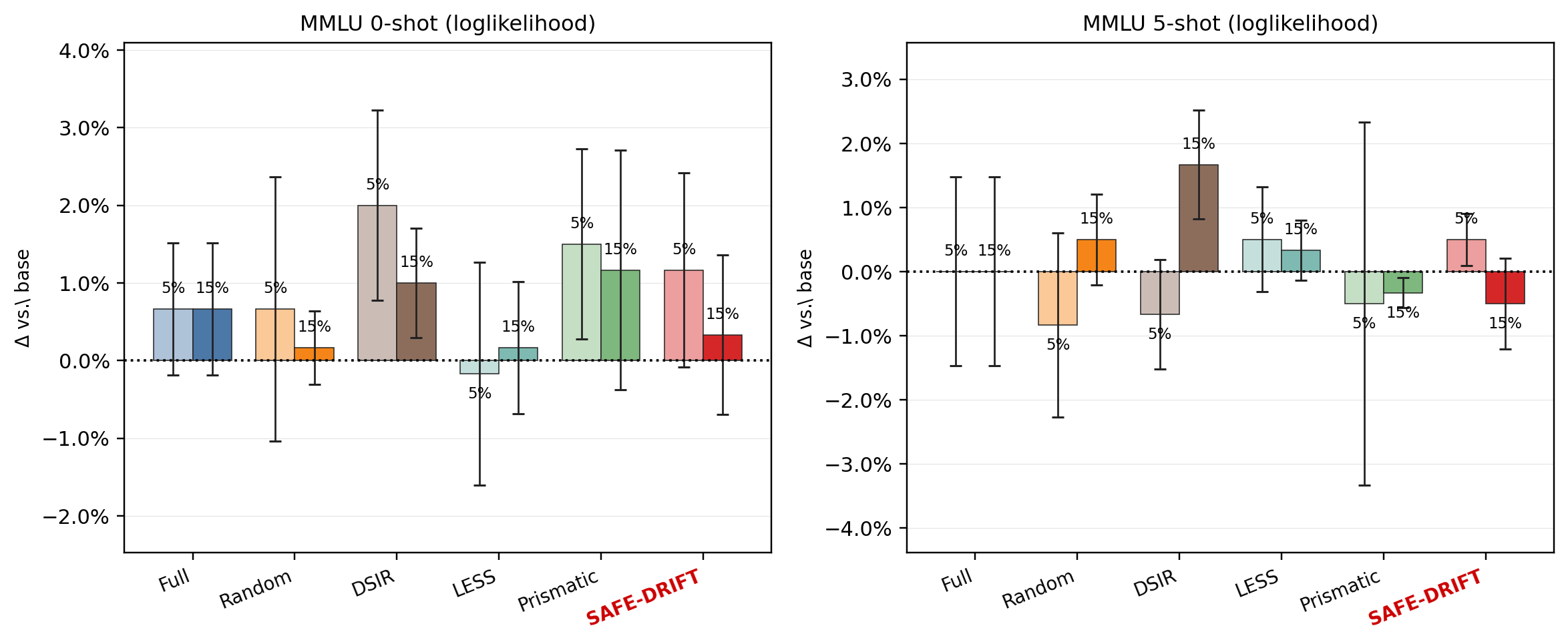}
    \caption{Relative change of MMLU log likelihood of correct answers on the held-out 200 questions compared to the base model: Panels show the 0-shot setting (left) and 5-shot setting (right). The lighter bar corresponds to a 5\% selection budget and the darker bar to a 15\% budget. }
    \label{fig:mmlu_delta_setting4}
\end{figure}

We provide the results of Setting 4 with two selection budgets: 5\% and 15\% of the full candidate pool in Figure~\ref{fig:ifeval_delta_setting4} for IFEval (target task) and Figure~\ref{fig:mmlu_delta_setting4} for MMLU (off-target task). Because the target pool is extremely small, with $42$ IFEval-derived examples in Table~\ref{tab:datasets}, DSIR is particularly brittle in this setting. From Table \ref{tab:datasets}, we know only $22\%$ of candidates fall in $Q_{\mathrm{clean}}$ and $26\%$ in $Q_{\mathrm{ent}}$, so $48\%$ have positive residual-task-block alignment $d_T$. This is not the fraction with positive total target alignment, which also includes the reference block. Figure~\ref{fig:entanglement_plot_setting4} provides complementary block-energy diagnostics, without proving a necessary trade-off in benchmark performance. 

Under this setting, \safe{} concentrates on the highest-scoring target-beneficial examples at the $5\%$ budget, while at $15\%$, \safe{} selects some weakly target-aligned or reference-entangled examples, diluting the target update and preserving its behavior on the reference task. Comparing with LESS, \safe{} gives a more favorable target--reference trade-off and attains more target gains. Random and Prismatic selects a broader subset from the candidate distribution, and they also select many entangled directions that result in MMLU preservation comparable to \safe{}. 

\subsection{Computational Cost}
\label{app:compute}
\safe{} has two computational bottlenecks. The first is per-example gradient construction over the candidate pool, which scales linearly in $N_C$ and is equally required by LESS and Prismatic; the second is the low-rank estimation of the reference Fisher and target gradient, whose marginal cost is small by comparison. For large candidate pools the pool can be subsampled before selection, and for large models, LoRA keeps the per-example gradients tractable.

Table~\ref{tab:flops_full} reports FLOP counts for Settings 1--3. LESS, Prismatic, and \safe{} have equal reported selection counts within each base model, reflecting the shared computation for gradient construction over the candidate pool. 
\begin{table}[t]
\centering
\small
\setlength{\tabcolsep}{5pt}
\resizebox{\textwidth}{!}{%
\begin{tabular}{lllrrrrc}
\toprule
Setting & Model & Method & Select FLOPs & Train FLOPs & Select + Train & Select share & Runs \\
\midrule
\multirow{10}{*}{Setting 1}
 & \multirow{5}{*}{OLMo-2-7B}
   & Random             & $0$                  & $7.24\times10^{15}$ & $7.24\times10^{15}$ & $0.0\%$ & 3 \\
 & & DSIR               & $<10^{8}$            & $4.13\times10^{15}$ & $4.13\times10^{15}$ & $<0.1\%$ & 3 \\
 & & LESS               & $5.54\times10^{16}$  & $6.05\times10^{15}$ & $6.15\times10^{16}$ & $90.2\%$ & 3 \\
 & & Prismatic          & $5.54\times10^{16}$  & $7.74\times10^{15}$ & $6.31\times10^{16}$ & $87.7\%$ & 3 \\
 & & \textsc{Safe-Drift}& $5.54\times10^{16}$  & $1.16\times10^{16}$ & $6.70\times10^{16}$ & $82.7\%$ & 3 \\
\cmidrule(lr){2-8}
 & \multirow{5}{*}{OLMo-2-13B}
   & Random             & $0$                  & $1.39\times10^{16}$ & $1.39\times10^{16}$ & $0.0\%$ & 3 \\
 & & DSIR               & $<10^{8}$            & $7.92\times10^{15}$ & $7.92\times10^{15}$ & $<0.1\%$ & 3 \\
 & & LESS               & $1.06\times10^{17}$  & $1.30\times10^{16}$ & $1.19\times10^{17}$ & $89.1\%$ & 3 \\
 & & Prismatic          & $1.06\times10^{17}$  & $1.48\times10^{16}$ & $1.21\times10^{17}$ & $87.7\%$ & 3 \\
 & & \textsc{Safe-Drift}& $1.06\times10^{17}$  & $2.19\times10^{16}$ & $1.28\times10^{17}$ & $82.9\%$ & 3 \\
\midrule
\multirow{5}{*}{Setting 2}
 & \multirow{5}{*}{OLMo-2-7B-Instruct}
   & Random             & $0$                  & $3.056\times10^{15}$ & $3.06\times10^{15}$ & $0.0\%$ & 5 \\
 & & DSIR               & $4.865\times10^{7}$  & $3.560\times10^{15}$ & $3.56\times10^{15}$ & $<0.1\%$ & 5 \\
 & & LESS               & $1.219\times10^{17}$ & $3.568\times10^{15}$ & $1.25\times10^{17}$ & $97.2\%$ & 5 \\
 & & Prismatic          & $1.219\times10^{17}$ & $3.035\times10^{15}$ & $1.25\times10^{17}$ & $97.6\%$ & 5 \\
 & & \textsc{Safe-Drift}& $1.219\times10^{17}$ & $1.625\times10^{15}$ & $1.24\times10^{17}$ & $98.7\%$ & 5 \\
\midrule
\multirow{5}{*}{Setting 3}
 & \multirow{5}{*}{OLMo-2-7B-Instruct}
   & Random             & $0$                  & $4.420\times10^{15}$ & $4.42\times10^{15}$ & $0.0\%$ & 5 \\
 & & DSIR               & $1.054\times10^{8}$  & $2.951\times10^{15}$ & $2.95\times10^{15}$ & $<0.1\%$ & 5 \\
 & & LESS               & $4.988\times10^{17}$ & $5.405\times10^{15}$ & $5.04\times10^{17}$ & $98.9\%$ & 5 \\
 & & Prismatic          & $4.988\times10^{17}$ & $4.371\times10^{15}$ & $5.03\times10^{17}$ & $99.1\%$ & 5 \\
 & & \textsc{Safe-Drift}& $4.988\times10^{17}$ & $3.006\times10^{15}$ & $5.02\times10^{17}$ & $99.4\%$ & 5 \\
\bottomrule
\end{tabular}%
}
\caption{\textit{Selection and training compute.} All runs use Adam with LoRA. ``Select + Train'' sums the reported selection and training FLOP estimates and excludes unitemized overhead. ``Select share'' is selection FLOPs divided by this subtotal, calculated before display rounding. These subtotals do not represent end-to-end compute.
\label{tab:flops_full}}
\end{table}

\end{document}